\newif\ifarxiv
\arxivtrue

\documentclass[10pt,twocolumn,letterpaper]{article}
\usepackage{cvpr}              

\usepackage{microtype}

\renewcommand{\paragraph}[1]{\vspace{.5em}\noindent\textbf{#1.}}

\usepackage{xcolor}         
\usepackage{array}
\usepackage{caption}       
\usepackage{subcaption}
\usepackage{pgffor} 
\usepackage{multirow}
\usepackage{makecell}
\newcolumntype{C}[1]{>{\centering\arraybackslash}m{#1}} 
\newcommand{\coltitle}[1]{%
  \parbox[c][2.4em][c]{\linewidth}{\centering\textbf{\scriptsize #1}}%
}
\newcommand{\rowlabel}[2][0cm]{%
  \raisebox{\dimexpr .25#1 - .25ex\relax}[.5#1][.5#1]{%
    \rotatebox[origin=c]{90}{%
      \makebox[#1][c]{\bfseries\footnotesize #2}%
    }%
  }%
}

\newcommand{\coltitlelarge}[1]{%
  \parbox[c][2.0em][c]{\linewidth}{\centering\textbf{#1}}%
}

\newenvironment{finalfigure}
  {\begingroup\setkeys{Gin}{draft=false}\begin{figure}}
  {\end{figure}\endgroup\setkeys{Gin}{draft=true}}
\newenvironment{finalfigure*}
  {\begingroup\setkeys{Gin}{draft=false}\begin{figure*}}
  {\end{figure*}\endgroup\setkeys{Gin}{draft=true}}

\definecolor{cvprblue}{rgb}{0.21,0.49,0.74}
\usepackage[pagebackref,breaklinks,colorlinks,allcolors=cvprblue]{hyperref}

\def\paperID{15445} 
\def\confName{CVPR}
\def\confYear{2026}

\title{Using Gaussian Splats to Create High-Fidelity Facial Geometry and Texture}

\author{Haodi He\\
Epic Games, Stanford University\\
{\tt\small hardyhe@stanford.edu}
\and
Jihun Yu\\
Epic Games\\
{\tt\small jihun.yu@epicgames.com}
\and
Ronald Fedkiw\\
Epic Games, Stanford University\\
{\tt\small rfedkiw@stanford.edu}
}

\begin{document}
\maketitle

\begin{abstract}
We leverage increasingly popular three-dimensional neural representations in order to construct a unified and consistent explanation of a collection of uncalibrated images of the human face. Our approach utilizes Gaussian Splatting, since it is more explicit and thus more amenable to constraints than NeRFs. We leverage segmentation annotations to align the semantic regions of the face, facilitating the reconstruction of a neutral pose from only 11 images (as opposed to requiring a long video). We soft constrain the Gaussians to an underlying triangulated surface in order to provide a more structured Gaussian Splat reconstruction, which in turn informs subsequent perturbations to increase the accuracy of the underlying triangulated surface. The resulting triangulated surface can then be used in a standard graphics pipeline. In addition, and perhaps most impactful, we show how accurate geometry enables the Gaussian Splats to be transformed into texture space where they can be treated as a view-dependent neural texture. This allows one to use high visual fidelity Gaussian Splatting on any asset in a scene without the need to modify any other asset or any other aspect (geometry, lighting, renderer, etc.) of the graphics pipeline. We utilize a relightable Gaussian model to disentangle texture from lighting in order to obtain a delit high-resolution albedo texture that is also readily usable in a standard graphics pipeline. The flexibility of our system allows for training with disparate images, even with incompatible lighting, facilitating robust regularization. Finally, we demonstrate the efficacy of our approach by illustrating its use in a text-driven asset creation pipeline.



\end{abstract}
\section{Introduction}
Facial avatars are essential for a wide range of applications including virtual reality, video conferencing, gaming, feature films, etc. As virtual interactions become more prevalent, the demand for compelling digital representations of human faces will continue to grow. A person's face avatar should accurately reflect their identity, while also being controllable, relightable, and efficient enough to use in real-time applications. These requirements have driven significant research endeavors in computer vision, computer graphics, and machine learning. However, it is still challenging to create such avatars in a scalable and democratized way, i.e.~with commodity hardware and limited input data and without using multiple calibrated cameras or a light-stage. 

Neural Radiance Fields (NeRFs)~\cite{mildenhall2021nerf} have become increasingly prevalent in both computer vision and computer graphics due to their impressive ability to reconstruct and render 3D scenes from 2D image collections. 
In particular, various authors have achieved impressive photorealistic results on 3D human faces. The implicit nature of the NeRF representation facilitates high-quality editing (see e.g. \cite{comas2024magicmirror}), since it keeps edits both non-local and smooth (similar to splines, but dissimilar to triangulated surfaces). In addition, the regularization provided by the low dimensional latent space keeps edits of faces looking like faces. It is far more difficult for non-expert users to edit triangulated surface geometry and textures directly.



Although only recently proposed, Gaussian Splatting~\cite{kerbl20233d} has quickly become remarkably popular. Its explicit nature makes it significantly more amenable to various constraints than other (typically implicit) neural models. Importantly, NeRFs and Gaussian Splatting complement each other, as one can edit a NeRF representation before converting it into a Gaussian Splatting model for various downstream tasks that would benefit from having a more structured and constrained representation.

    
\begin{figure*}[t]
  \centering
  \includegraphics[width=0.8\linewidth]{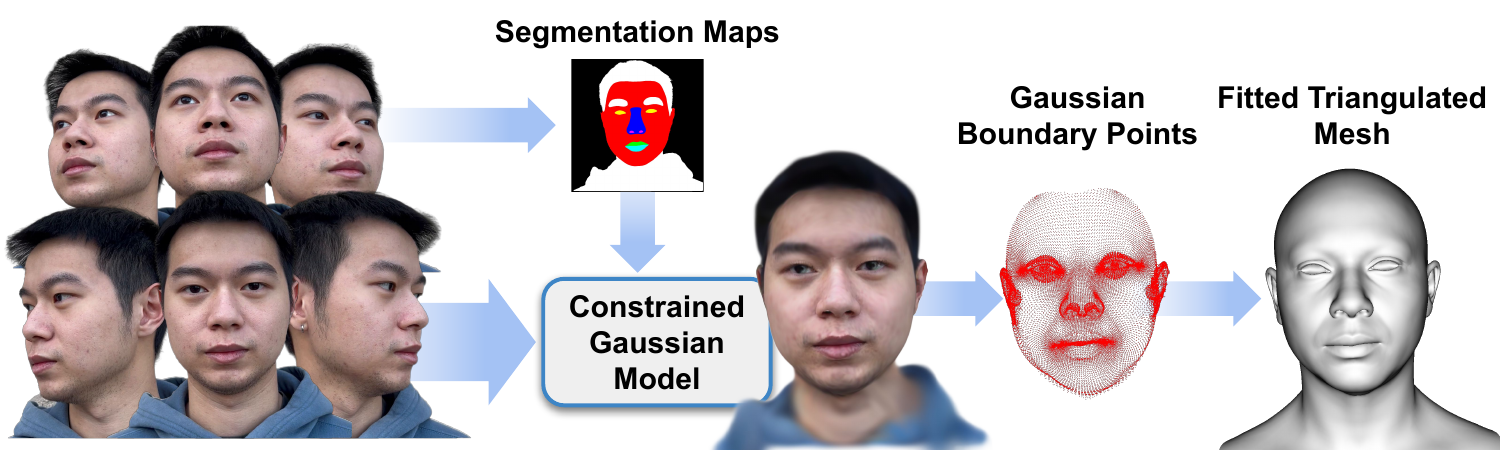}
  \caption{Using a small number of self-captured uncalibrated multi-view images, we use segmentation annotations along with size and shape constraints to force the Gaussians to move instead of deform. In addition, soft constraints are used to keep the Gaussians tightly coupled and close to the triangulated surface. After training, in a post process, the triangulated surface is deformed to better approximate the Gaussian reconstruction.}
  \label{fig:geo_pipeline}
\end{figure*} 

The standard graphics pipeline has been refined over decades via both software and hardware optimizations and has become quite mature, especially for real-time applications. In fact, in spite of the maturity and impact of ray tracing, it has only recently been somewhat incorporated into gaming consoles and other real-time applications. Silicon chip development is often a zero-sum game where adding one capability necessarily removes another. However, ray tracing has been embraced in non-realtime applications and is often used to create content for real-time applications. It is not a stretch to assume that neural rendering methods will be treated similarly. Thus, converting neural models, such as Gaussian Splatting, into triangulated surfaces with textures disentangled from lighting necessarily increases their immediate impact on real-time computer graphics applications. In this work, we address this by introducing a pipeline that transforms a Gaussian Splatting model trained with self-captured uncalibrated multi-view images into a triangulated surface with de-lit textures. 

Our method offers significant advantages over traditional mesh-based geometry reconstruction, since it does not rely on the ability of advanced shading models to overcome the domain gap between synthetic and real images. Instead, neural rendering is used to close the domain gap, while constraints are used to tightly connect the neural rendering degrees of freedom to an explicit triangulated surface. The Gaussian Splatting model is modified to more tightly couple it to the triangulated surface in two ways: segmentation annotations along with size and shape constraints are used to force the Gaussians to move (instead of deform) in order to explain the data, and soft constraints are used to keep the Gaussians tightly coupled and close to the triangulated surface. After training, in a post process, the triangulated surface is deformed to better approximate the Gaussian reconstruction. See Fig.~\ref{fig:geo_pipeline} and Fig.~\ref{fig:ablation}.

\begin{finalfigure}[bh]
  \centering

  \centering
  \setlength{\tabcolsep}{2pt}
  \renewcommand{\arraystretch}{1.0}
  \newcolumntype{M}[1]{>{\centering\arraybackslash}m{#1}}

  \begin{tabular}{M{0.07\linewidth} C{0.27\linewidth} C{0.27\linewidth} C{0.27\linewidth}}
    {} &
    \coltitle{Reconstructed Image} &
    \coltitle{Gaussians} &
    \coltitle{Mesh} \\[0.4em]

    \rowlabel{Our method} &
    \includegraphics[width=\linewidth]{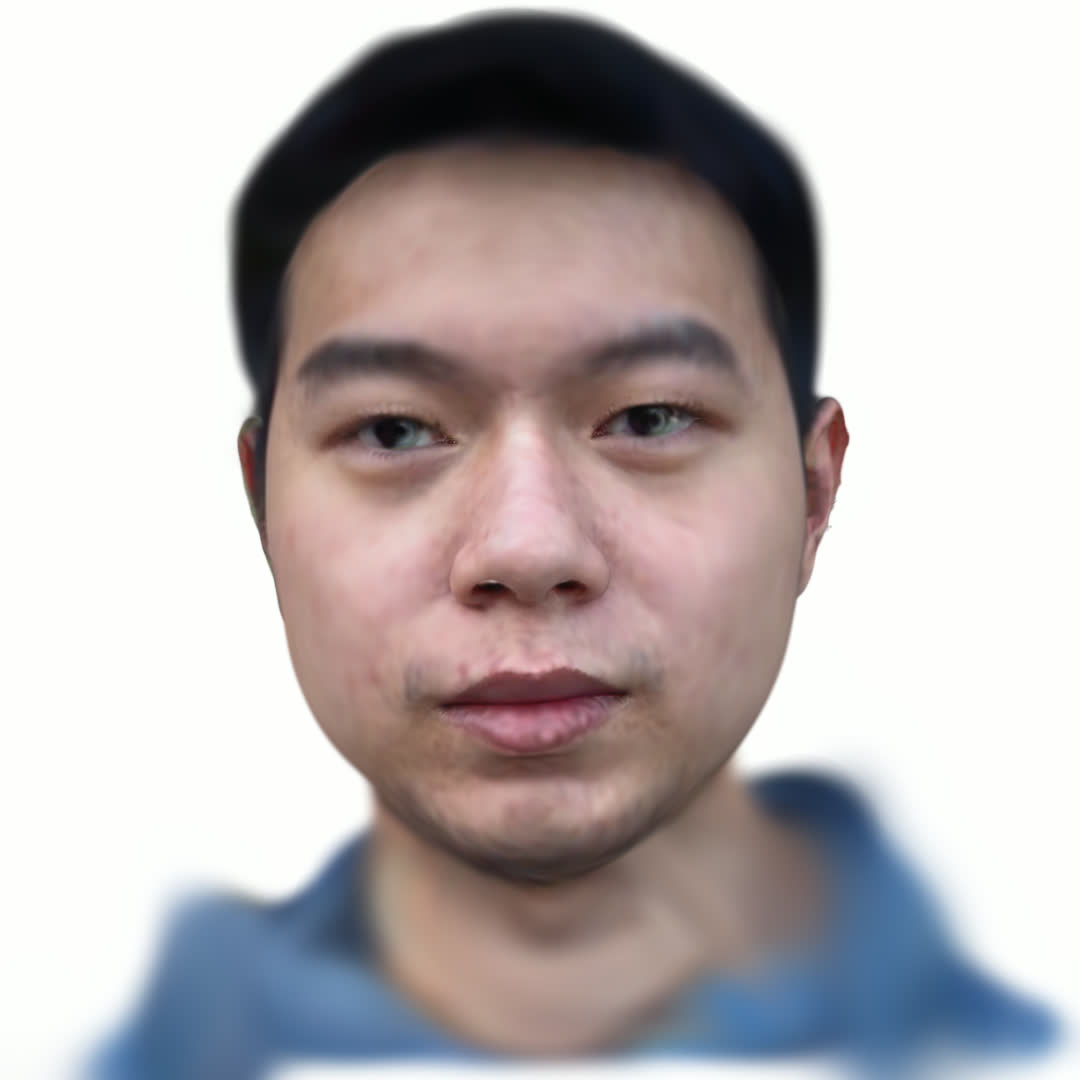} &
    \includegraphics[width=\linewidth]{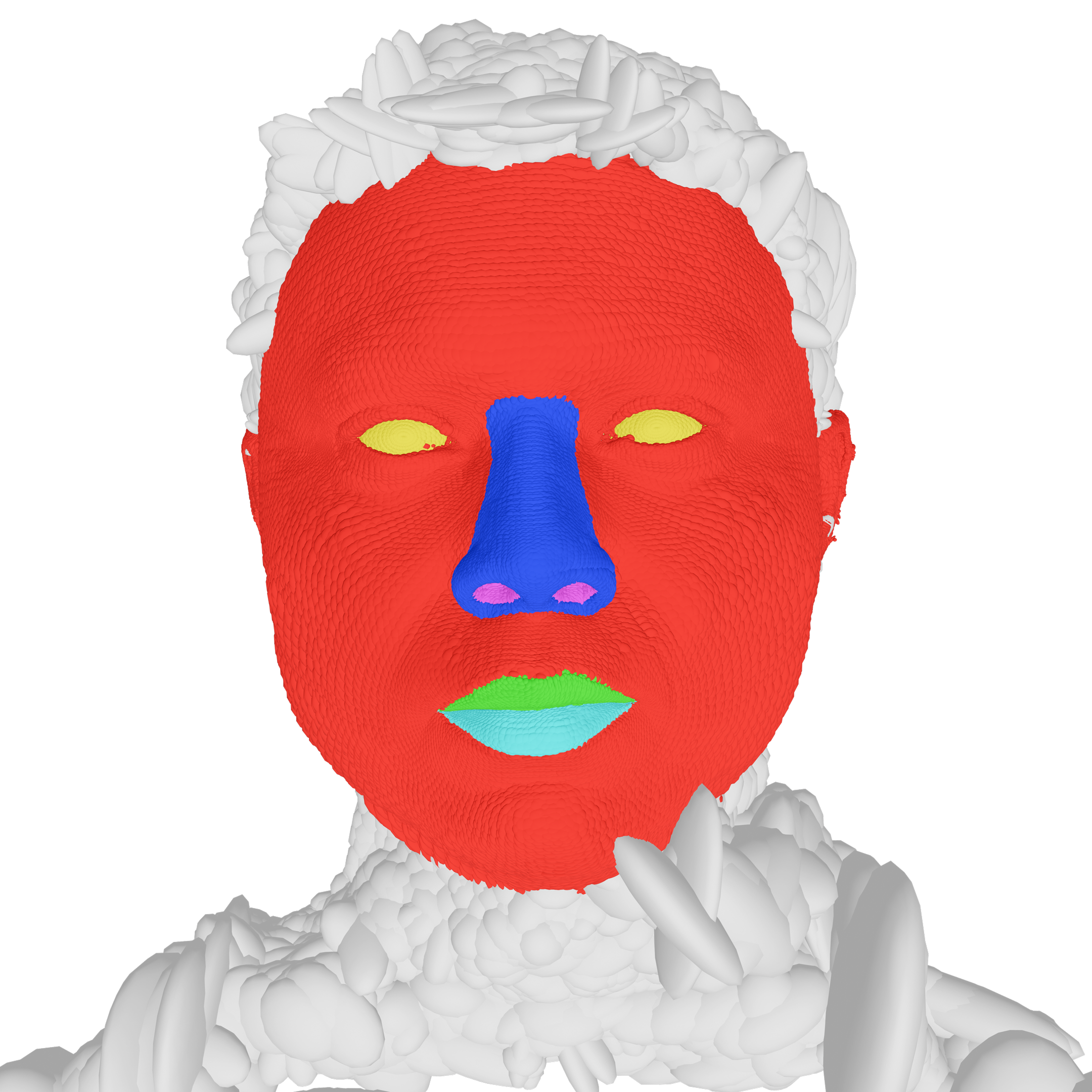} &
    \includegraphics[width=\linewidth]{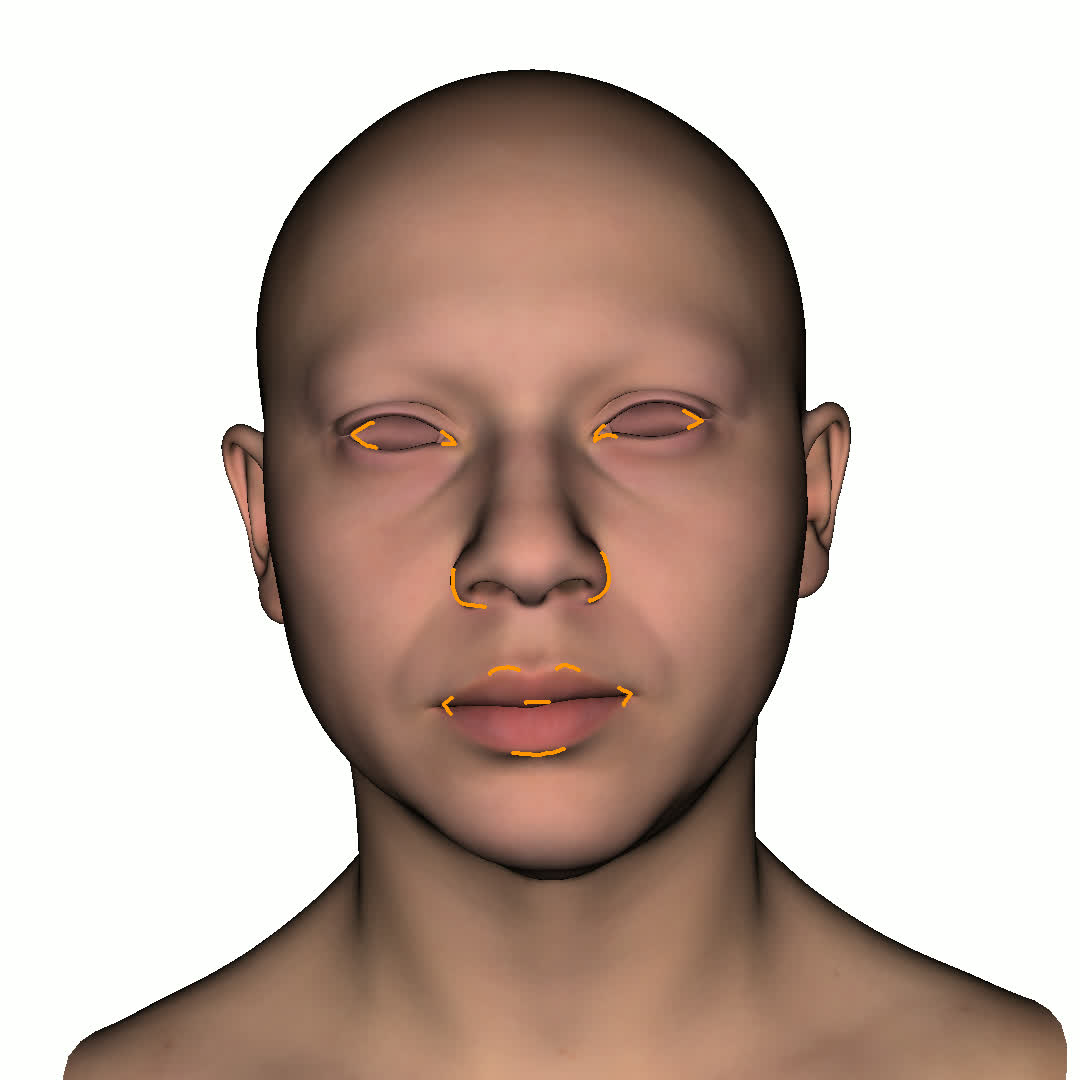} \\[0.35em]

    \rowlabel{w/o Segmentation} &
    \includegraphics[width=\linewidth]{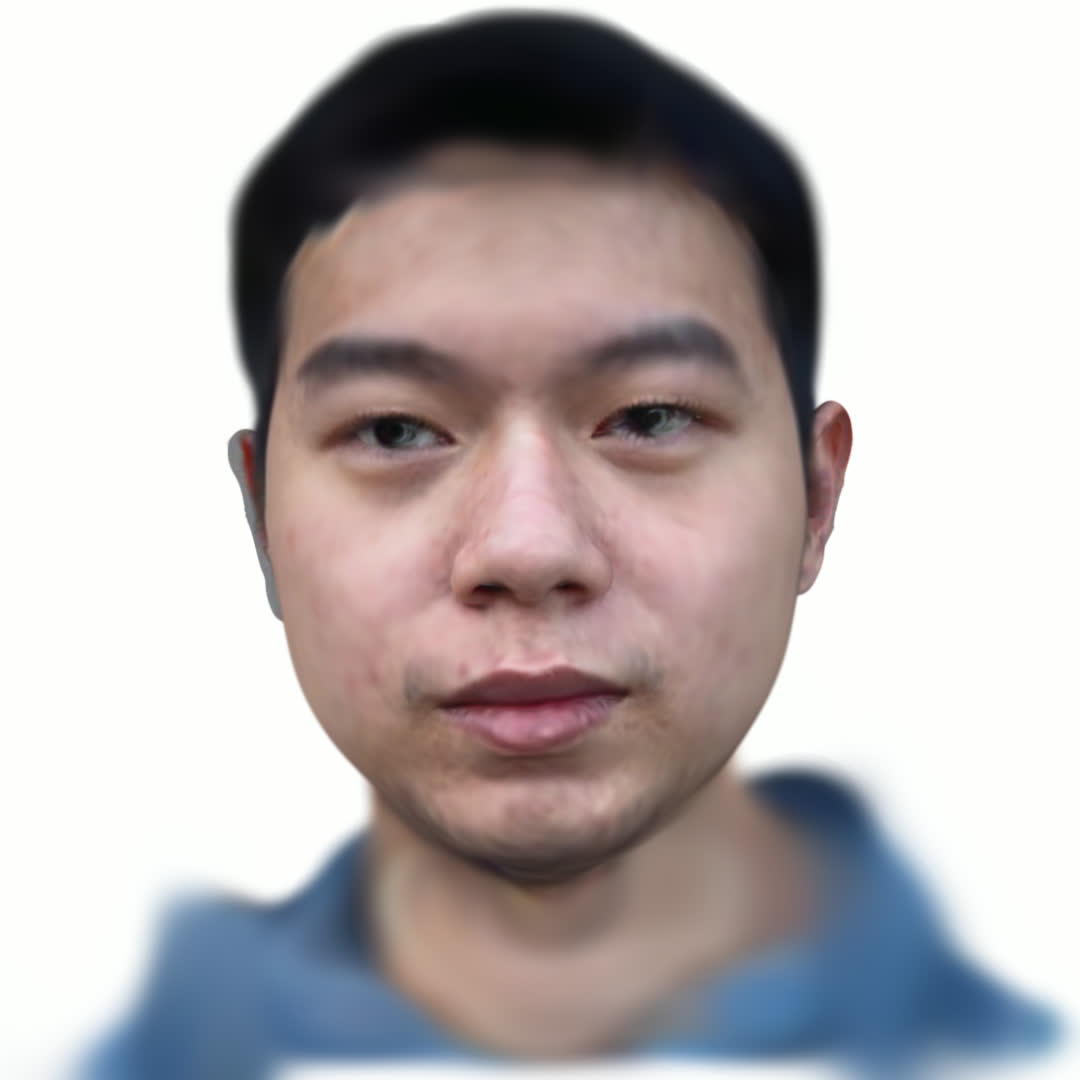} &
    \includegraphics[width=\linewidth]{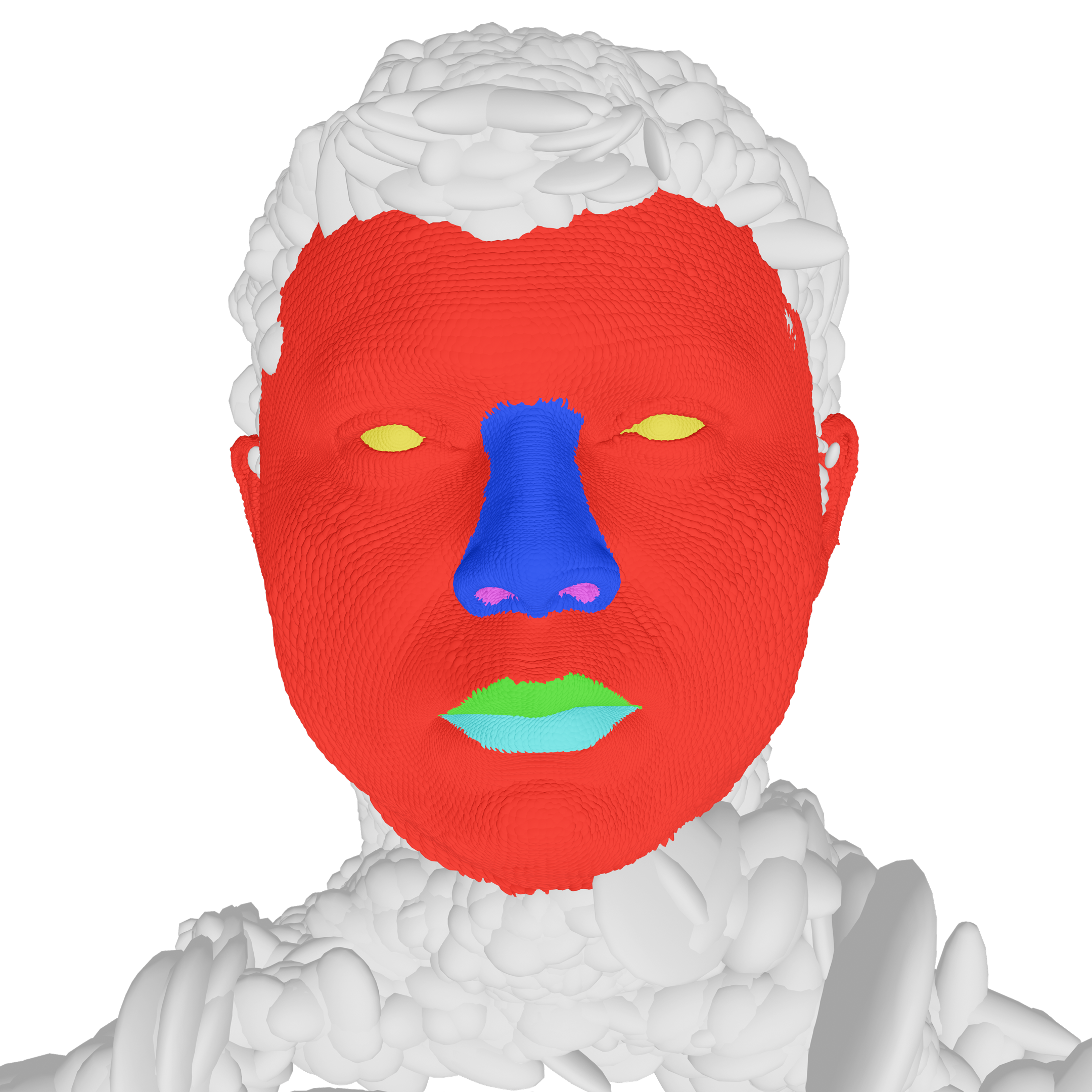} &
    \includegraphics[width=\linewidth]{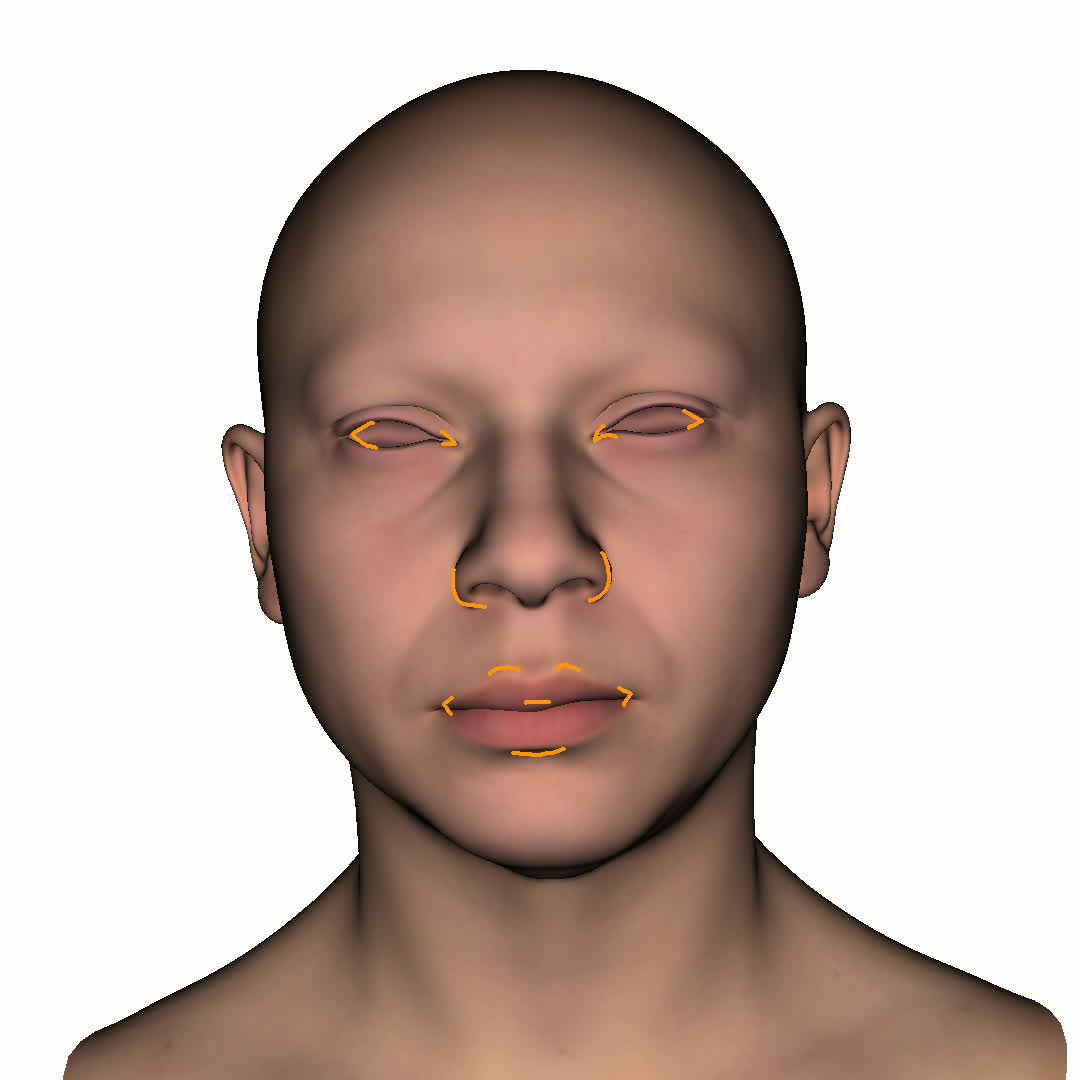} \\

    \rowlabel{w/o Constraints} &
    \includegraphics[width=\linewidth]{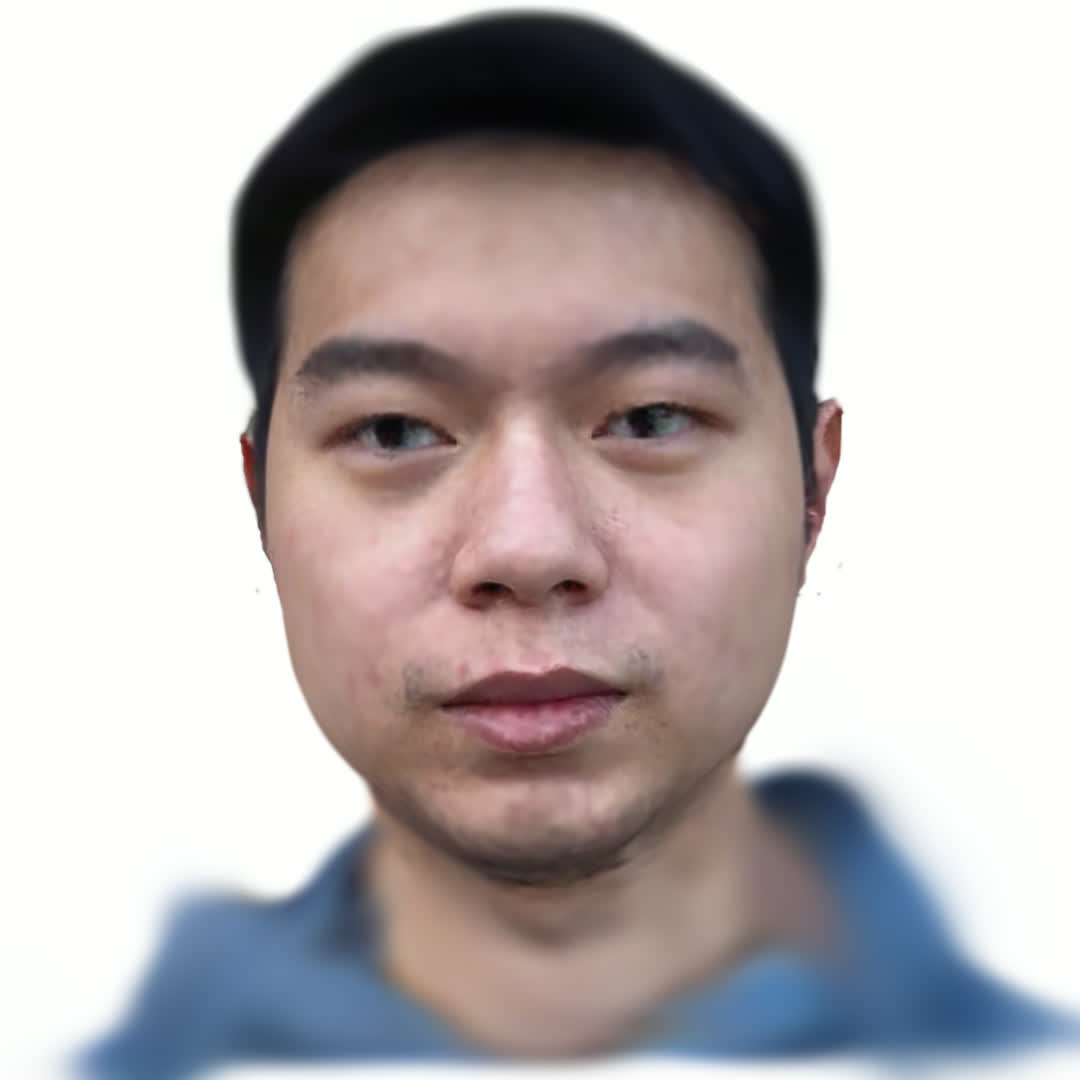} &
    \includegraphics[width=\linewidth]{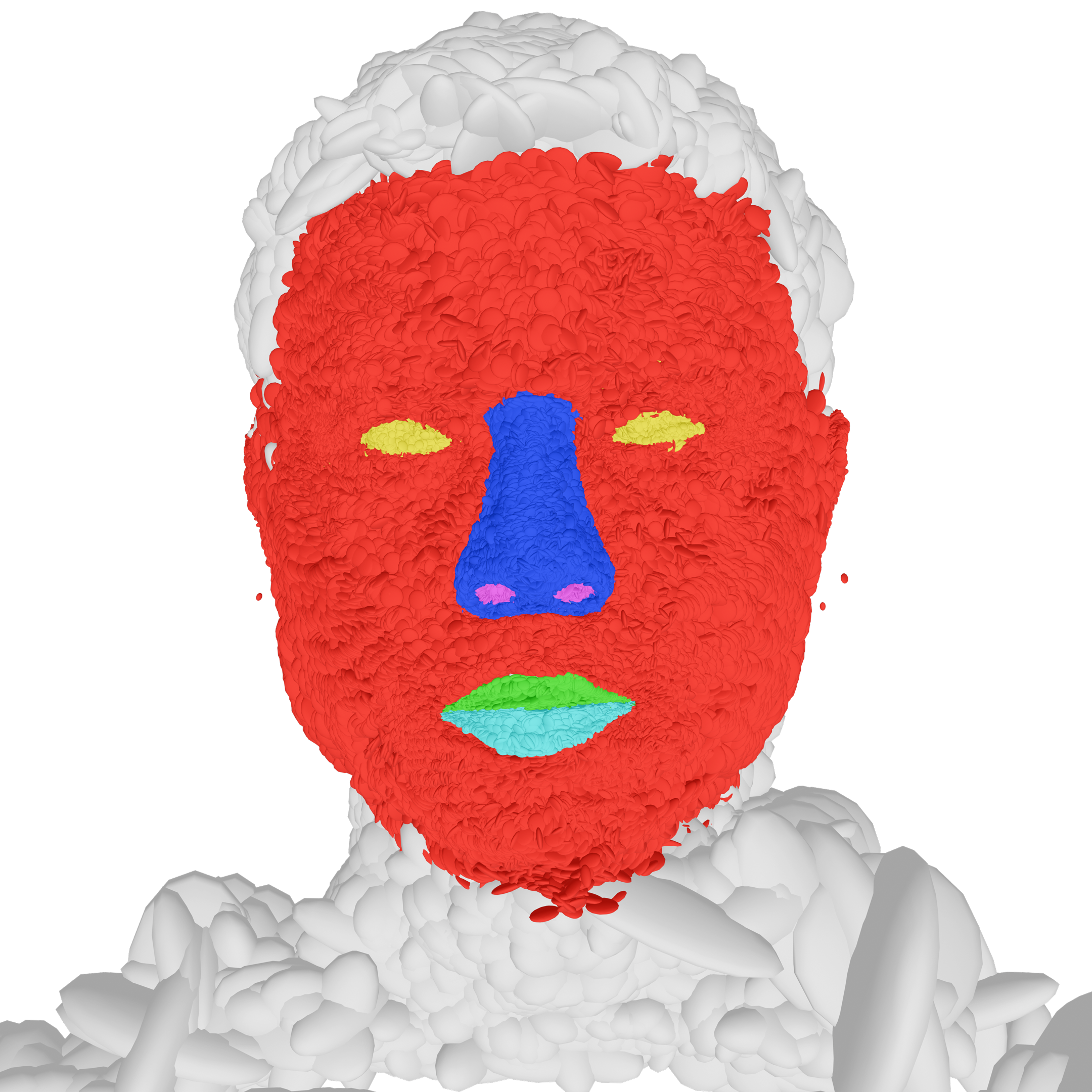} &
    \includegraphics[width=\linewidth]{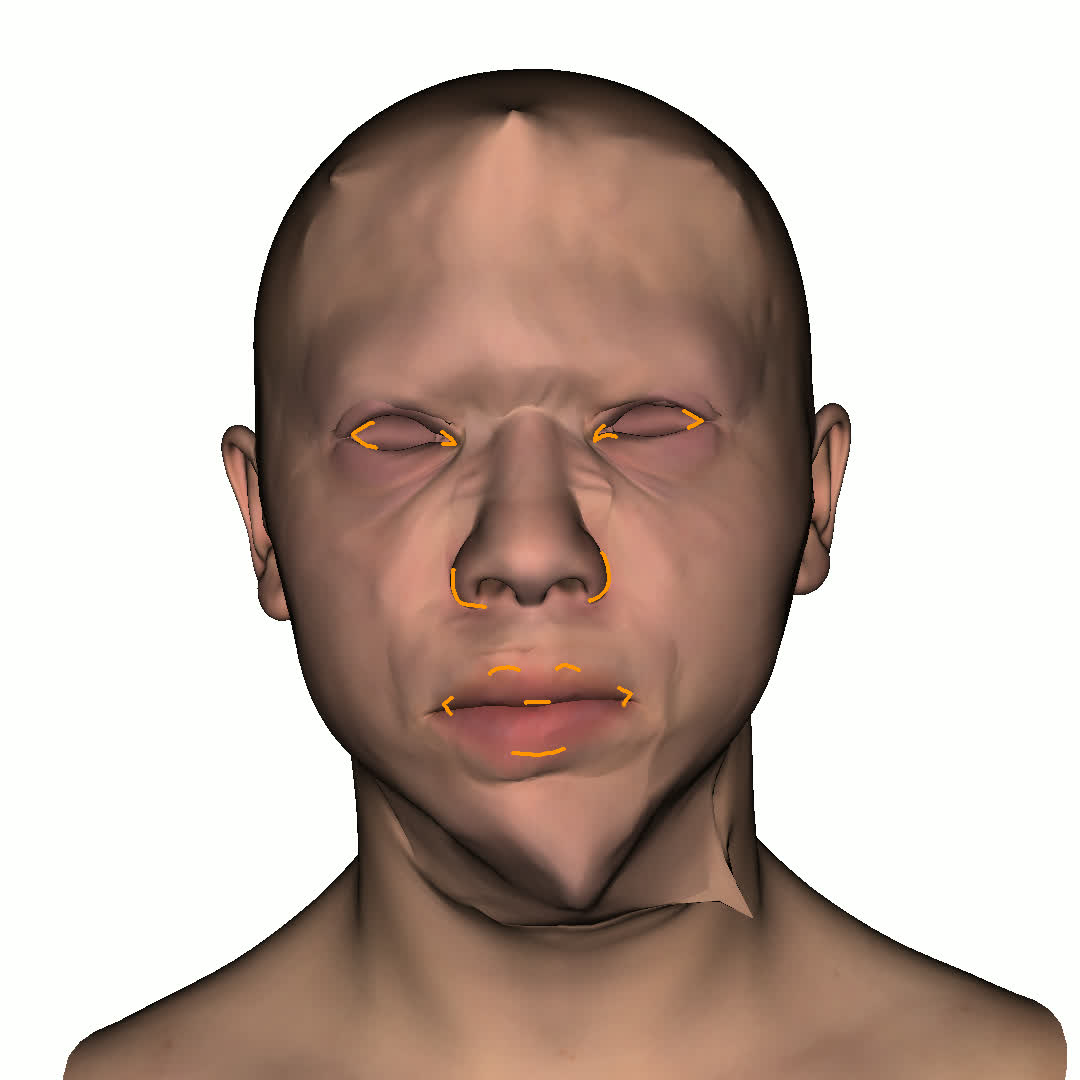} \\

  \end{tabular}
  
  \caption{The ground-truth is well-reconstructed (column 1) not only by our method but also in the ablation tests where segmentation supervision (row 2) and constraints (row 3) have been omitted. Omitting segmentation supervision allows the Gaussians to incorrectly explain regions of the image that their triangles should not be associated with (compare row 2 column 2 to row 1 column 2), resulting in spurious geometry (row 2 column 3). Omitting soft constraints disconnects the Gaussians from their triangles resulting in very spurious geometry (row 3 column 3). }
  \label{fig:ablation}
\end{finalfigure}

 In order to obtain a de-lit texture with albedo disentangled from normal and lighting, without the need of a light-stage, we regularize this underconstrained problem by utilizing a PCA representation of a mesh-based facial texture derived from the Metahuman dataset~\cite{metahuman}. We optimize the PCA coefficients to reconstruct albedo as faithfully as possible, while minimizing reliance on a Relightable Gaussian Splatting model that is used to capture residual differences between a target image and a rendering of the textured mesh.

In summary, we obtain an accurate triangulated surface reconstruction with de-lit high-resolution textures, compatible with a standard graphics pipeline. Moreover, we demonstrate the efficacy of our approach by illustrating its use in a text-driven asset creation pipeline. Key contributions include:
\begin{itemize}
    \item We propose two key modifications of Gaussian Splatting in order to enable accurate triangulated surface reconstruction: Soft constraints encourage Gaussians to be more tightly coupled to the underlying mesh, so that Gaussians perturbations can more accurately drive mesh deformation. Segmentation annotations supervise the Gaussians, so that they do not attempt to explain regions of the target image that they should not be associated with.

    \item We present a method that disentangles albedo textures from lighting and normals. The PCA coefficients of the textured mesh are optimized to capture as much albedo color as possible while minimizing the contributions from the relightable Gaussians that are used to capture differences between the synthetic rendering of the mesh and the target image.

    \item Our method avoids the need of controlled capture setups and instead requires only commodity hardware and limited number of views. In fact, the flexibility of our approach allows for joint training using data from different capture setups; as an example, we combine images from our capture method with those obtained from so-called flashlight capture (see e.g. \cite{han2024high}).

    \item Finally, and perhaps most importantly, we illustrate that the geometry obtained via our pipeline is accurate enough to use in conjunction with a view-dependent neural texture. In particular, we purpose a novel Gaussian Splatting approach to view-dependent neural textures, emphasizing that this enables the utilization of high visual fidelity Gaussian Splatting on any asset in a scene without the need to modify any other asset or any other aspect (geometry, lighting, renderer, etc.) of the graphics pipeline.
\end{itemize}

    
    
    
\section{Related Work}
\paragraph{NeRFs:} Generally speaking, NeRFs disentangle camera intrinsics and extrinsics from facial and other representations by predicting color and density for volumetric ray marching. Some frameworks utilize parameterized models such as 3DMM (e.g.~\cite{gafni2021dynamic,hong2022headnerf,zhuang2022mofanerf}), FLAME (e.g.~\cite{athar2023flame,duan2023bakedavatar}), or related parameterization (e.g.~\cite{gao2022reconstructing}) as inputs; then, expression parameters are used to morph a neutral expression-less face from a canonical space into the scene (e.g.~\cite{teotia2024hq3davatar}). Although NeRFs do not disentangle geometry, coarse geometry is sometimes used to drive the morph (e.g.~\cite{zielonka2023instant,bai2023learning}). Some works (e.g.~\cite{yang2024vrmm, sarkar2023litnerf}) have aimed to relight NeRFs, but the coarse geometry and the entanglement between geometry and texture hinders the ability to disentangle lighting in a manner that facilitates the incorporation of NeRFs into a standard graphics pipeline. Still, the implicit representation and the latent space regularization make NeRFs a powerful editing framework (e.g.~\cite{comas2024magicmirror,li2023instructpix2nerf,sun2022fenerf,jiang2022nerffaceediting}). The latent space regularization also makes NeRFs good candidates for democratized personal avatar generation using only a few images or a quick phone scan (e.g.~\cite{cao2022authentic,buhler2023preface,buehler2024cafca,yang2024vrmm}).

\paragraph{Gaussian Splatting:} Similar to NeRFs, Gaussian Splatting methods use expression parameters to deform a canonical space representation (e.g.~\cite{xiang2024flashavatar, zhao2024psavatar, wang2023gaussianhead, xu2024gaussian}), often with the aid of a triangulated surface (e.g.~\cite{teotia2024gaussianheads, chen2024monogaussianavatar, shao2024splattingavatar, qian2024gaussianavatars}). In fact, triangulated surfaces can be used to parent Gaussians for other structures as well, such as hair (e.g.~\cite{wang2024mega}). Gaussian Splatting has surpassed NeRFs in a number of areas, including for democratized personal avatar generation (e.g.~\cite{chu2024generalizable, saunders2024gaspgaussianavatarssynthetic,he2025lam,liao2025soap,liu2025gaussianavatareditorphotorealisticanimatablegaussian}). We utilize a relightable Gaussian Splatting model (e.g.~\cite{saito2024relightable,li2024uravatar}) in order to disentangle texture from lighting. Most similar to our approach,~\cite{li2024topo4d} leverages Gaussian Splatting to reconstruct a sequence of triangulated surface meshes with high-resolution textures from multi-view light stage videos; in contrast, our method does not require a controlled capture setup and enables de-lighting of the texture.

\paragraph{Other Neural Representations:} Besides NeRFs and Gaussian Splatting, there are number of other neural approaches. The vast majority of these are based on GANs~\cite{goodfellow2020generative}, see e.g.~\cite{trevithick2023real, lin20223d}. Most of the three-dimensional approaches rely on implicit formulations and SDFs (e.g.~\cite{yenamandra2021i3dmm, zheng2022avatar, giebenhain2024mononphm, han2024high}), although some approaches do use explicit geometry (see e.g.~\cite{zheng2023pointavatar}, which uses a point-based approach). Several works (e.g.~\cite{ma2021pixel,grassal2022neural,wang2024mega,sun2025svg}) employ meshes with view and expression dependent neural textures, following~\cite{thies2019deferred}, to further enhance photorealism.

\paragraph{Geometry Reconstruction:} Mesh-based geometry reconstruction typically relies on PCA for regularization to combat noise. Popular examples include 3DMM~\cite{blanz2023morphable}, FLAME~\cite{li2017learning}, and other variants (e.g.~\cite{cao2013facewarehouse, paysan20093d, bao2021high, booth20163d, wang2022faceverse}). Some methods compute additional per-vertex deformations that are added on top of the parameterized geometry. Older approaches use various approximations to rendering (e.g.~\cite{garrido2013reconstructing,garrido2016reconstruction}), while newer approaches use either a fully differentiable ray tracer (e.g.~\cite{dib2023s2f2,wang2022faceverse}) or neural rendering (e.g.~\cite{ ma2021pixel, grassal2022neural, thies2019deferred}). Machine learning approaches rely on the ability to pretrain a model on large datasets (e.g.~\cite{richardson20163d,zhu2016face,cao2013facewarehouse,liu2015deep,sagonas2016300,bagdanov2011florence,wood2021fake,chung2018voxceleb2,yin20063d,raman2023mesh}), inferring mesh parameters using either a single view (e.g.~\cite{richardson20163d, tran2018nonlinear, tewari2018high, tran2019towards, koizumi2020look, zhang2023accurate, yang2020facescape, chai2023hiface, wood20223d}) or multiple views (e.g.~\cite{tewari2019fml, deng2019accurate, bolkart2023instant}). 

\paragraph{Texture and Lighting:} Although chrome spheres have been used to estimate on-set lighting conditions for feature films for three decades (see e.g.~\cite{debevec2008rendering}) and the main ideas behind them originated even two decades before that (see e.g.~\cite{blinn1976texture,miller1984illumination}), much of the prior work still does not disentangle texture from lighting. Of those that do aim to obtain de-lit textures and lighting estimates, the higher-end efforts utilize a light-stage (e.g.~\cite{debevec2000acquiring, ma2007rapid, ghosh2011multiview}), see for example~\cite{saito2024relightable, li2024topo4d}. Using large-scale light-stage datasets (e.g.~\cite{li2020learning, lattas2020avatarme}), neural networks can be trained to predict de-lit textures, surface normals, and lighting conditions (e.g.~\cite{lattas2021avatarme++, sun2019single, nestmeyer2020learning}). See also~\cite{yamaguchi2018high, pandey2021total, wang2020single, kim2024switchlight}. In addition to approaches that leverage a light-stage or light-stage data, a number of approaches aim to disentangle texture from lighting for in-the-wild single-view images under unconstrained conditions:~\cite{tran2019towards, dib2021practical, dib2023s2f2} incorporate physical or statistical regularizations,~\cite{hou2021towards, hou2022face} model shadows to better separate shading from albedo, and~\cite{bai2023ffhq, zhou2019deep} generate pseudo ground-truth data to aid training. Similarly, various approaches utilize multi-view images (e.g.~\cite{wang2023sunstage, zheng2023pointavatar, li2024uravatar, bao2021high}). 




\section{Preliminaries}
\subsection{Gaussian Splatting}\label{sec:gaussian}
Gaussian Splatting~\cite{kerbl20233d} represents a scene using a collection of 3D Gaussian primitives, 
\begin{equation}
    G(\mathbf{x}) = e^{-\tfrac{1}{2} (\mathbf{x} - \boldsymbol{\mu})^\top \Sigma^{-1} (\mathbf{x} - \boldsymbol{\mu})}
\end{equation}
with mean position $\boldsymbol{\mu} \in \mathbb{R}^3$ and covariance matrix $\Sigma \in \mathbb{R}^{3 \times 3}$. The covariance matrix is constructed via $\Sigma = R S S^\top R^\top$ with diagonal anisotropic scaling $S$ and rotation $R$. During rendering, each 3D Gaussian is projected to a 2D Gaussian in screen space to obtain $\boldsymbol{\mu}_{\text{2D}}$ and $\Sigma_{\text{2D}} = J W \Sigma W^\top J^\top$ where $W$ is the viewing transformation and $J$ is the Jacobian of the projective transformation evaluated at $W \boldsymbol{\mu}$. Given an opacity $\alpha \in [0,1]$, the Gaussian makes a contribution of
\begin{equation}
    w(p) = \alpha e^{-\tfrac{1}{2} (p - \boldsymbol{\mu}_{\text{2D}})^\top \Sigma_{\text{2D}}^{-1} (p - \boldsymbol{\mu}_{\text{2D}})}
\end{equation}
 at pixel $p$. Pixel colors are computed using front-to-back alpha compositing via
\begin{equation}\label{eq:comp}
    \sum_{i} \left( \prod_{j<i} (1 - w_j(p)) \right) w_i(p) \mathbf{c}_i(\boldsymbol{v}(\boldsymbol{\mu}_i))
\end{equation}
where the viewing direction $\boldsymbol{v}$ is computed using the mean position of the Gaussian, and the view-dependent color $\mathbf{c}_i$ is computed using $\boldsymbol{v}$ and the Gaussian's learned per-channel spherical harmonic coefficients.






\subsection{Gaussian Avatars} \label{sec:gaussian_avatars}
Gaussian Avatars~\cite{qian2024gaussianavatars} represent deformable head models by attaching Gaussians to the triangles of a mesh. The position and covariance of each Gaussian is defined in the local coordinate system of its parent triangle and driven by the triangle’s motion via
\begin{align}
    \boldsymbol{\mu}(\theta) &= R(\theta)\boldsymbol{\mu}_{\text{local}} + \mathbf{T}(\theta) \\
    \Sigma(\theta) &= s(\theta)^2 R(\theta)\Sigma_{\text{local}} R(\theta)^\top
\end{align}
where $R$, $\mathbf{T}$, and $s$ are defined for each frame via construction. The translation $\mathbf{T}$ is set to be equal to the centroid of the triangle. The first and second consistently chosen edges are used to construct an orthogonal basis for the plane of the triangle via Gram-Schmidt; then, columns of $R$ are defined using those two orthogonal vectors and their cross product. The (scalar) scale factor $s$ is defined by averaging the length of a consistently chosen edge and its altitude. Ideally, $R$, $\mathbf{T}$, and $s$ capture a significant portion of $\boldsymbol{\mu}_{\text{local}}$ and $\Sigma_{\text{local}}$, allowing the learned $\boldsymbol{\mu}_{\text{local}}$ and $\Sigma_{\text{local}}$ to focus on the fine-scale geometric variations that are not captured by the triangle mesh. Moreover, using the same canonical space version of $\boldsymbol{\mu}_{\text{local}}$ and $\Sigma_{\text{local}}$ for every frame adds regularization to combat overfitting. 

Following~\cite{qian2024gaussianavatars}, we regularize the local scaling ($S$ in Sec.~\ref{sec:gaussian}, which is $S_\text{local}$ in $\Sigma_{\text{local}}$) via 
\begin{equation}
\mathcal{L}_{\text{scale}} = \left\| \sum_i \max \left( \hat{\boldsymbol{e}}_i^\top S_{\text{local}}\hat{\boldsymbol{e}}_i, 
\epsilon_{\text{scale}} \right) \hat{\boldsymbol{e}}_i \right\|_2
\end{equation}
aiming to shrink the diagonal entries of $S_{\text{local}}$ that are larger than $\epsilon_{\text{scale}} = 0.6$. Note, $\hat{\boldsymbol{e}}_i$ are the standard basis vector in $\mathbb{R}^3$.

\subsection{Relightable Gaussian Avatars}\label{sec:relightable}

Relightable Gaussian Avatars~\cite{saito2024relightable} modify the color in Eq~\ref{eq:comp}, replacing it with the sum of a diffuse and a specular term. For each of the three color channels, the diffuse component is computed by integrating over the surface of a sphere
\begin{equation}
\label{eq:relightable_diffuse}
    c^{\text{diffuse}} = \rho  \int_{\mathbb{S}} L(\boldsymbol{\omega}) d(\boldsymbol{\omega}) \, d\boldsymbol{\omega}  = \rho \sum_{k}L_k d_k
\end{equation}
where $\boldsymbol{\rho}$ is the learned albedo, $\mathbf{L}$ is the environment lighting, and $\mathbf{d}$ is a learned radiance transfer function. The integral in Eq.~\ref{eq:relightable_diffuse} is simplified into a summation by representing both $\mathbf{L}$ and $\mathbf{d}$ via their spherical harmonics approximations. The view-dependent specular term is defined by
\begin{equation}
\label{eq:relightable_specular}
  \begin{aligned}
    &\mathbf{c}^{\text{specular}}(\boldsymbol{\mu}_i, \mathbf{n}_i, \sigma_i, \widetilde{v}_i) \\&= \widetilde{v}_i(\boldsymbol{v}(\boldsymbol{\mu}_i)) \int_{\mathbb{S}} \mathbf{L}(\boldsymbol{\omega})\, G(\boldsymbol{\omega}, \boldsymbol{v}_r(\boldsymbol{v}(\boldsymbol{\mu}_i), \mathbf{n}_i), \sigma_i)\, d\boldsymbol{\omega}
    \end{aligned}  
\end{equation}
where $\boldsymbol{v}_r = 2(\boldsymbol{v} \cdot \mathbf{n})\, \mathbf{n} - \boldsymbol{v}$ is the mirror reflection direction about a learned surface normal $\mathbf{n}$, and the spherical Gaussian kernel $\text{G}$ depends on a learned specular sharpness $\sigma$. Note that~\cite{saito2024relightable} combined a number of view-dependent terms and aggressively dropped their dependence on $\boldsymbol{\omega}$ so that they could be pulled out of the integral and be represented by the learnable $\widetilde{v}$.
\section{Method}

To summarize, our method reconstructs a (de-lit) textured triangulated surface mesh, suitable for use in a standard graphics pipeline, from either a short monocular video or a set of images. In Section~\ref{sec:data_init}, we describe the data acquisition and initialization steps, which include using landmark detection in order to estimate head pose and to generate a coarse approximation to the triangulated surface.  In Section~\ref{sec:gauss_opt}, we discuss how we use soft constraints and segmentation annotations in order to more tightly couple the Gaussian Splatting model to the underlying triangulated surface. In Section~\ref{sec:geometry_opt}, we discuss how our modified Gaussian Splatting model can be used to drive deformation of the underlying triangulated surface in order to obtain detailed and representative reconstructed geometry. In Section~\ref{sec:neural_texture}, we demonstrate that the geometry obtained via our pipeline is accurate enough to use in conjunction with a view-dependent neural texture. In Section~\ref{sec:lighting_est} and~\ref{sec:texture_opt}, we explain how we generate high-resolution textures disentangled from lighting. Finally, in Section~\ref{sec:mh_generation} we wrap up with the discussion on how to convert our final result into a MetaHuman framework so that it can be used in wide variety of applications.

\subsection{Data Acquisition and Initialization}\label{sec:data_init}

We capture a 4K-resolution monocular video of the subject using the rear camera of an iPhone $14$ held in a fixed position. During recording, the subject slowly rotates their head while maintaining a neutral expression. This is done outdoors in shaded conditions in order to minimize harsh lighting and specular highlights. Afterwards, each video frame is center-cropped to a resolution of 2160 × 2160 pixels. A pretrained landmark detection network~\cite{EpicGames_MetaHumanAnimator} is used to infer both facial and skull landmarks on each frame. Assuming the camera is fixed (as it is in our data acquisition), the skull landmarks are used to estimate the head pose (rotation and translation) for each frame by comparing the inferred landmarks with corresponding landmarks on a 3D canonical template. Although this assumes that the subject and the template have the same size head, we only require a rough pose estimation during the initialization stage.


\begin{figure}[t]
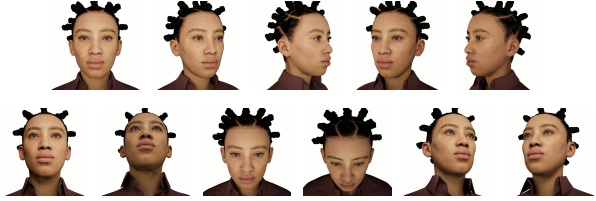

    \centering
    \foreach \i in {00,01,02,03,04} {
        \begin{subfigure}[t]{0.14\linewidth}
            \centering
            \includegraphics[width=\linewidth]{figs/head_poses/\i.jpg}
        \end{subfigure}
    }
    \par\medskip
    \foreach \i in {05,06,07,08,09,10} {
        \begin{subfigure}[t]{0.14\linewidth}
            \centering
            \includegraphics[width=\linewidth]{figs/head_poses/\i.jpg}
        \end{subfigure}
    }
    \caption{The 11 predefined target head poses used for reconstruction.}
    \label{fig:head_poses}
\end{figure}

Eleven predefined target head poses are chosen for the reconstruction, shown in Fig.~\ref{fig:head_poses}.  Given a predefined target head pose, we select the closest non-blurry frame. Closeness is measured by comparing the estimated rotation to the target rotation, and blurriness is measured using the variance of the Laplacian of the pixel values in the grayscale image (as is typical, see e.g. \cite{pech2000diatom}). Weights are used to combine closeness and blurriness into an objective function, so that the process can be automated via minimization.




Starting with a front-facing image, we obtain a rough estimation to depth via~\cite{Agarwal_2023_WACV}. Then, the image, depth estimate, and landmarks are used as input into the MetaHuman Animator~\cite{EpicGames_MetaHumanAnimator} in order to obtain a triangulated surface. The resulting mesh is topologically consistent with other MetaHumans making it compatible with modern graphics pipelines; however, it does not well-represent the subject, given all the rough approximation used in the process. On the other hand, it does provide a good initial guess for Gaussian Splatting. To simplify the model, we remove the triangles associated with the teeth, inner mouth, and 
eyelashes resulting in a face mesh with $47{,}944$ triangles.

\subsection{Modified Gaussian Splatting Model}\label{sec:gauss_opt}

Similar to Gaussian Avatars~\cite{qian2024gaussianavatars}, we attach Gaussians to the triangles of the mesh; however, we do not jointly optimize the triangulated surface along with the Gaussian parameters during training. This decoupling allows us to more readily leverage various improvements in Gaussian Splatting as they become available. It also allows us to independently use mesh regularization and other mesh-based considerations without adversely affecting the Gaussian Splatting. We assign exactly one Gaussian to each mesh triangle and disable both densification and pruning during training. This one-to-one correspondence between Gaussians and mesh triangles makes geometry regularization and subsequent surface reconstruction more straightforward.

It is worth noting two other approaches that modify the Gaussian Splatting model in order to facilitate triangulated surface reconstruction. \cite{Huang2DGS2024} uses flattened two-dimensional Gaussians along with depth distortion and normal consistency regularization terms, and~\cite{guedon2024sugar} uses SDF-based regularization. These modifications all encourage the Gaussians to better align with the surfaces they represent.




\subsubsection{Soft Constraints for Geometric Regularization}
\label{sec:soft_reg}

We promote local geometric consistency of the Gaussians by introducing soft regularization terms that encourage specific geometric features to not vary too much across the mesh. This is accomplished via 
\begin{equation}\label{eq:lap}
\mathcal{L}_{\text{reg}} = \sum_i \left\| \mathbf{z}_i - \frac{1}{|\mathcal{E}(i)|} \sum_{j \in \mathcal{E}(i)} \mathbf{z}_j \right\|^2
\end{equation}
where $\mathcal{E}(i)$ represents Gaussians associated with mesh triangles that share an edge with the triangle containing Gaussian $i$. Note that $i \not\in \mathcal{E}(i)$. $\mathcal{L}_{\text{reg}}$ encourages each $\mathbf{z}_i$ to be equal to the average $\mathbf{z}$ value of its neighbors. This incorrectly weighted Laplacian smoothing (see~\cite{desbrun1999implicit}) is more similar to a standard deviation with the global mean replaced by local means. 

To better regularize the eyeballs, which are disconnected from the face mesh, we identify bi-directional nearest neighbor pairs between triangle centroids of the eyeballs and triangle centroids of the face mesh; then, Gaussians associated with these nearest neighbor pairs are included in the construction of $\mathcal{E}$. 

\paragraph{Center Displacements}
Here, we consider the displacement between each Gaussian center $\boldsymbol{\mu}_i$ and its corresponding mesh triangle centroid $\mathbf{T}_i$ as the geometric feature of interest via $\mathbf{z}_i = \boldsymbol{\mu}_i - \mathbf{T}_i$ in Eq. \ref{eq:lap} in order to obtain $\mathcal{L}_{\text{reg}}^{\text{center}}$. This aims to keep Gaussians as close to the mesh as their neighbors are, helping to regularize subsequent mesh deformation that will be driven in part by these offsets. 

\paragraph{Local Normal} 

Here, we consider the geometric feature $\mathbf{z}_i = \boldsymbol{n}_{i,\text{local}}$, the local normal of Gaussian $i$, in Eq. \ref{eq:lap} in order to obtain $\mathcal{L}_{\text{reg}}^{\text{normal}}$. This aims to keep disagreements in the normal direction between Gaussians and the mesh smoothly varying over the mesh, again helping to regularize subsequent mesh deformation. Note that the method proposed in~\cite{qian2024gaussianavatars}, discussed in Section \ref{sec:gaussian_avatars}, for choosing $R$ causes $\Sigma_\text{local}$ and thus $\boldsymbol{n}_{\text{local}}$ (the third column of $\Sigma_\text{local}$) to vary inconsistently across the mesh. This can be remedied by constructing $R$ more consistently from triangle to triangle, instead of using edge directions that vary significantly across the mesh.
In order to do this, we use UV texture coordinates of the triangle vertices in order to reconstruct fairly consistent U and V directions across the mesh (see e.g.~\cite[Sec.~6.8]{Lengyel2003Math3D2}). Note that discontinuities in the texture coordinates can be ignored by removing neighbors from $\mathcal{E}$. Since U and V are not necessarily orthogonal, we retain U as our consistent direction and orthogonalize V to be perpendicular to U.

\paragraph{Boundary Displacements} \label{sec:boundary} Here, we consider $\mathbf{z}_i = \mathbf{x}_i^* - \mathbf{T}_i$, which is the distance from the outer boundary point $\mathbf{x}_i^*$ of the Gaussian to the triangle centroid $\mathbf{T}_i$, in Eq. \ref{eq:lap} in order to obtain $\mathcal{L}_{\text{reg}}^{\text{boundary}}$. This provides additional regularization (in addition to $\boldsymbol{\mu}_i - \mathbf{T}_i$) that constrains the shape of the Gaussians and more directly affects silhouette boundaries. Whereas~\cite{li2024topo4d} extracts boundary points using only a single Gaussian for each, we include the contribution from neighboring Gaussians in order to more accurately reflect the visible boundaries. In order to do this, we first define $\mathcal{N}(i)$ as a set that contains the Gaussian $i$ itself along with its $k$-nearest neighbors in the UV texture space. Then, a ray is defined via $\mathbf{x}_i = \boldsymbol{\mu}_i + t\mathbf{n}_i$ where $\boldsymbol{\mu}_i$ is the center of the Gaussian, $\mathbf{n}_i$ is the normal of the triangle the Gaussian is associated with, and $t$ is the ray parameter. Next, for each Gaussian $j \in \mathcal{N}(i)$, we solve $\alpha_j G_j(\mathbf{x}_i) = \tau$ in order to find a boundary point of Gaussian $j$ near Gaussian $i$ based on the density threshold $\tau$. Note that we do not consider Gaussians $j \in \mathcal{N}(i)$ that do not have $\alpha_j G_j(\boldsymbol{\mu}_i) < \tau$, which indicates that Gaussian $j$ contains the center of Gaussian $i$ based on the density threshold $\tau$. Finally, $\mathbf{x}_i^*$ is defined via $t_i^*$, where $t_i^*$ is the maximum of the $t_j$ found by solving $\alpha_j G_j(\mathbf{x}_i) = \tau$ for each $j \in \mathcal{N}(i)$.

\subsubsection{Semantic Segmentation for Supervision} 
\label{sec:seg_sup}

\begin{figure}[b]
  
  \centering
  \begin{subfigure}[t]{0.31\linewidth}

    \centering
    \includegraphics[width=\linewidth]{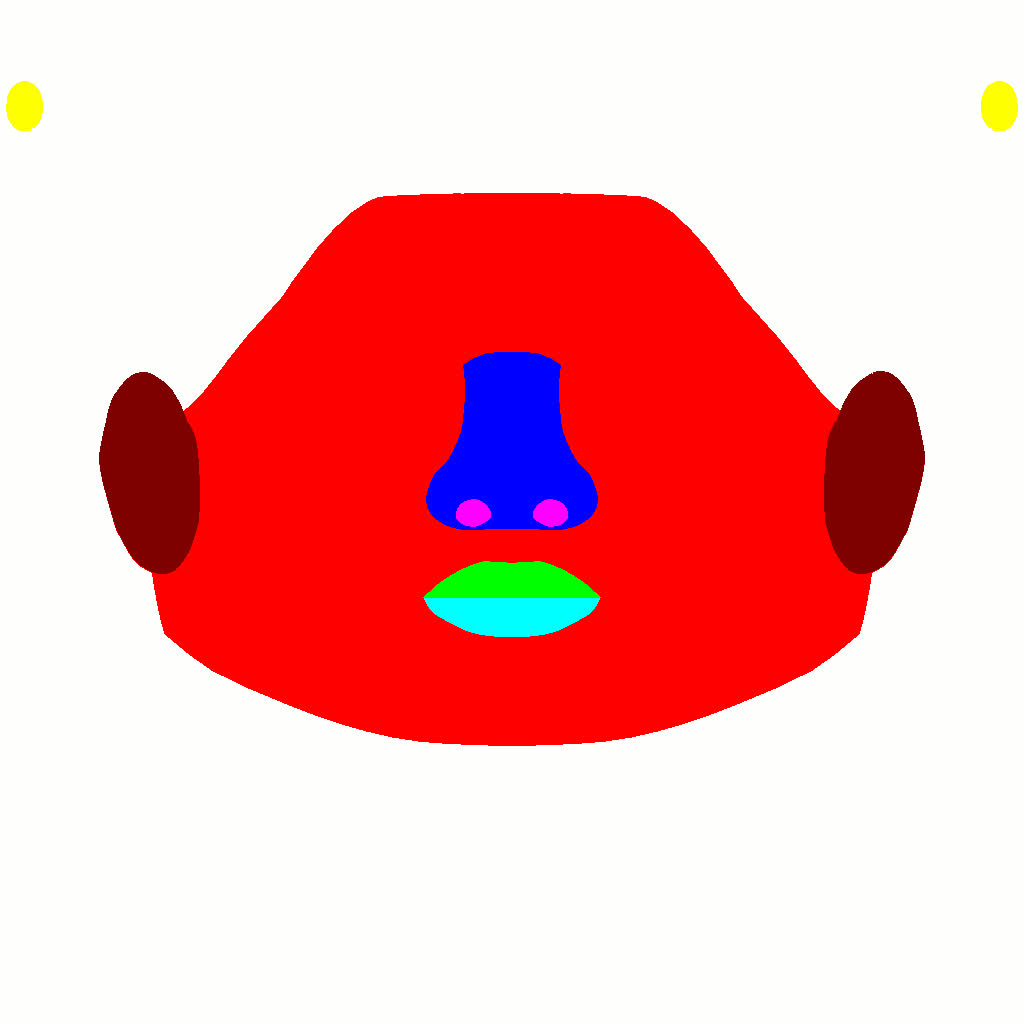}
    \label{fig:seg_map}
  \end{subfigure}
  \hfill
  \begin{subfigure}[t]{0.31\linewidth}
    \centering
    \includegraphics[width=\linewidth]{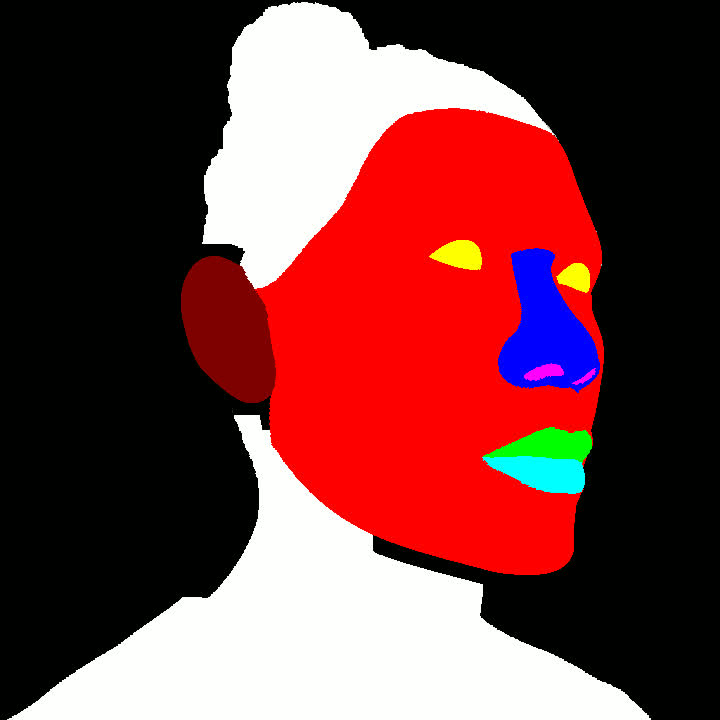}
  \end{subfigure}
  \hfill
  \begin{subfigure}[t]{0.31\linewidth}
    \centering
    \includegraphics[width=\linewidth]{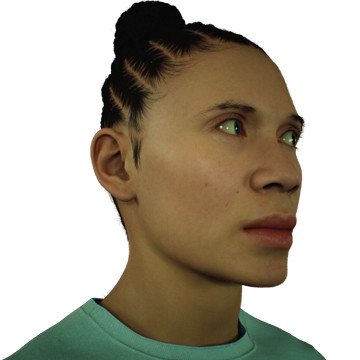}
  \end{subfigure}

  \caption{The annotated texture map (left) used to create training data: face (red), nose (blue), nostril
(pink), top lip (green), bottom lip (cyan), eyes (yellow), ears (dark red), non-face (white). The labeled image (middle) has an additional (black) background label. Note how the (black) background label has been expanded to portions of the foreground in order to emphasize occlusion boundaries.  The segmentation network is trained to recover Fig.~\ref{fig:seg_example} (middle) from Fig.~\ref{fig:seg_example} (right).}

  \label{fig:seg_example}
\end{figure}

 We train a segmentation model using the architecture from Mask2Former~\cite{cheng2021mask2former} with synthetic training data. Since MetaHuman textures are already semantically labeled, it is straightforward to generate synthetic training data in texture space. Moreover, since the labels are consistent from one MetaHuman to another, any manual adjustments to the labels only need to be done once. Fig.~\ref{fig:seg_example} (left) shows the labeled texture map with the neck, body, and hair all receiving the same (white) non-face label. The labeled images, Fig.~\ref{fig:seg_example} (middle), have an additional (black) label indicating the background region. In order to emphasize occlusion boundaries, the (black) background label is expanded to portions of the foreground to bound regions where the normals are becoming perpendicular to the view directions. This provides strong cues for the face silhouette. 

\begin{figure}[t]
  
  \centering
 \hspace{0.12\linewidth}
\begin{subfigure}[t]{0.31\linewidth}
  \centering
  \includegraphics[width=\linewidth]{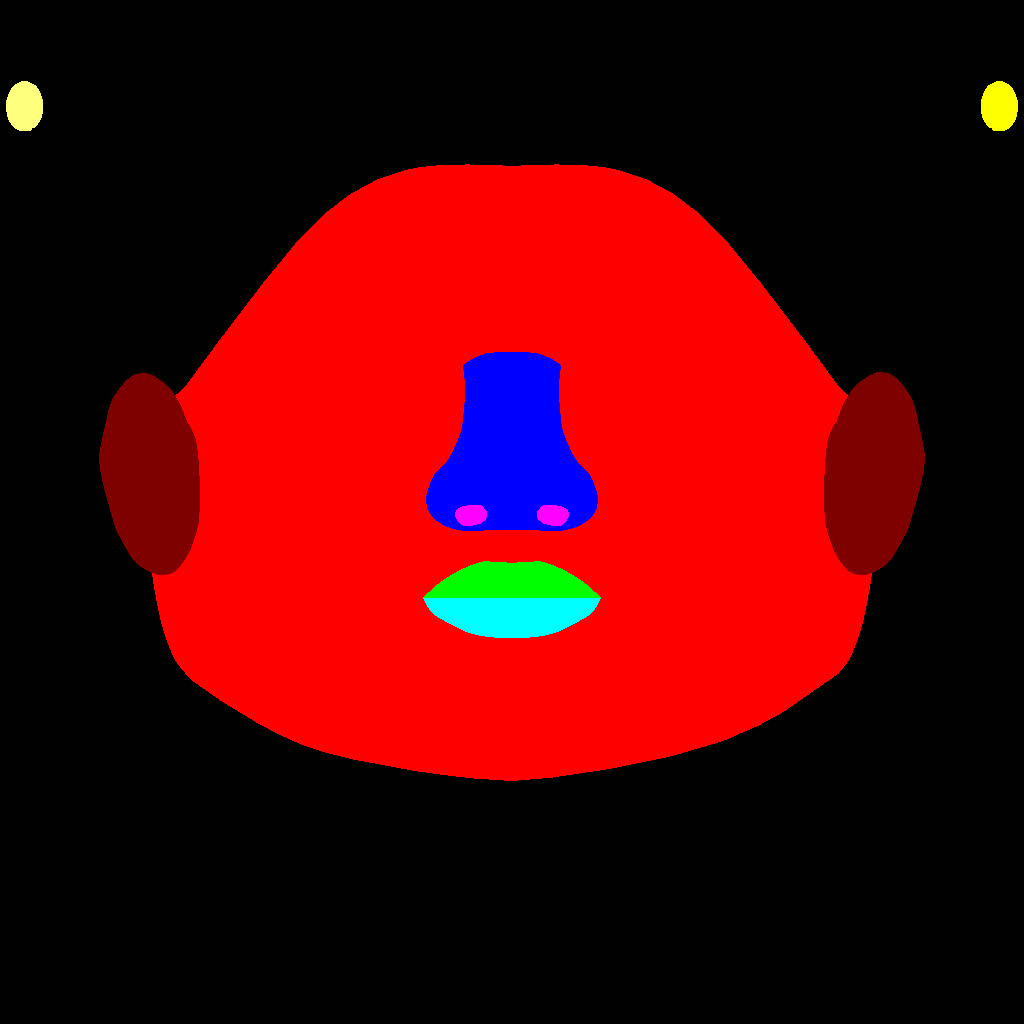}
  \label{fig:seg_map}
\end{subfigure}\hfill
\begin{subfigure}[t]{0.31\linewidth}
  \centering
  \includegraphics[width=\linewidth]{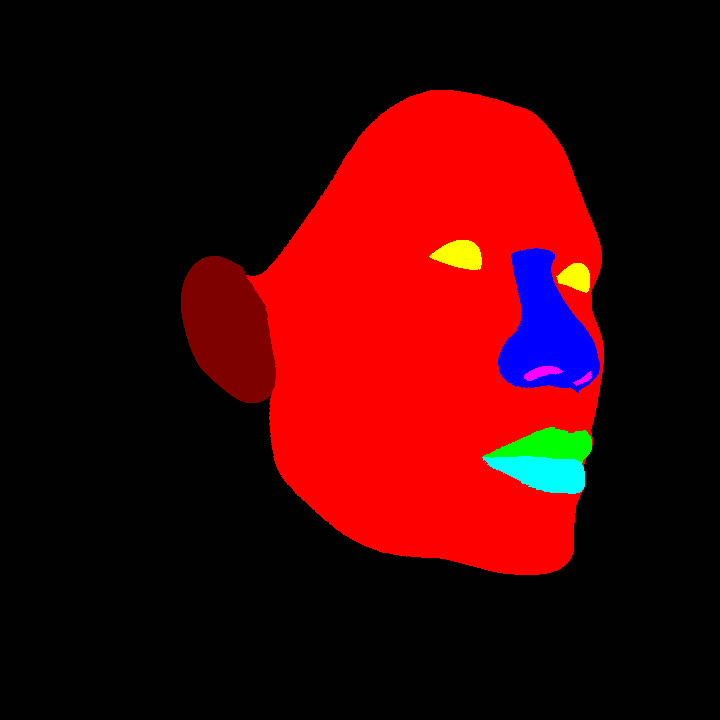}
\end{subfigure}
\hspace{0.12\linewidth}

  \caption{The template texture (left) used to assign labels to Gaussians: face (red), nose (blue), nostril (pink), top lip (green), bottom lip (cyan), eyes (yellow), ears (dark red), background (black). Note how the face label (red) has been expanded into both the hair and neck regions, as compared to Fig.~\ref{fig:seg_example}. Note the differences between the labeled triangles (right) and what one would expect to obtain as image labels from the segmentation network (Fig.~\ref{fig:seg_example} middle).}

  \label{fig:seg_gauss}
\end{figure}

Each Gaussian inherits a label from its associated triangle, and the triangle labels are derived from a template texture, Fig.~\ref{fig:seg_gauss} (left), that intentionally differs from Fig.~\ref{fig:seg_example} (left). Hairlines vary from person to person, and forcing triangles to align with an incorrect hairline can lead to spurious deformations of the forehead; thus, we expand the (red) face label in the template texture so that it includes regions that may or may not be covered with hair. It can be difficult to predict the boundary between the face and the neck, especially due to the non-linear deformations that occur in that region when someone turns their head; thus, we expand the face label in the template texture there as well. The remaining areas of the texture are given (black) background labels, in contrast to the (white) non-face foreground labels in Fig.~\ref{fig:seg_example} (left).



For each target image, the network from~\cite{meyer2025ben} is used to segment the foreground from the background; then, our segmentation network is used to predict foreground labels similar to Fig.~\ref{fig:seg_example} middle. Pixels with (white) non-face foreground labels are ignored during training, allowing Gaussians with any label to explain those regions of the image. In contrast, the (black) background labels are retained to emphasize occlusion boundaries. Let $P$ be the set of pixels in the image that are not ignored and $L$ be the number of labels (including the background) being used. Then, the segmentation loss is
\begin{equation}
\mathcal{L}_{\text{seg}} = \frac{1}{|P|} \sum_{p \in P} \left\| \hat{S}(p) - \text{onehot}_{L}(S(p)) \right\|^2
\end{equation}
where $S$ is the segmentation label of pixel $p$ and $\text{onehot}_{L}$ converts from the set of segmentation labels to the set of basis functions $\{\mathbf{e}_1, ..., \mathbf{e}_L\}$. $\hat{S}$ is constructed using all visible Gaussians that overlap a pixel by alpha blending the one-hot encoding of their segmentation labels. $\mathcal{L}_{\text{seg}}$ coerces Gaussians to explain the parts of the images that they belong to, which in turn will help to improve the alignment between the triangulated surface and the images (when the triangulated surface is later deformed, see Sec.~\ref{sec:geometry_opt}).





\subsubsection{Other Modifications}\label{sec:other_considerations}
For each target image, the mesh is rendered from the current camera view in order to identify back-facing and occluded triangles, and Gaussians associated with those triangles are ignored with a certain probability. This makes Gaussians associated with visible triangles primarily responsible for explaining the target image. Note that the Gaussians associated with hidden triangles are not entirely discarded, since they may actually correspond to a portion of the target image. This allows them to drag their hidden triangles towards the appropriate portion of the target image. 

\begin{figure}[h]
  \centering
  \setlength{\tabcolsep}{0.01\linewidth}        
  \begin{tabular}{C{0.31\linewidth} C{0.31\linewidth} C{0.31\linewidth}}
    \coltitle{Target Image} &
    \coltitle{w/o Eyes Loss} &
    \coltitle{w/ Eyes Loss}  \\[0.4em]

    \multirow{2}{*}{\includegraphics[width=\linewidth]{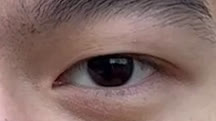}} &
    \includegraphics[width=\linewidth]{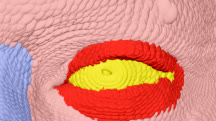} &
    \includegraphics[width=\linewidth]{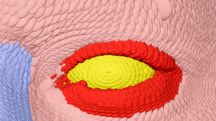} \\[0.35em]

    {} &
    \includegraphics[width=\linewidth]{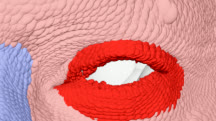} &
    \includegraphics[width=\linewidth]{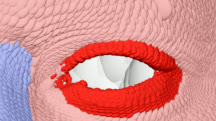} \\
  \end{tabular}

  \caption{The results obtained without and with the eye regularization loss ($\mathcal{L}_{\text{eyes}}$ in Eq.~\ref{eq:eyes_reg}), respectively. The top row shows all the Gaussians, while the bottom row omits the (yellow) eyeball Gaussians in order to more clearly illustrate the errors in the (red) eye socket Gaussians. Although both variations match the target image, the result obtained without $\mathcal{L}_{\text{eyes}}$ achieves this by allowing the eyeball Gaussians to occlude the eye socket Gaussians. The resulting mesh would have an inaccurately small eye socket. This is alleviated by $\mathcal{L}_{\text{eyes}}$.}
    
  \label{fig:eyes-reg}
\end{figure}

An additional soft constraint is introduced to keep the eyeball and eye socket Gaussians from interfering with each other. Let $x^{*}_{\text{eyeball}}$ and $x^{*}_{\text{socket}}$ be the outer boundary points of the eyeball and eye socket Gaussians, respectively. For each eye socket Gaussian, let 
\begin{equation}
\hat{x}^*_{\text{eyeball}, i} = \arg\min_j\big\lVert x^*_{\mathrm{eyeball},j} - x^*_{\mathrm{socket}, i} \big\rVert_2
\end{equation} be the closest eyeball Gaussian, and let $\mathbf{n}_{i}$ be the unit normal that points from the center of the eyeball (computed as the mean of all $x^{*}_{\text{eyeball}, j}$) to $\hat{x}^*_{\text{eyeball},i}$. Then,
\begin{multline}
\label{eq:eyes_reg}
\mathcal{L}_{\text{eyes}}
= \frac{1}{N_s} 
  \sum_{i=1}^{N_s}
  \bigl\|
    \max \bigl(\bigl(
      \hat{x}^*_{\text{eyeball,i}} -\, x^*_{\text{socket},i}
      \bigr) 
      \cdot \mathbf{n}_{i},
      0
    \bigr)
  \bigr\|_2
\end{multline}
penalizes interference between the eyeball and eye socket Gaussians. $N_s$ is the number of Gaussians representing the eye socket. See Fig.~\ref{fig:eyes-reg}. 

\subsection{Triangulated Surface Deformation}\label{sec:geometry_opt}
We use our modified Gaussian Splatting to jointly optimize the camera extrinsics and Gaussian parameters, as is typical.
 Afterwards, the camera extrinsics are held fixed during geometry refinement. Each iteration, the Gaussian parameters are re-optimized in order to provide supervisory information for subsequent deformation of the triangulated surface. The supervisory information takes the form of outer boundary points for each Gaussian, i.e.~the $\mathbf{x}_i^*$ from Sec.~\ref{sec:boundary}, except that the combined contribution (replacing the max in Sec.~\ref{sec:boundary} with a sum) of each Gaussian and its $k$-nearest neighbors is used.
That is, for each Gaussian $i$,

\begin{equation}
\sum_{j \in \mathcal{N}(i)} \alpha_j\,G_j\bigl(\boldsymbol{\mu}_i \;+\; t_i^*\,\mathbf{n}_i) \;=\; \tau
\end{equation}
is solved to find a ray parameter $t_i^*$ that defines the outer boundary point $\mathbf{x}_i^* = \boldsymbol{\mu}_i+ t_i^*\mathbf{n}_i$. 

The vertices $\mathbf{v}$ of the triangulated surface are perturbed by minimizing 

\begin{equation} \label{eq:geo_refinement}
\mathcal{L}_{\text{centroid}} = \sum_{i} \Bigl\|\,\mathbf{v}_{i}^\text{centroid} \;-\; \mathbf{x}_i^*\Bigr\|^2 
\end{equation}
where the triangle centroid $\mathbf{v}_{i}^\text{centroid}$ is the average of the triangle vertices. In addition to the data term in Eq.~\ref{eq:geo_refinement}, two regularization terms are also included in the minimization. $\mathcal{L}_{\text{reg}}^{\text{vertex}}$ aims to keep the perturbation of each vertex equal to average perturbation of its edge-connected neighbors, and $\mathcal{L}_{\text{reg}}^{\text{normal}}$ penalizes the difference between changes in the normal direction. Additional regularization can be had by optimizing over the coefficients of a PCA basis, instead of optimizing over the individual vertex positions directly. We perform two iterations of geometry refinement, using a MetaHuman per-region PCA formulation~\cite{metahuman} in the first iteration and individual vertex positions in the second. As mentioned above, the Gaussians parameters are re-optimized (using the current best guess for the geometry) before each of these iterations. 

\subsection{Neural Texture Approach}
\label{sec:neural_texture}
Although Gaussian Splatting approaches typically use a triangulated surface that only roughly approximate the geometry, we illustrate the benefits of having an accurately reconstructed triangulated surface in the section. Importantly, this accurately reconstructed triangulated surface is valuable even when one prefers the look of Gaussian Splatting to that which can be obtained via a standard graphics pipeline utilizing textured triangles. This emphasizes the efficacy of our contribution. Although neural approaches have become quite popular in the literature, they have not been widely adopted in industry. It took decades to develop the software and hardware used in the modern graphics pipeline, and the various development were shaped by the need to embrace artist creativity and control in every aspect of the pipeline. Thus, less invasive approaches are likely to facilitate quicker industry adoption. In this vein, we propose moving the Gaussian Splatting model out of world space and into texture space. This retains all the conveniences and efficiencies of the software and hardware graphics pipeline while still using Gaussian Splatting on assets of interest via a view-dependent neural texture. Of course, neural textures require more accurate geometry, again emphasizing the efficacy of our contribution.

It is straightforward to transform world space triangles to texture space, since their vertices already have UV coordinates. The resulting transforms can be used to move the Gaussians into a 3D UVW texture space in the same manner that the Gaussians are moved from canonical space to world space (see Sec.~\ref{sec:gaussian_avatars}). Note that W encodes the orthogonal distance from each Gaussian to its parent triangle. A view-dependent neural texture can then be computed by splatting the Gaussians perpendicularly with an ``orthographic camera'' while still defining their color as a function of the world space camera view as usual. Since the Gaussians live in a canonical space, it is straightforward to compute both their world space and texture space locations. Although there are an infinite number of ways to represent a view-dependent neural texture, we argue that the Gaussians provide an exceptionally good basis, as evidenced by their efficacy in providing world space structure to camera space images.

\begin{figure}[t]
  \centering


  \begin{minipage}[t]{0.225\linewidth}
  \centering
  \coltitle{World Space}\\[0.3em]
  \includegraphics[width=\linewidth]{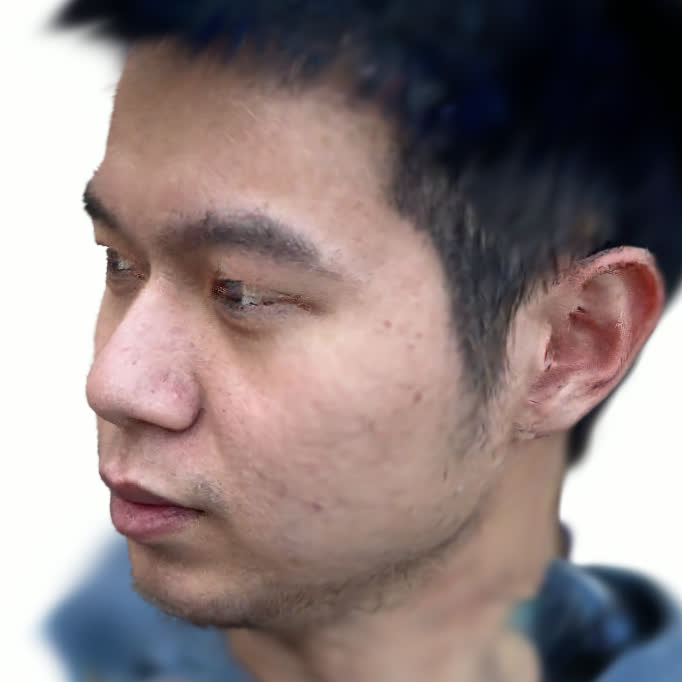}
\end{minipage}%
\hspace{0.05\linewidth}%
\hfill
\begin{minipage}[t]{0.225\linewidth}
  \centering
  \coltitle{{\vspace{1.6ex}\\Pre Fine-Tuning}}\\[0.3em]
  \includegraphics[width=\linewidth]{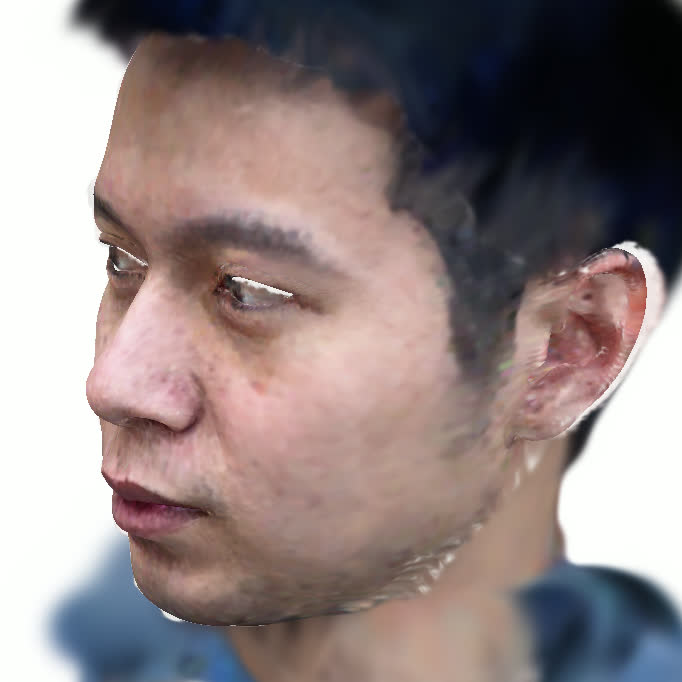}
\end{minipage}%
\hfill
\begin{minipage}[t]{0.225\linewidth}
  \centering
  \coltitle{Texture Space\\Fine-Tuned}\\[0.3em]
  \includegraphics[width=\linewidth]{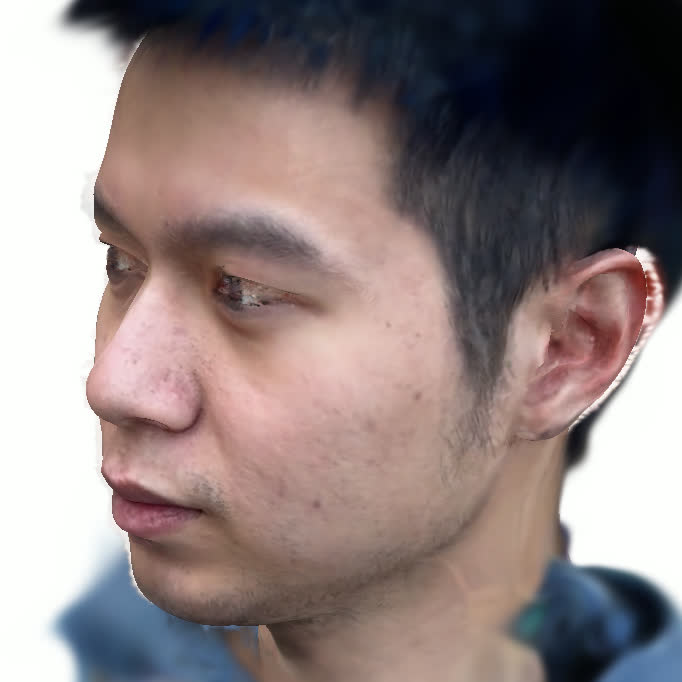}
\end{minipage}%
\hfill
\begin{minipage}[t]{0.225\linewidth}
  \centering
  \coltitle{{\vspace{1.6ex}\\w/o Occlusions}}\\[0.3em]
  \includegraphics[width=\linewidth]{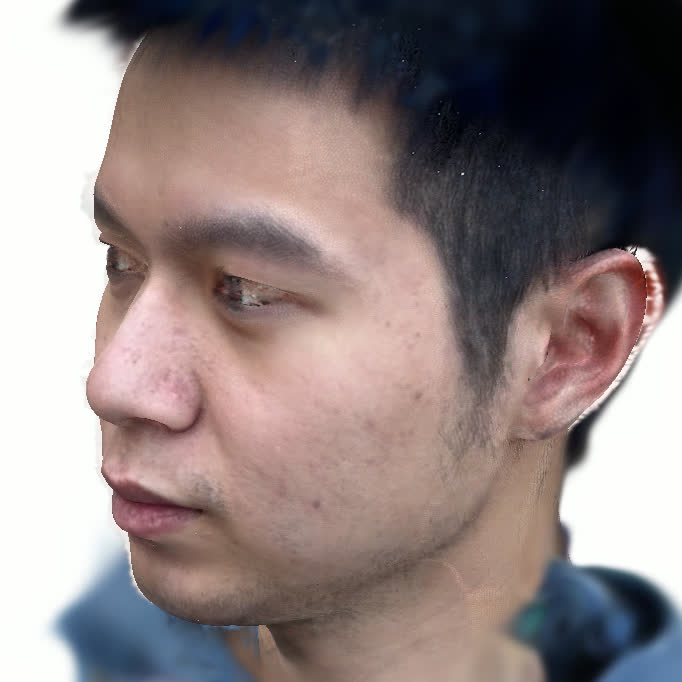}
\end{minipage}%

  \caption{A comparison of standard Gaussian Splatting (first image) to a view-dependent neural texture (other three images) for a novel view. The second image shows the results obtained by using parent triangles to transform the Gaussians into UVW texture space, and the third image shows the results obtained after fine-tuning those transformed Gaussians using the original views (this novel view is not used for the fine-tuning). The fourth image removes the $\alpha$-blending by replacing it with a direct summation, so that the Gaussians can be defined in a 2D UV texture space instead of a 3D UVW texture space for the sake of efficiency (if desired). For the sake of comparison, all the images still utilize world-space Gaussians for the non-face (hair, neck, eyeballs, etc.) regions.}
    
  \label{fig:neural_texture}
\end{figure}

\begin{figure}[t]
  \centering
  \newlength{\imgH}
  \setlength{\imgH}{0.1\linewidth}
  \begin{minipage}[t]{0.63\linewidth}
  \centering
  \coltitle{World Space}\\[0.6em]
  \begin{minipage}[t][\imgH][b]{0.4\linewidth}
    \centering
    \coltitle{(From Left)}\\[-0.3em]
    \includegraphics[height=\imgH]{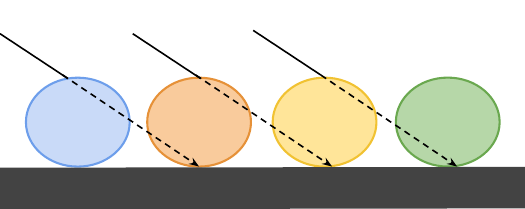}
  \end{minipage}%
  \hspace{0.03\linewidth}
  \begin{minipage}[t][\imgH][b]{0.4\linewidth}
    \centering
    \coltitle{(From Right)}\\[-0.3em]
    \includegraphics[height=\imgH]{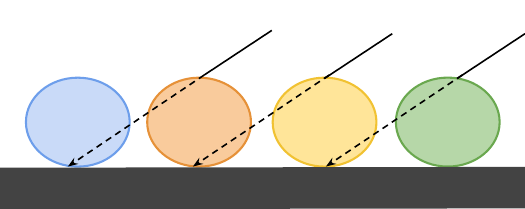}
  \end{minipage}
\end{minipage}%
\hspace{0.05\linewidth}%
\begin{minipage}[t]{0.27\linewidth}
  \centering
  \coltitle{Texture Space}\\[0.6em]
  \begin{minipage}[t][\imgH][b]{\linewidth}
    \centering
    \includegraphics[height=\imgH]{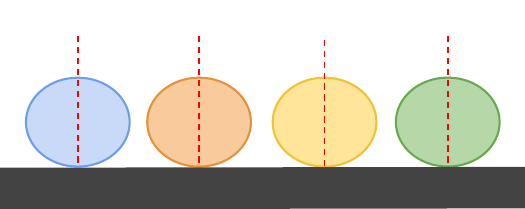}
  \end{minipage}
\end{minipage}
  \caption{In the world space approach, color accumulation is performed along the ray direction; thus, the mixing of Gaussians is view-dependent. When viewed from the left, Gaussians to the left of each point on the surface need to properly combine with the Gaussians above those points. When viewed from the right, Gaussians to the right of each point on the surface need to properly combine with the Gaussians above those points. In contrast, the texture space approach always accumulates in the same direction, orthogonal to the texture, regardless of the view direction. This helps to prevent the blurriness issue for novel views, where the Gaussians are accumulated in ways that were unanticipated during training. In addition, reducing the influence of neighboring Gaussians increases the locality and compactness of the texture space approach (as compared to the world space approach) creating a smaller circle of confusion and thus intrinsically higher resolution. }
  \label{fig:texture_space}
\end{figure}

Rendering the triangulated surface with a simple ambient-only shader allows for a straightforward comparison between standard Gaussian Splatting (Fig.~\ref{fig:neural_texture}, first image) and our newly proposed neural texture approach (Fig.~\ref{fig:neural_texture}, second image). Notably, the neural texture computes the color at each point on the triangulated surface by $\alpha$-blending Gaussians along the normal direction instead of the usual approach of $\alpha$-blending along the ray direction. See Fig.~\ref{fig:texture_space}. In addition, pixels that overlap Gaussians but do not overlap the triangulated surface are not shaded at all. Both of these differences emphasize the need for the accurate geometry provided by our method.

Since the Gaussians are being used differently, it makes sense to retrain them (using their current parameters as a warm start). After fine-tuning, the model generates even cleaner details (Fig.~\ref{fig:neural_texture}, third image) than the standard approach (Fig.~\ref{fig:neural_texture}, first image). For the sake of computation efficiency, we also experimented with replacing the usual $\alpha$-blending with a direct summation (along the lines of~\cite{zhang2024gaussianimage}). This allows for purely 2D Gaussians in UV space (Fig.~\ref{fig:neural_texture}, fourth image).


Treating the Gaussians as a neural texture has other advantages as well. For example, a mipmap can be created by reducing both the number of texels and the number of Gaussians. Notably, it is quite straightforward to accomplish this in texture space in contrast to the difficulties associated with significantly reducing the number of Gaussians for level of detail representations in world space.

Importantly, Gaussian Splatting does a good job dealing with camera extrinsics, allowing for the creation of an accurate triangulated surface geometry. However, after asset creation, it seems that Gaussian Splatting can be replaced with a simpler neural or even spherical harmonic texture with no loss of efficacy.

\subsection{Lighting Estimation}\label{sec:lighting_est}
Spherical harmonics~\cite{ramamoorthi2001efficient} is used to approximate the lighting conditions via $\sum_k \boldsymbol{l}_k  A_k \, Y_k(\boldsymbol{n}_p)$, where $\boldsymbol{n}_p$ is a per-pixel unit-length surface normal, \( Y_k \) are diffuse spherical harmonic basis functions, $A_k$ are normalization constants, and the $\boldsymbol{l}_k$ are learned three-color-channel coefficients. In order to encourage uniform values across color channels to prevent color overfitting, the zeroth order coefficients are regularized via $\mathcal{L}_{\text{lighting}} = \left\| \boldsymbol{l}_0 - \bar{\boldsymbol{l}}_0 \right\|_2$, where $\bar{\boldsymbol{l}}_0$ is the mean value across the three color channels. Each \(\boldsymbol{n}_p\) is computed by perturbing the interpolated vertex averaged normal by a learned per-pixel rotation $R_p$ meant to mimic a normal map. Each Gaussian is given an additional learnable parameter $R_i$, and the $R_i$ are alpha-blended (as usual) in order to determine per-pixel $R_p$. Regularization of the form $\mathcal{L}_{\text{rotation}} = \left\| R_p - I \right\|^2_F$ is added to limit the magnitude of the rotation. This normal map reduces color artifacts around fine-scale features such as wrinkles.

\begin{figure}[t]
  \centering

  \hfill
  \hspace{0.12\linewidth}
  \begin{minipage}[t]{0.31\linewidth}
    \centering
    \coltitle{w/o Occlusion map}\\[0.3em]
    \includegraphics[width=\linewidth]{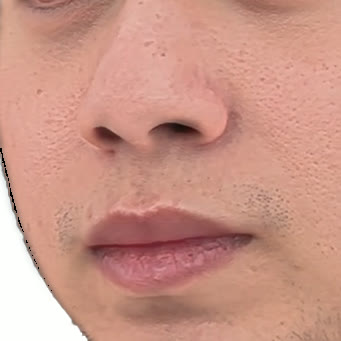}
  \end{minipage}%
  \hfill
  \begin{minipage}[t]{0.31\linewidth}
    \centering
    \coltitle{w/ Occlusion map}\\[0.3em]
    \includegraphics[width=\linewidth]{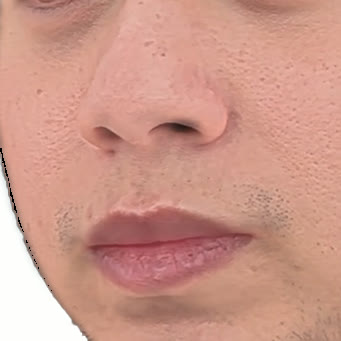}
  \end{minipage}%
  \hspace{0.12\linewidth}%
  \hfill
  \caption{De-lit images recovered without and with the occlusion map, respectively. The occlusion map reduces baked-in ambient shadows, particularly in recessed regions such as below the nose and along the lip crease, improving the de-lit textures.}
    
  \label{fig:occlusion-map}
\end{figure}

To improve rendering quality near occlusion boundaries and reduce baked-in shadows, we precompute a screen-space occlusion map for each target image, similar to~\cite{herholz2011screen}. For each pixel, the corresponding point $\boldsymbol{x}_p$ on the triangulated surface is identified; then, Monte Carlo integration over the hemisphere centered about $(\boldsymbol{x}_p, \boldsymbol{n}_p)$ is used to compute a per-channel visibility fraction
\begin{equation}
f_{\text{visible}}(p; l) = \frac{\sum_k \l_k \sum_{j}  V(\boldsymbol{x}_p, \boldsymbol{d}_j)  Y_k(\boldsymbol{d}_j) \boldsymbol{n}_p \cdot \boldsymbol{d}_j }{\sum_k l_k  \sum_{j} Y_k(\boldsymbol{d}_j) \boldsymbol{n}_p \cdot \boldsymbol{d}_j }
\end{equation}
where the visibility in direction $\boldsymbol{d_j}$, i.e. $V(\boldsymbol{x}_p, \boldsymbol{d_j}) \in \{0, 1\}$, is computed based on the mesh geometry. Since $f_{\text{visible}}$ is expensive to evaluate, it is only computed occasionally using the current best guess for $\boldsymbol{n}_p$ and the $\boldsymbol{l}_k$.  Denoting the lagged $\boldsymbol{l}_k$ as $\boldsymbol{l}_\text{visible}$, the per-pixel per-channel lighting estimate is given by 
\begin{equation}
\label{eq:lighting_est}
L(p; l, l_\text{visible}) = f_{\text{visible}}(p; l_\text{visible})\sum_k l_k \, A_k Y_k(\boldsymbol{n}_p)
\end{equation}
 where $\boldsymbol{n}_p$ depends on both the triangulated surface and the learned normal map. See Fig.~\ref{fig:occlusion-map}.

\subsection{De-lit Texture Generation}\label{sec:texture_opt}
\begin{figure*}[t]
  \centering
  \includegraphics[width=0.7\linewidth]{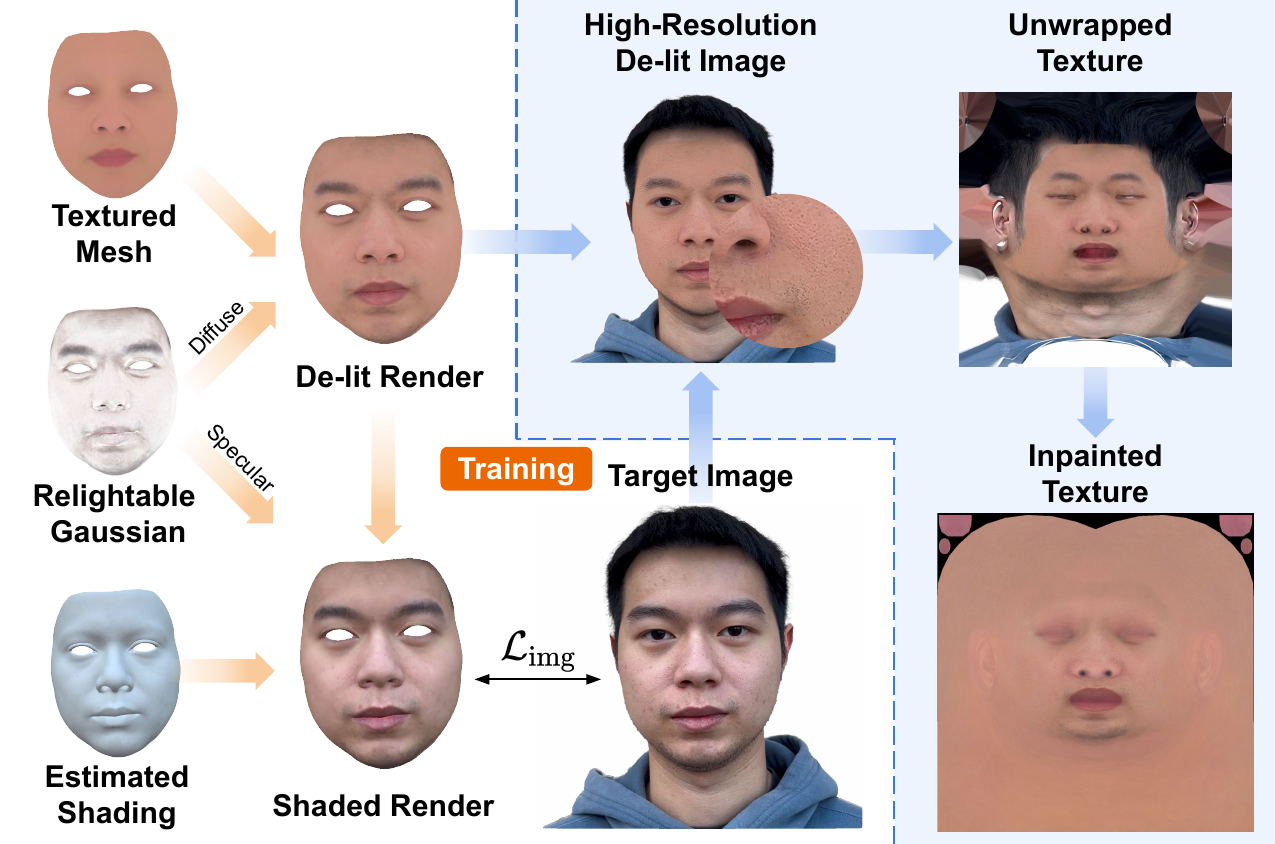}
  \caption{Our method combines a textured mesh with a relightable Gaussian model and estimated spherical-harmonics lighting to reproduce a target image. After training, we render the view-independent albedo component to obtain a de-lit image and reintroduce high-frequency details extracted from the reference image. The resulting high-resolution de-lit images are then unwrapped and inpainted to form a complete high-resolution de-lit texture.}
  \label{fig:tex_pipeline}
\end{figure*}

Although our method is motivated by~\cite{saito2024relightable}, their method relies on a large light stage dataset and thus cannot be applied directly to our limited number of images taken by a single camera without known or constrained lighting. To address this, we regularize the problem with a rendering of the reconstructed triangulated surface using the low-dimensional texture space available in the MetaHuman framework; in particular, we use the first 20 (out of a total of 137) PCA coefficients. This leads to a texture $\sum_j c_j \boldsymbol{T}_j$ where the $\boldsymbol{T}_j$ are three-channel PCA basis functions and the $c_j$ are learned during training. Augmenting this regularized texture with Gaussian Splatting allows the model to match the training data. The goal is to limit the contribution of the Gaussians to high-frequency texture details and lighting variations in order to obtain a de-lit mesh-based texture. See Fig.~\ref{fig:tex_pipeline} for an overview of our approach.


For each pixel, the per-pixel lighting estimate (see Sec.~\ref{sec:lighting_est}) is multiplied channel-by-channel by the triangulated surface texture interpolated to that pixel. A per-pixel compositing weight $\beta_p$ is used to combine the contribution from the Gaussians with a $1 - \beta_p$ contribution from the triangulated surface. Each Gaussian is given an additional learnable parameter $\beta_i$, and the $\beta_i$ are alpha-blended (as usual) in order to determine per-pixel $\beta_p$. During training, the $\beta_i$ are regularized toward zero via $\mathcal{L}_{\text{blending}} = \left\| \beta_p \right\|_2$ in order to favor reliance on the mesh.


For the Gaussian model, we modify the method proposed in Sec.~\ref{sec:relightable} in a number of ways in order to provide regularization aimed to compensate for our much more limited access to data. The diffuse term in Eq.~\ref{eq:relightable_diffuse} is calculated by alpha compositing the learned albedo $\boldsymbol{\rho}$ along the lines of Eq.~\ref{eq:comp} before multiplying it by the per-pixel lighting estimate (from Eq.~\ref{eq:lighting_est}). The specular term in Eq.~\ref{eq:relightable_specular} and simplifications thereof seem intractable in the face of limited data; thus, 
we replace Eq.~\ref{eq:relightable_specular} with an extra view-dependent color term $\boldsymbol{c}$, identical to that which was discussed in and after Eq.~\ref{eq:comp}. That is, motivated by~\cite{saito2024relightable}, we augment~\cite{kerbl20233d} to include an extra view-independent diffuse term evaluated with the aid of a spherical harmonic lighting estimate. The intent is to have this diffuse term explain as much of the lighting as possible, so that the original view-dependent color term can be minimized via $\mathcal{L}_{\text{view}} = \left\| \boldsymbol{c} \right\|_2$.

After training, de-lit images can be obtained by ignoring the view-dependent color term and by setting the per-pixel lighting estimate equal to a value of one. Lost or damaged high-frequency texture details can be restored by extracting them from the target images (via a high-pass filter) and using them to replace the equivalent high-frequency content in the de-lit images. Note that the target images are warped to better align with the triangulated surface geometry (see~\cite{zhu2024democratizingcreationanimatablefacial}) before the high-frequency content is extracted. The method proposed in~\cite{lin2022leveraging} is used to project the corrected de-lit image data into texture space where it can be gathered to texels in order to obtain a de-lit texture.

Although using~\cite{lin2022leveraging} to project and gather a de-lit texture results in a high-quality result for most of the face, there are some problematic areas. The eyebrows and other hair regions will have artifacts baked into the texture, the neck region suffers from limited visibility, the hairline on the scalp will be erratic, etc. We remedy these issues by creating a mask in texture space that separates the high-quality results from the regions that are typically problematic. The texture in the high-quality region is projected into the MetaHuman texture space to find PCA coefficients for a global texture that can be used to inpaint the problematic regions; afterwards, blending is used to soften the texture transition at region boundaries.

\subsection{MetaHuman Generation}\label{sec:mh_generation}
\begin{figure}[t]
  \centering

  \hfill
  \hspace{0.12\linewidth}
  \begin{minipage}[t]{0.31\linewidth}
    \centering
    \coltitle{Target Image}\\[0.3em]
    \includegraphics[width=\linewidth]{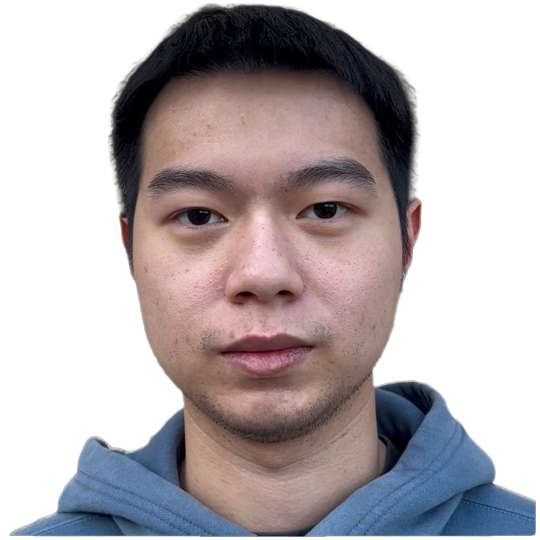}
  \end{minipage}%
  \hfill
  \begin{minipage}[t]{0.31\linewidth}
    \centering
    \coltitle{MetaHuman}\\[0.3em]
    \includegraphics[width=\linewidth]{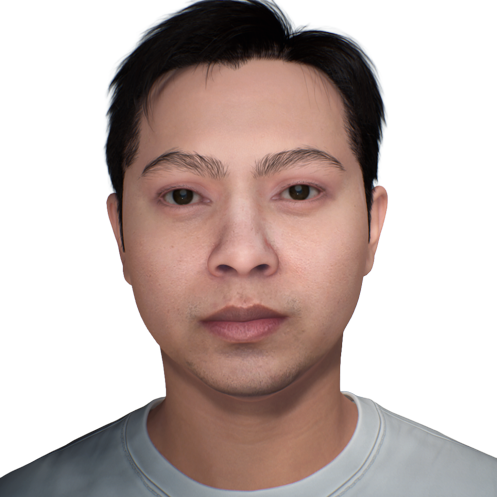}
  \end{minipage}%
  \hspace{0.12\linewidth}%
  \hfill
  \caption{Target image (left) and the MetaHuman reconstructed with our pipeline (right). Note that the hair style, eyebrow style, and eyeballs are manually selected from the MetaHuman database.}
    
  \label{fig:metahuman}
\end{figure}

Although we started with an initial guess that was consistent with the MetaHuman framework (see Section~\ref{sec:data_init}), the geometry perturbations discussed in Section~\ref{sec:geometry_opt} lead to a vertex layout (sometimes referred to as spans in the industry) that is inconsistent with the underlying Metahuman basis. Thus, we use the mesh-to-MetaHuman~\cite{epic_mesh_to_metahuman} tool to convert our reconstructed geometry back into the MetaHuman basis. Since this process perturbs vertex positions, the PCA basis function used for the texture no longer correspond to the same surface locations. We remedy this by repeating the lighting estimate and texture solve (discussed in Section~\ref{sec:lighting_est} and Section~\ref{sec:texture_opt}) on the new geometry. Obviously, it would be more efficient if only one lighting and texture solve were required; however, the mesh-to-Metahuman process requires an input texture (and better textures give better results). Although the teeth, inner mouth, and eyelashes are added back to the geometry via the mesh-to-MetaHuman process, we ignore them (as usual) during the lighting estimate and texture solve. Alternatively, MetaHuman Creator~\cite{epic2025metahuman_creator} in Unreal Engine 5.6, can be used to better preserve the input geometry; in addition, this alleviates the need to repeat the lighting and texture solve. See Fig.~\ref{fig:metahuman}

\section{Implementation Details}

Follow~\cite{zhao2016loss}, we combine a standard per-pixel $\mathcal{L}_1$ loss with a structural similarity loss (SSIM from~\cite{wang2004image}), i.e. 
\begin{equation}
\mathcal{L}_{\text{img}} = (1 - \lambda) \mathcal{L}_1 + \lambda \mathcal{L}_{\text{D-SSIM}}
\end{equation}
where $\mathcal{L}_{\text{D-SSIM}}$ multiplies the structural similarity by negative one so that it is maximized, and $\lambda = 0.2$. This approach is also followed by~\cite{qian2024gaussianavatars,kerbl20233d}. $\mathcal{L}_{\text{img}}$ is used in every stage of the Gaussian training. In both the camera and geometry optimization passes, the regularization terms discussed in Sec.~\ref{sec:soft_reg} are included; in addition, $\lambda_{\text{scale}} = 50$ for $\mathcal{L}_{\text{scale}}$ from Sec.~\ref{sec:gaussian_avatars}. In the camera optimization pass, $\lambda_{\text{reg}}^{\text{center}} = 10$, $\lambda_{\text{reg}}^{\text{normal}} = 10$, and $\lambda_{\text{reg}}^{\text{boundary}} = 50$. In the geometry optimization pass, $\lambda_{\text{reg}}^{\text{center}} = 20$, $\lambda_{\text{reg}}^{\text{normal}} = 10$, $\lambda_{\text{reg}}^{\text{boundary}} = 500$. In both passes, $\lambda_{\text{seg}} = 50$ for $\mathcal{L}_{\text{seg}}$ from Sec.~\ref{sec:seg_sup}, and $\lambda_{\text{eyes}} = 20$ for $\mathcal{L}_{\text{eyes}}$ from Sec.~\ref{sec:other_considerations}. During texture reconstruction, the regularization terms from Sec.~\ref{sec:lighting_est} and Sec.~\ref{sec:texture_opt} use $\lambda^{\text{lighting}} = 10^{-3}$, $\lambda^{\text{rotation}} = 0.2$, $\lambda^{\text{blending}} = 0.1$, and $\lambda^{\text{view}} = 10^{-3}$. 

The model is trained for 5000 iterations with one Gaussian per triangle in both the camera and geometry optimization passes. Before texture reconstruction, the mesh is subdivided (without perturbing vertex positions) and each Gaussian is split into four Gaussians. The new Gaussians are located at the centers of the sub-triangles and otherwise inherit the properties of their parent Gaussian, except for a reduction in scale by a factor of two. An additional $1000$ iterations are used to refine the densified model. During texture reconstruction, opacity, position, rotation, and scale are kept fixed, and the learnable parameters are initialized as follows: albedo and the view-dependent spherical harmonic coefficients are set randomly, compositing weights are set to $0.05$, and normal map rotations start at the identity. The texture reconstruction model is trained for $2000$ iterations.

We use a learning rate of $5\times10^{-3}$ for all Gaussian parameters, except for scale and compositing weights which use $1\times10^{-3}$ and $2\times10^{-2}$ respectively. Slowing the convergence of scale allows the model to better benefit from the segmentation supervision. In the camera optimization pass, the learning rate for camera extrinsics is set to $1 \times 10^{-2}$. During texture reconstruction, the learning rate for the lighting estimates is set to $5 \times 10^{-4}$ and the learning rate for the mesh-based texture PCA coefficients is set to $1 \times 10^{-2}$. We followed the training protocol of~\cite{qian2024gaussianavatars} for all other implementation details not explicitly specified. 

When perturbing the vertices of the triangulated surface as discussed in Sec.~\ref{sec:geometry_opt}, we optimize the parameters (either PCA coefficients or vertex positions) for $2000$ iterations using Adam~\cite{kingma2014adam} with a learning rate of $5\times 10^{-3}$. $\mathcal{L}_{\text{centroid}}$ is combined with the regularization terms via $\lambda_{\text{reg}}^{\text{vertex}} = 2$ and $\lambda_{\text{reg}}^{\text{normal}} = 0.2$. In the PCA optimization, we found it useful to penalize coefficients that exceeded a value of $1$ via $\lambda_{\text{reg}}^{\text{PCA}} = 1$. 

\subsection{Segmentation Model Training} 

The segmentation network, described in Sec.~\ref{sec:seg_sup}, was trained on 1600 randomly sampled identities from the MetaHuman Creator~\cite{epic2025metahuman_creator}. Five $720 \times 720$ resolution images were used for each identity, yielding 8000 total image pairs (Fig.~\ref{fig:seg_example}, middle and right). Random eye controls, including blinking, were included in the otherwise neutral facial expressions. Lighting conditions were randomly sampled from the 11 available environment presets. The $25^\circ$ FoV camera always pointed at the center of the head, and its distance from the center of the head was varied randomly from $45$ to $80$ cm. Measured from the horizontal ray through the center of the head that extends through the front middle of the face, the pitch and yaw were randomly sampled to be between $\pm 40^\circ$ and $\pm 80^\circ$, respectively. We trained the model for 40,000 iterations with a batch size of 4, and all other training details follow the default settings of~\cite{cheng2021mask2former,mmseg2020}. An additional 60 randomly sampled identities, yielding 300 total image pairs, were used for validation.

\section{Experiments}
\begin{figure*}[!t]
  \centering
  
  \begin{minipage}[t]{0.13\linewidth}
    \centering
    \coltitlelarge{Target Image}\\[0.3em]
    \includegraphics[width=\linewidth]{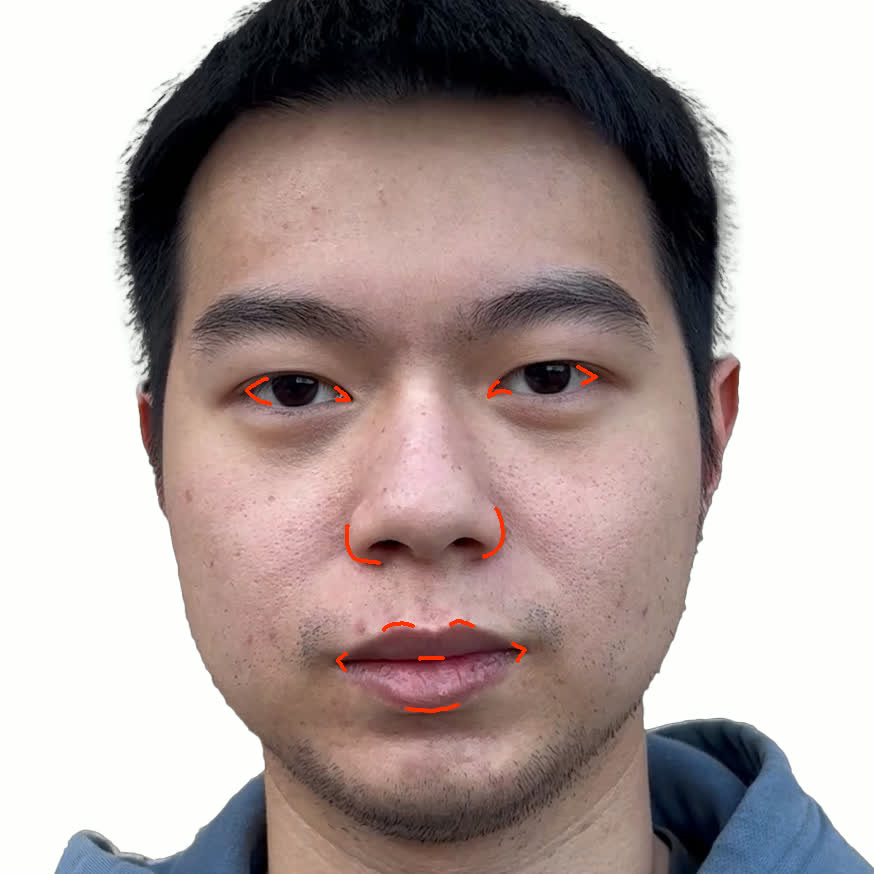}\\
    \includegraphics[width=\linewidth]{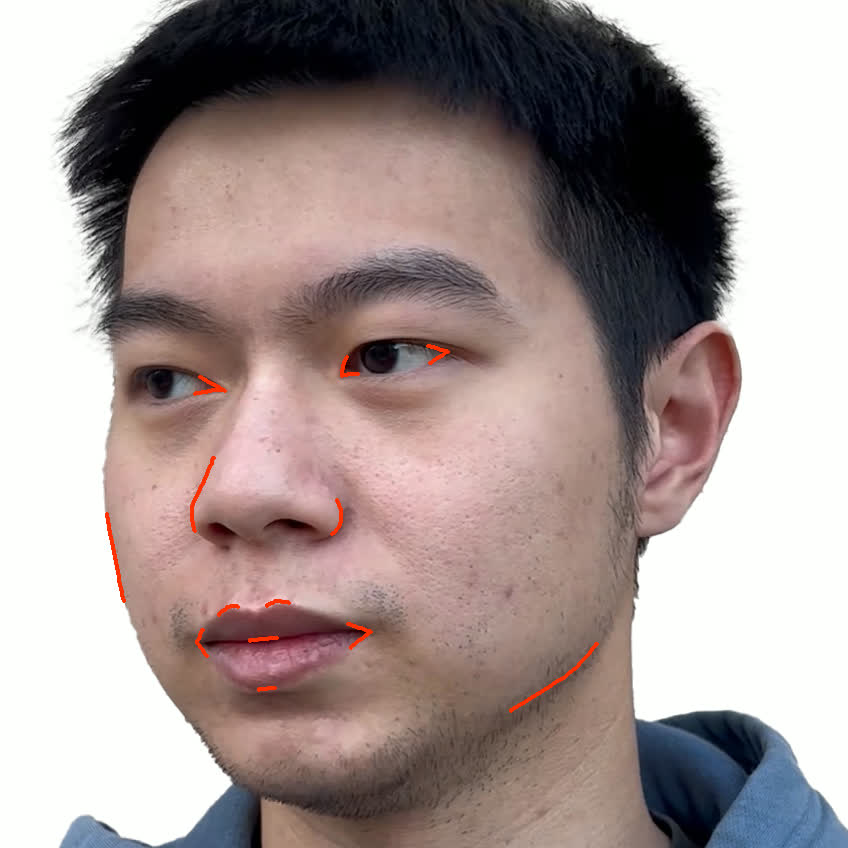}\\
    \includegraphics[width=\linewidth]{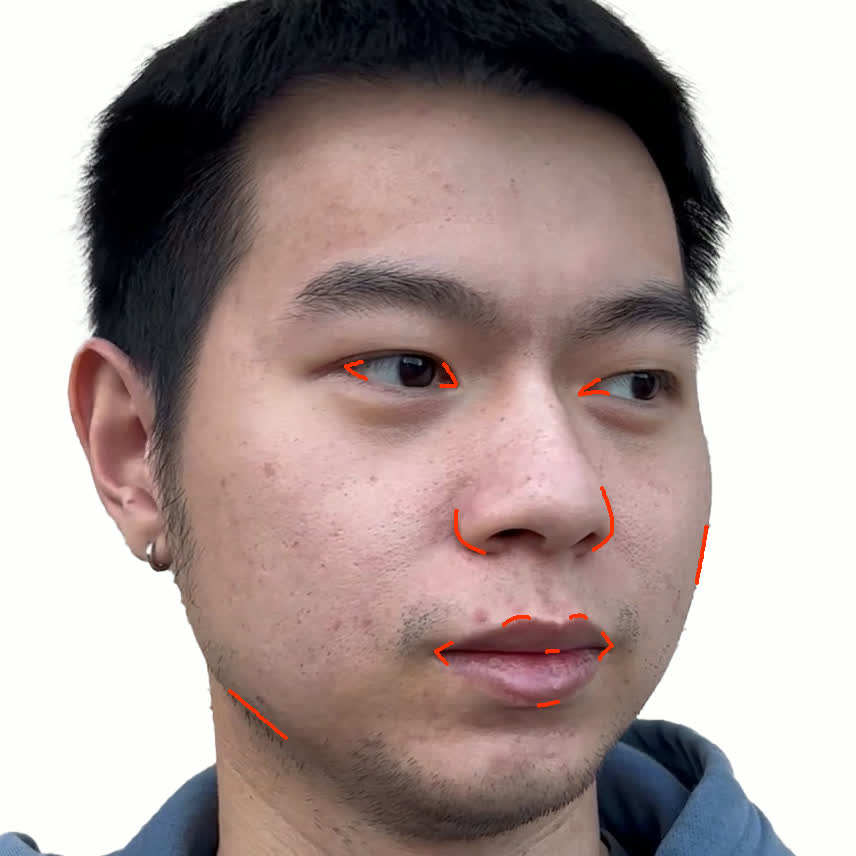}
  \end{minipage}%
  \hfill%
  \begin{minipage}[t]{0.26\linewidth}
    \centering
    \coltitlelarge{Ours}\\[0.3em]
    \begin{minipage}[t]{0.5\linewidth}
      \centering
      \includegraphics[width=\linewidth]{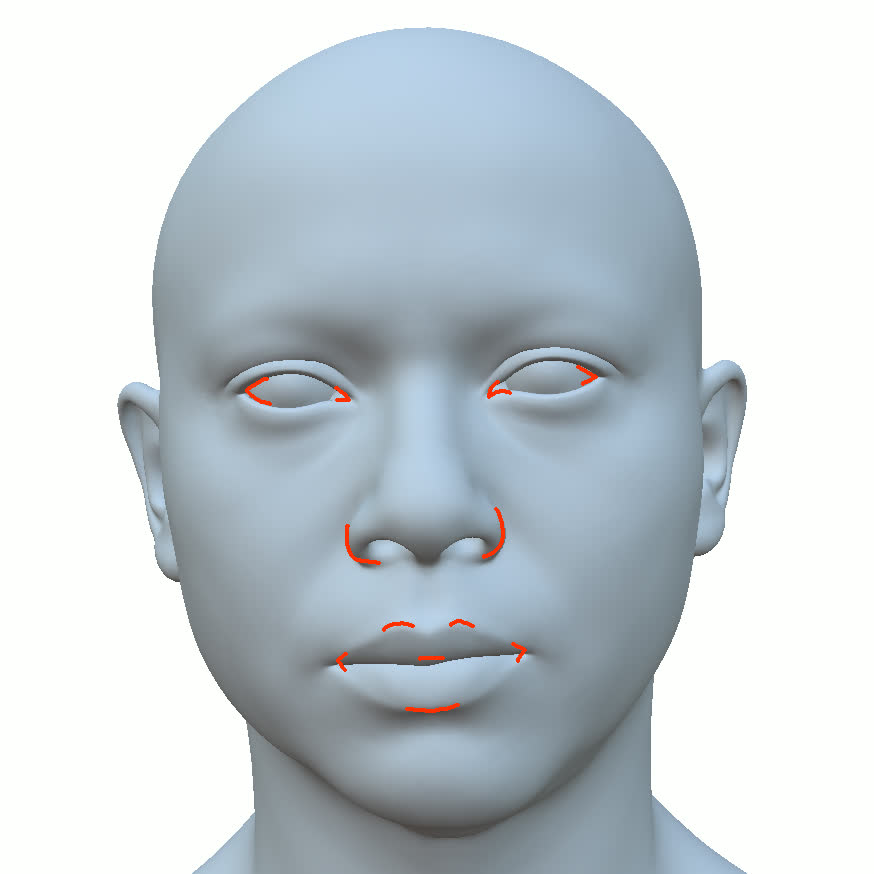}\\
      \includegraphics[width=\linewidth]{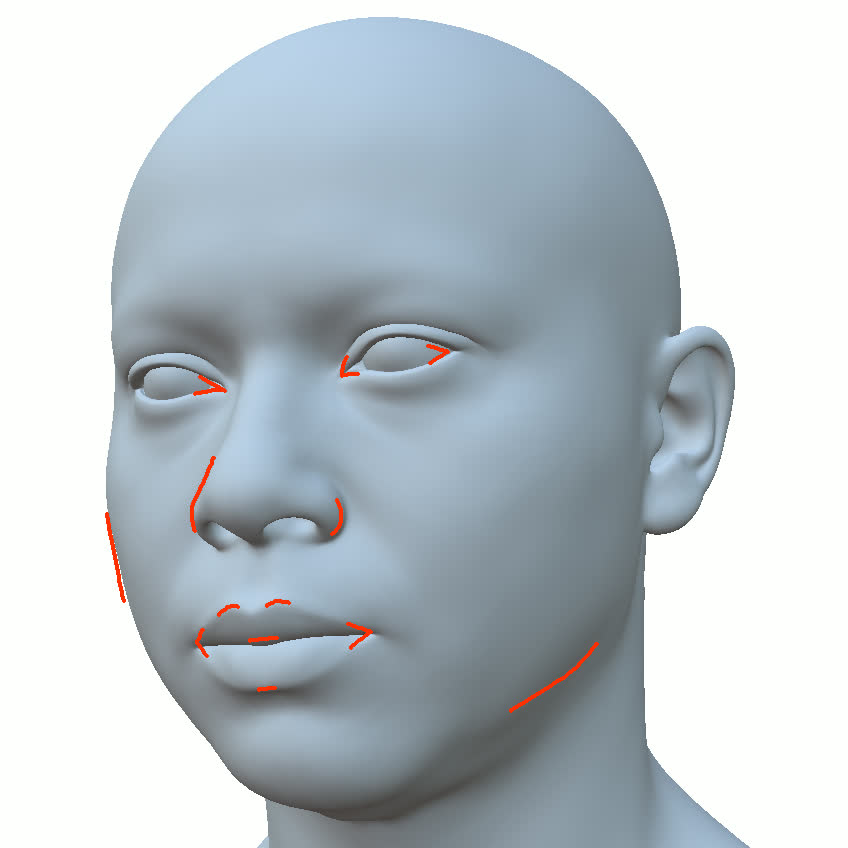}\\
      \includegraphics[width=\linewidth]{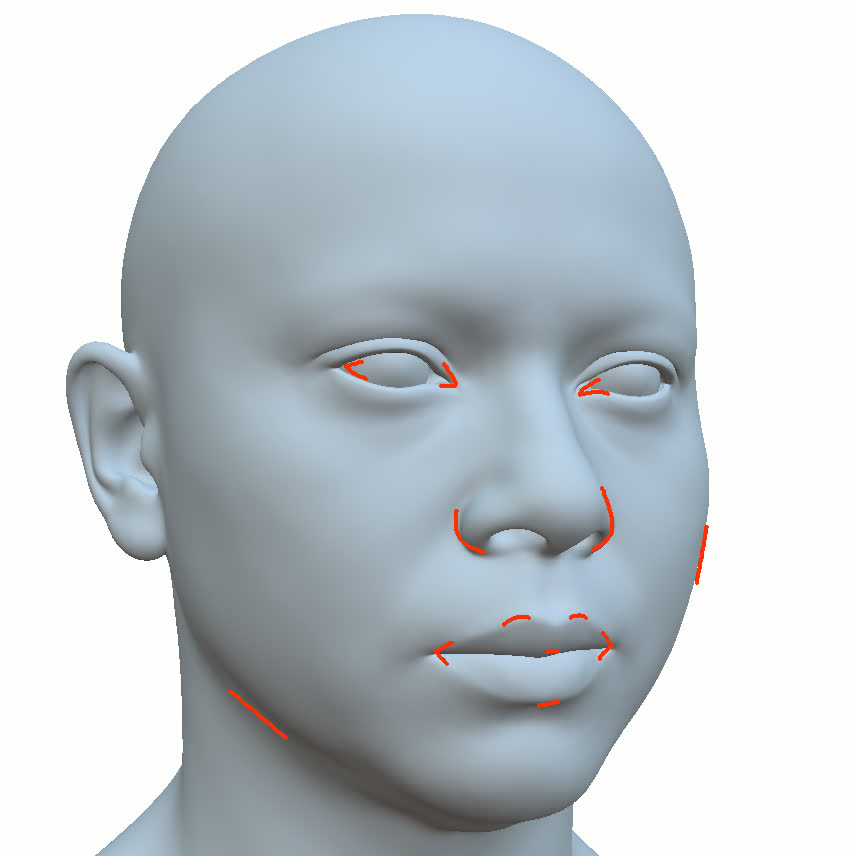}
    \end{minipage}%
    \begin{minipage}[t]{0.5\linewidth}
      \centering
      \includegraphics[width=\linewidth]{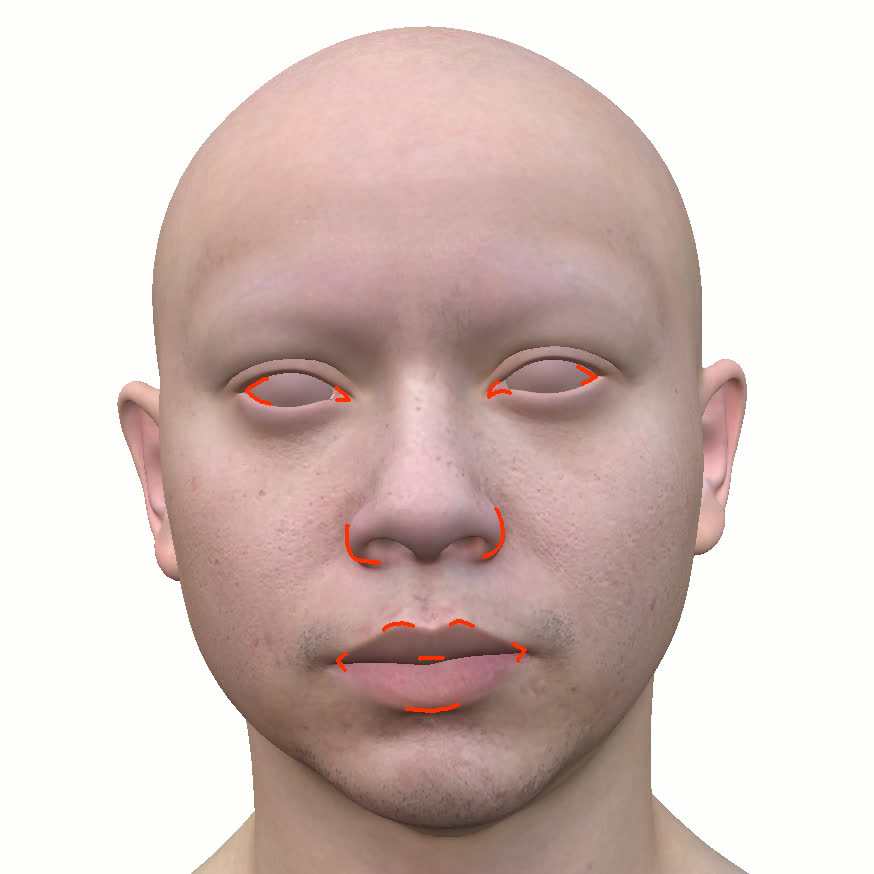}\\
      \includegraphics[width=\linewidth]{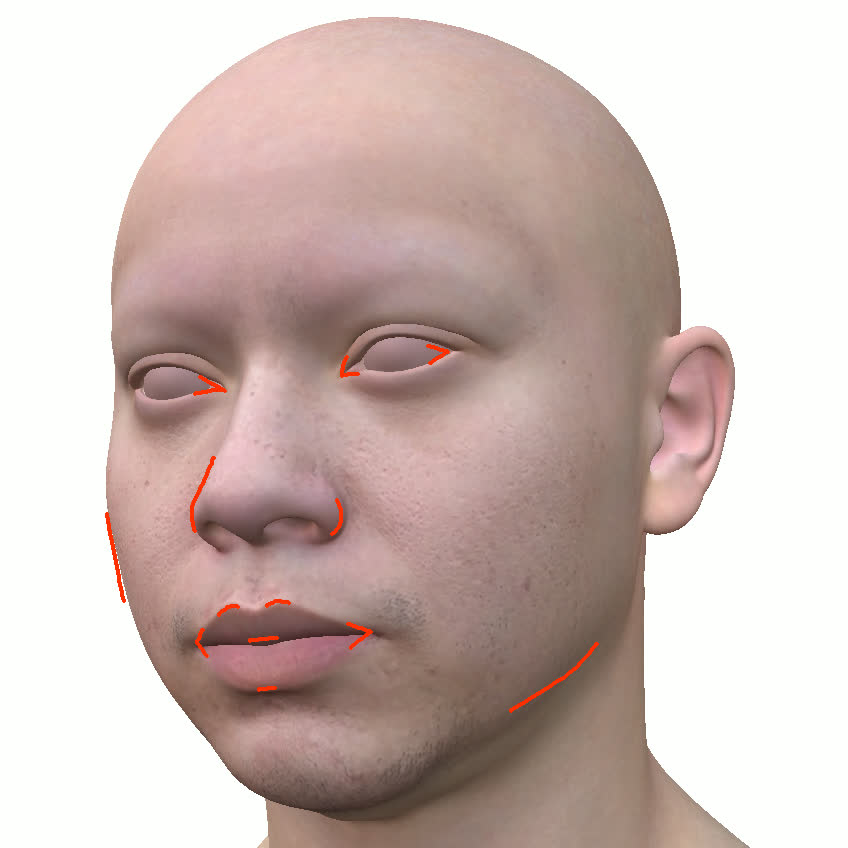}\\
      \includegraphics[width=\linewidth]{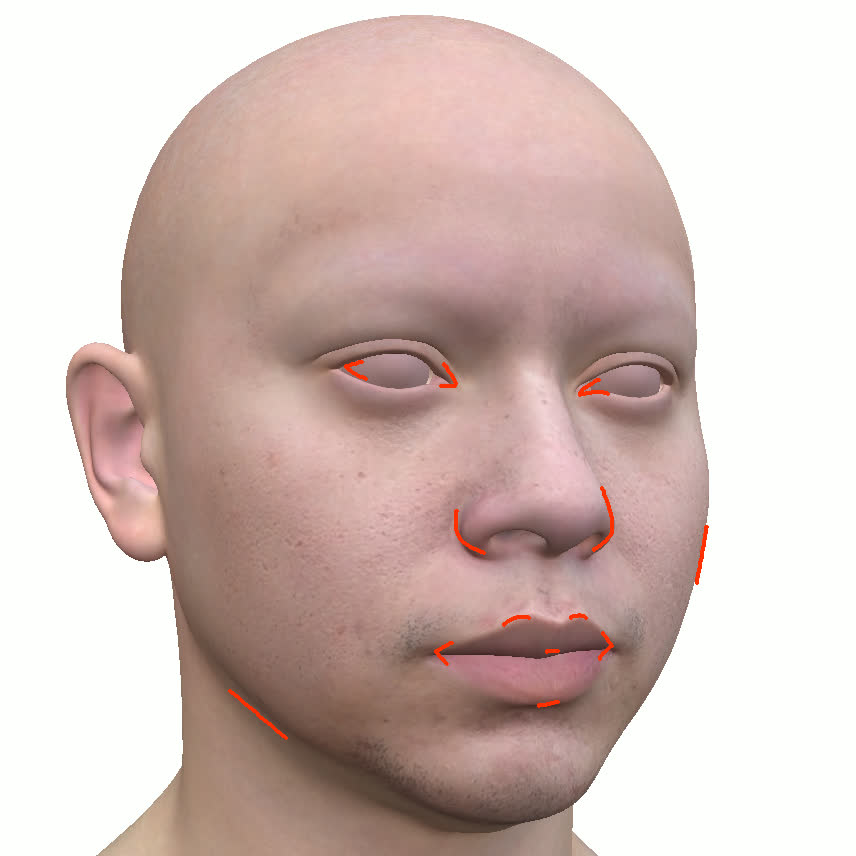}
    \end{minipage}
  \end{minipage}%
  \hfill%
  \begin{minipage}[t]{0.26\linewidth}
    \centering
    \coltitlelarge{NextFace}\\[0.3em]
    \begin{minipage}[t]{0.5\linewidth}
      \centering
      \includegraphics[width=\linewidth]{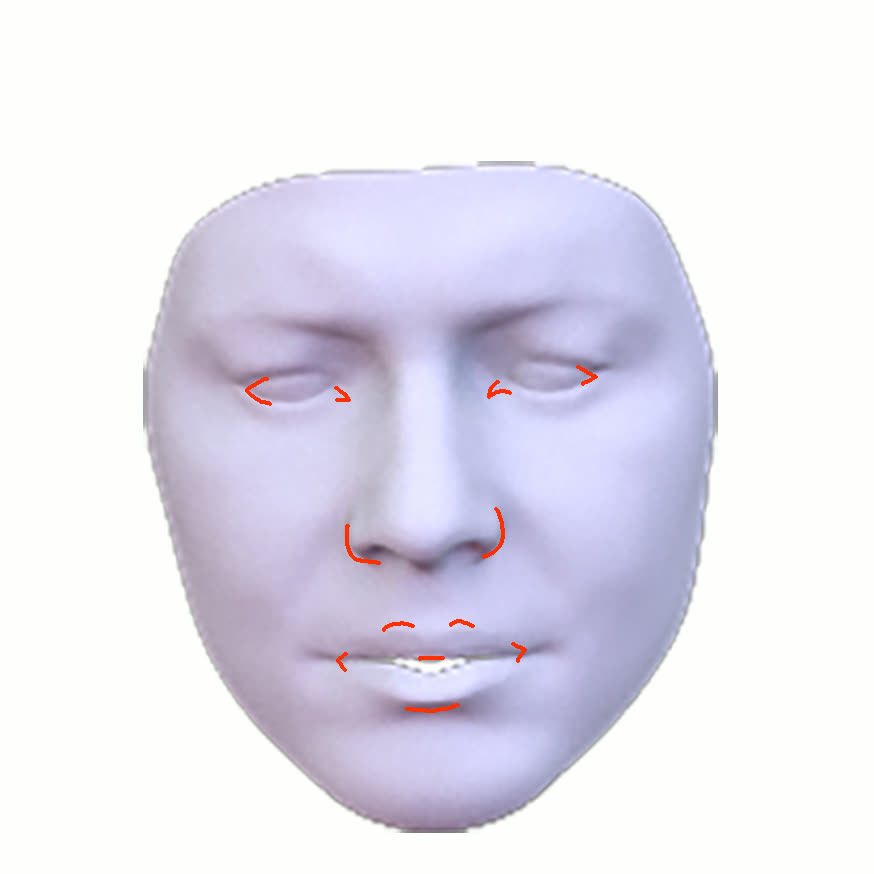}\\
      \includegraphics[width=\linewidth]{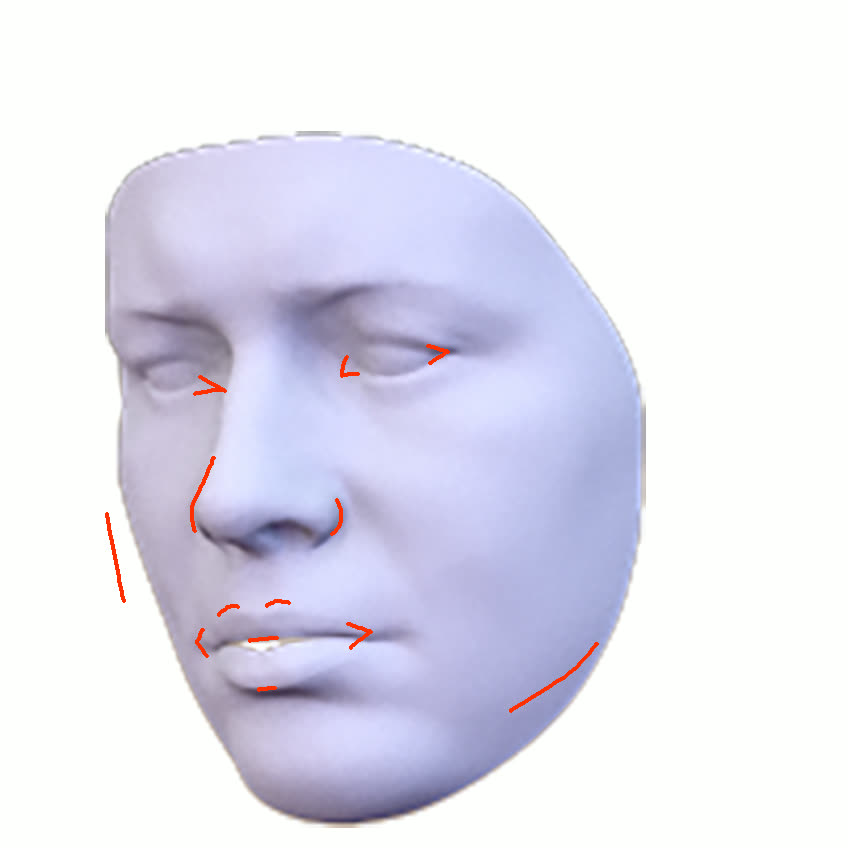}\\
      \includegraphics[width=\linewidth]{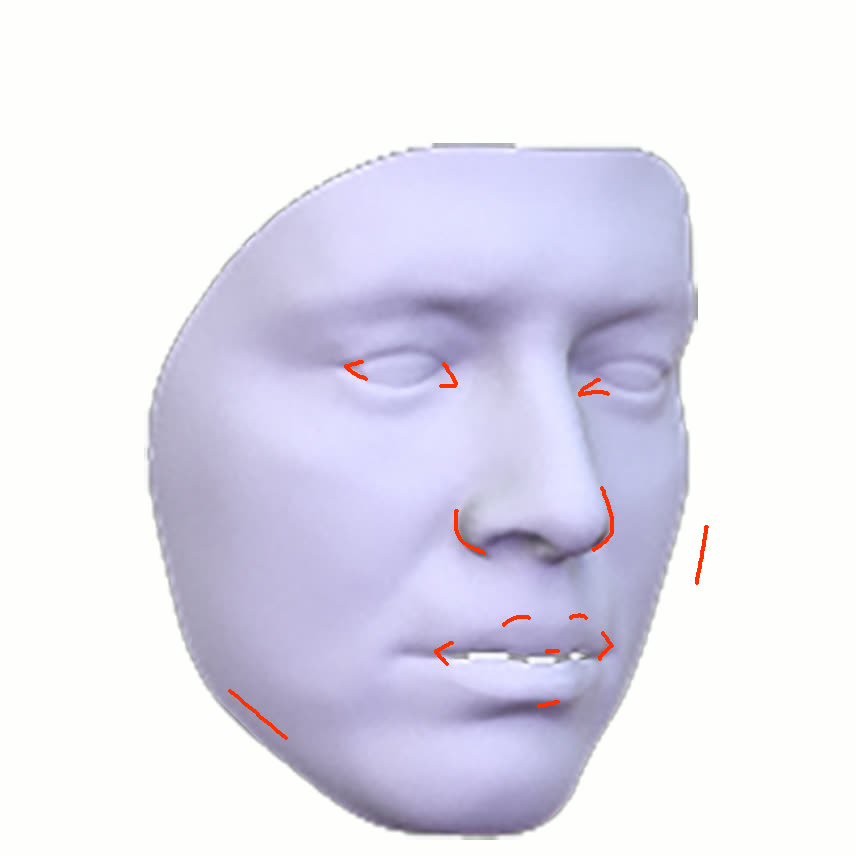}
    \end{minipage}%
    \begin{minipage}[t]{0.5\linewidth}
      \centering
      \includegraphics[width=\linewidth]{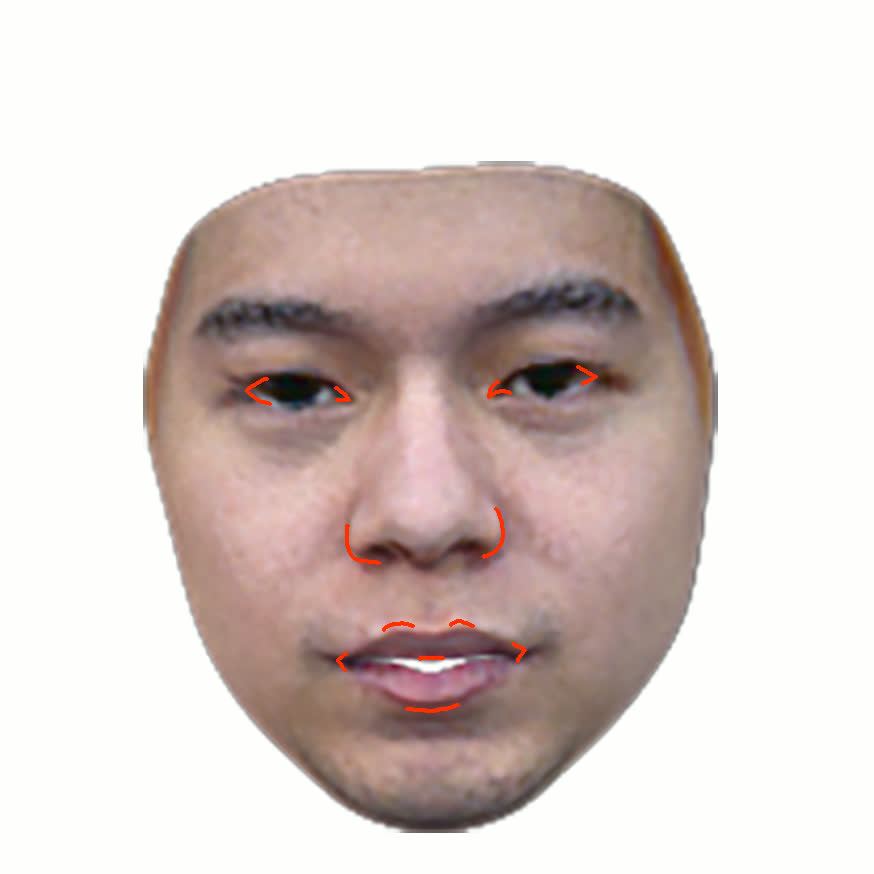}\\
      \includegraphics[width=\linewidth]{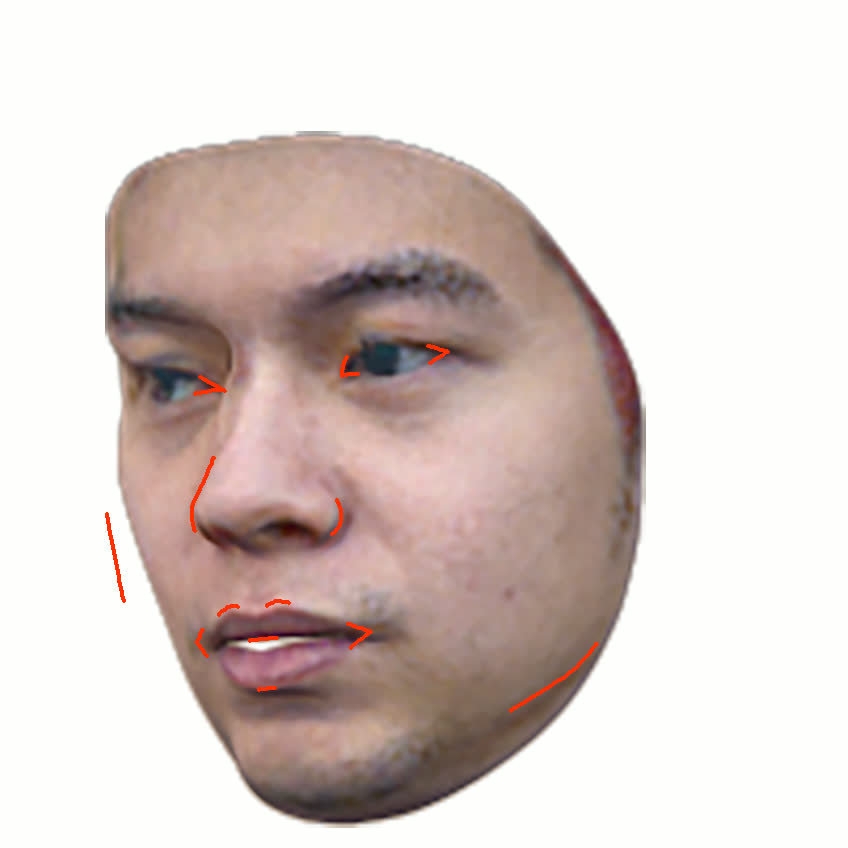}\\
      \includegraphics[width=\linewidth]{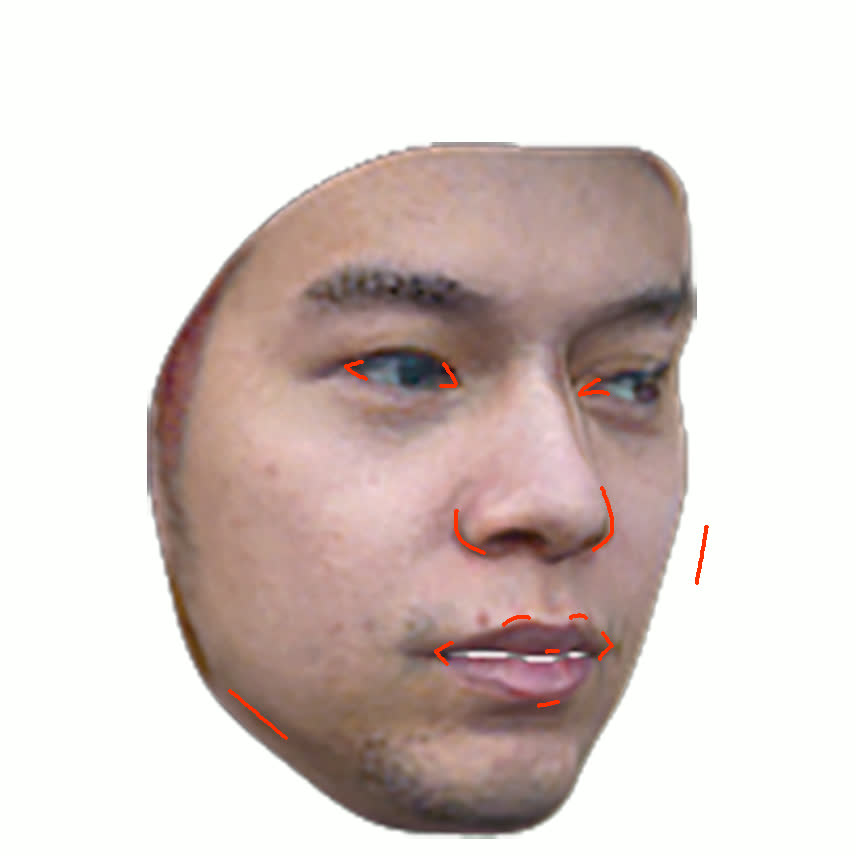}
    \end{minipage}
  \end{minipage}%
  \hfill%
  \begin{minipage}[t]{0.26\linewidth}
    \centering
    \coltitlelarge{NHA}\\[0.3em]
    \begin{minipage}[t]{0.5\linewidth}
      \centering
      \includegraphics[width=\linewidth]{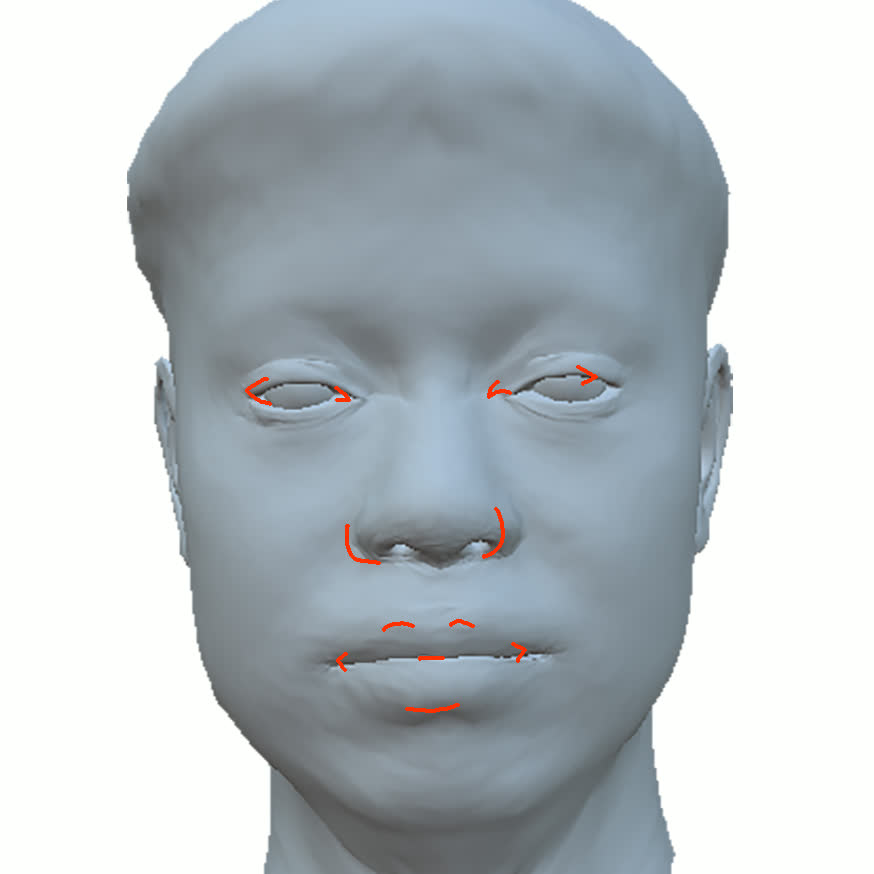}\\
      \includegraphics[width=\linewidth]{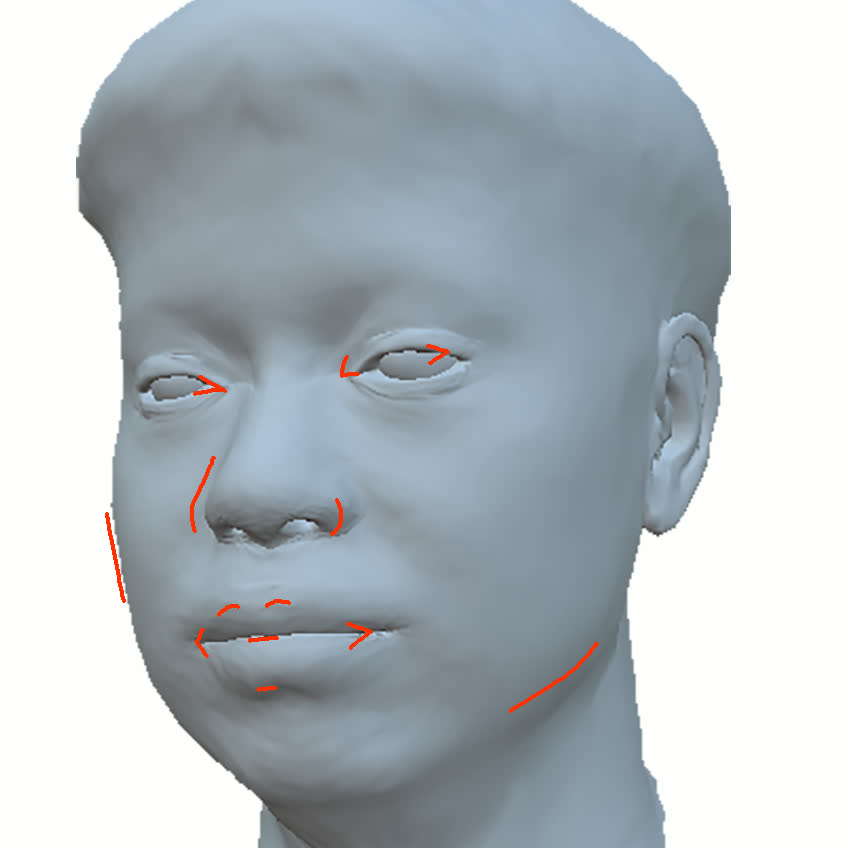}\\
      \includegraphics[width=\linewidth]{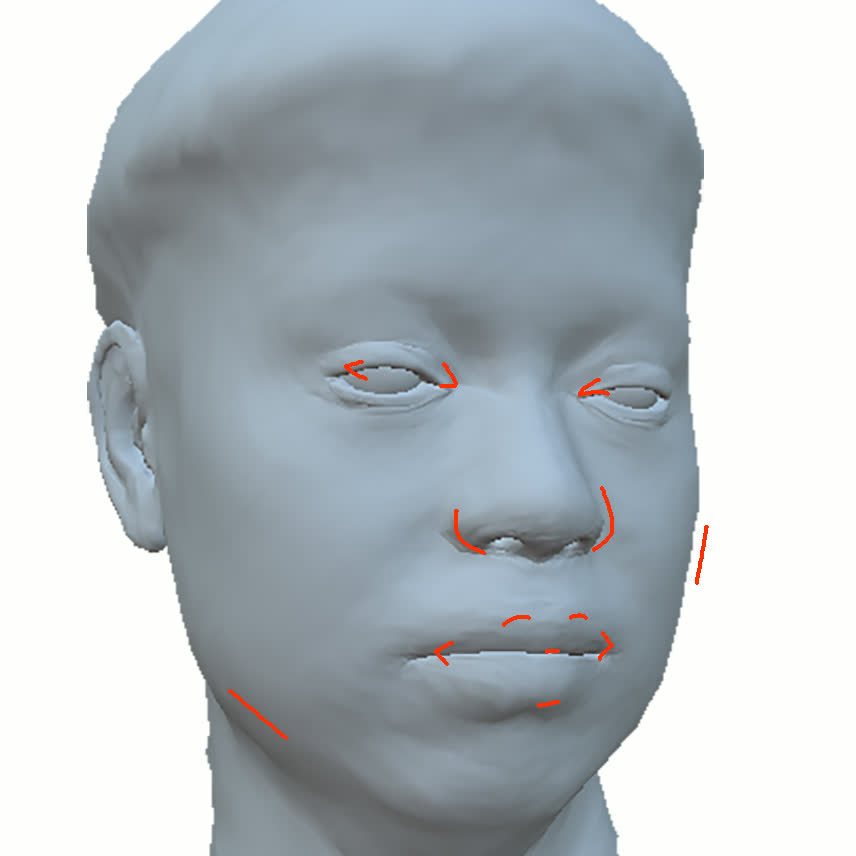}
    \end{minipage}%
    \begin{minipage}[t]{0.5\linewidth}
      \centering
      \includegraphics[width=\linewidth]{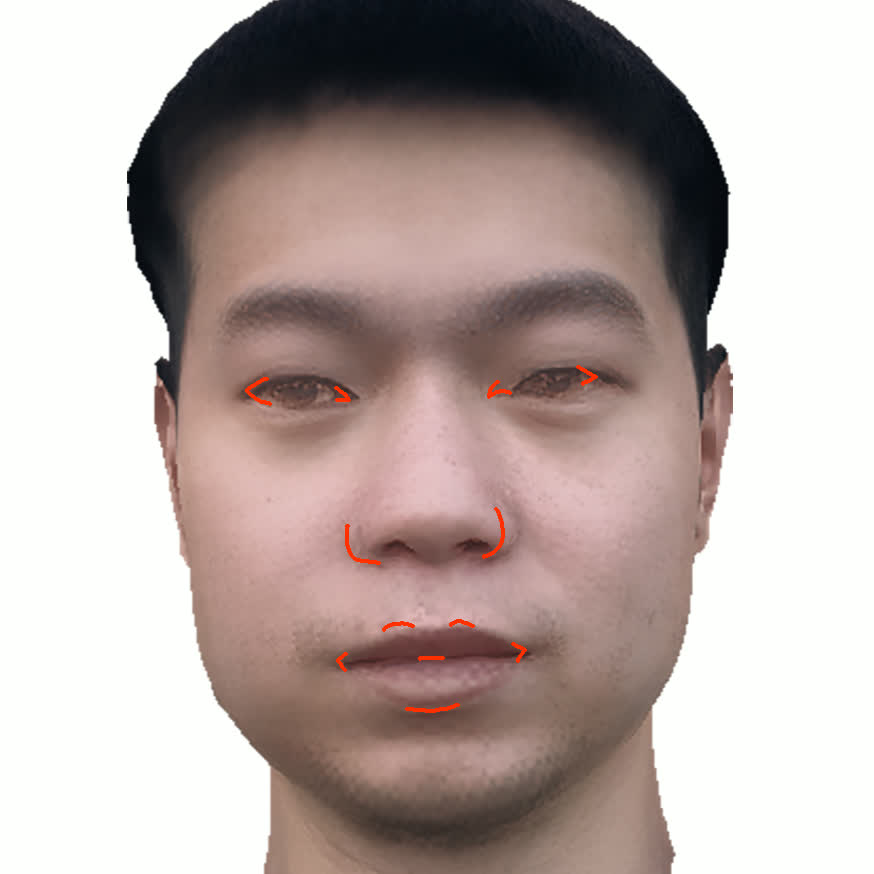}\\
      \includegraphics[width=\linewidth]{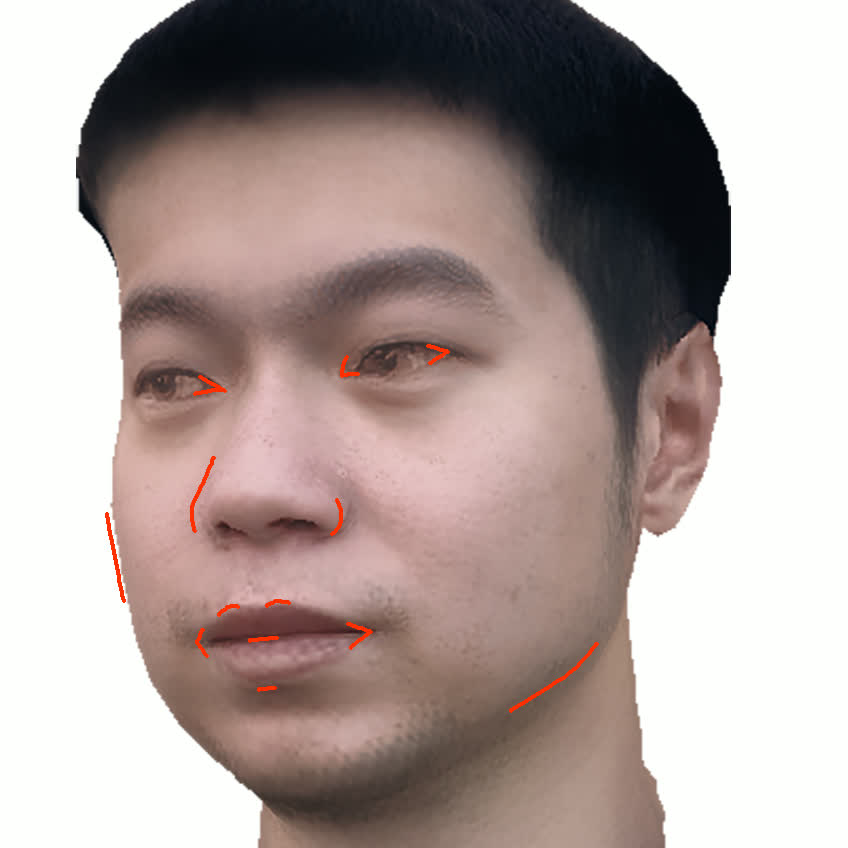}\\
      \includegraphics[width=\linewidth]{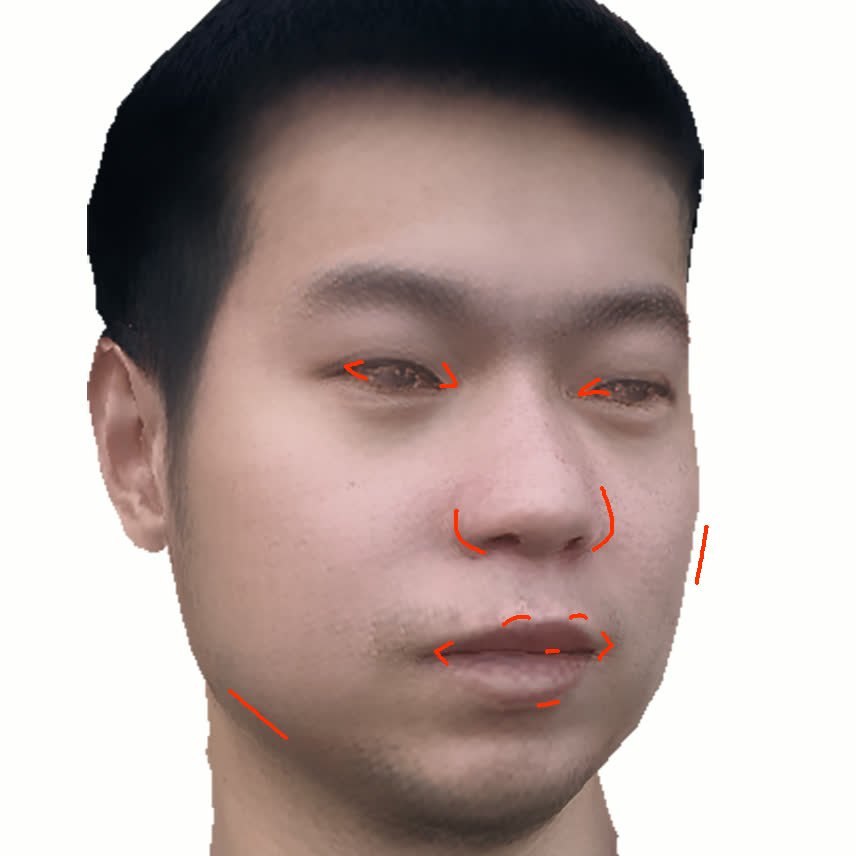}
    \end{minipage}
  \end{minipage}%

  \caption{
    Geometry reconstruction results comparing our method to NextFace~\cite{dib2021practical} and NHA~\cite{grassal2022neural}. Contour curves have been overlaid on key facial features in order to facilitate comparisons. Comparing column 2 to columns 4 and 6 illustrates that our method reconstructs better geometry. It is important to stress that the projection of colors (overfit to an image) onto geometry, as shown in columns 5 and 7, obfuscates errors in geometry reconstruction.}
  \label{fig:nf_comparison}
\end{figure*}
\begin{figure*}[htbp]
  \centering
  
  \hspace{0.12\linewidth}
  \begin{minipage}[t]{0.13\linewidth}
    \centering
    \coltitlelarge{Target Image}\\[0.3em]
    \includegraphics[width=\linewidth]{figs/nextface_comparison/00_gt.jpg}\\
  \end{minipage}%
  \hfill%
  \hfill%
  \begin{minipage}[t]{0.26\linewidth}
    \centering
    \coltitlelarge{NHA (w/ Expr.)}\\[0.3em]
    \begin{minipage}[t]{0.5\linewidth}
      \centering
      \includegraphics[width=\linewidth]{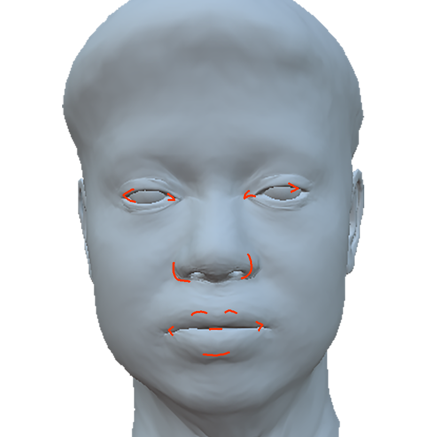}\\
    \end{minipage}%
    \begin{minipage}[t]{0.5\linewidth}
      \centering
      \includegraphics[width=\linewidth]{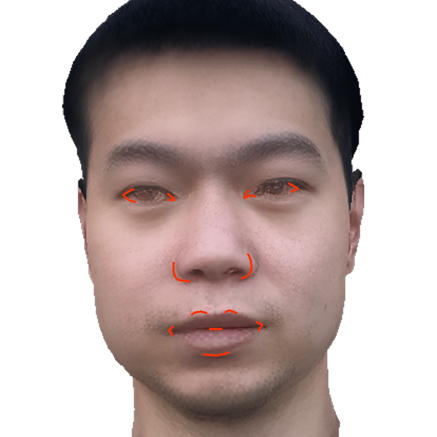}\\
    \end{minipage}
  \end{minipage}%
  \hfill%
  \begin{minipage}[t]{0.26\linewidth}
    \centering
    \coltitlelarge{NHA (Neutral)}\\[0.3em]
    \begin{minipage}[t]{0.5\linewidth}
      \centering
      \includegraphics[width=\linewidth]{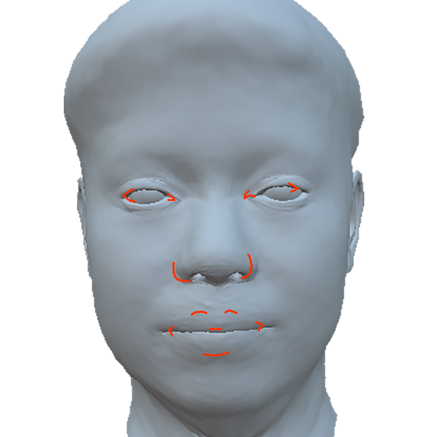}\\
    \end{minipage}%
    \begin{minipage}[t]{0.5\linewidth}
      \centering
      \includegraphics[width=\linewidth]{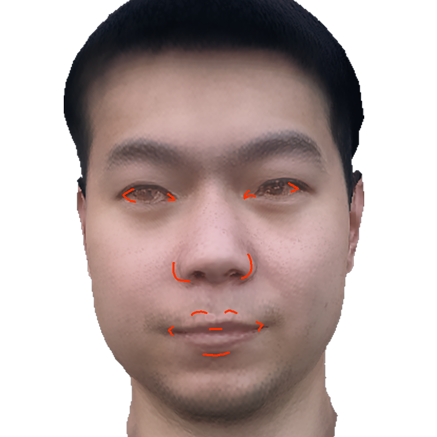}\\
    \end{minipage}
  \end{minipage}%
  \hspace{0.12\linewidth}

\caption{Allowing NHA~\cite{grassal2022neural} to overfit expressions to the target images does give slightly better results than those that are obtained by fixing the expression to be held in a neutral pose, as we did for Fig.~\ref{fig:nf_comparison} (compare the 2nd and 3rd images here to the top right of Fig.~\ref{fig:nf_comparison}). However, as is typical, this expression overfitting leads to a poor (unusable) neutral pose as shown in the last two images.}
\label{fig:nha_comparison}
\end{figure*}

For each in-the-wild reconstruction, we capture monocular video data as discussed in Sec.~\ref{sec:data_init}; then, eleven predefined target head poses are chosen for the reconstruction.

We ran various ablation tests to evaluate the efficacy of our contributions. Omitting our segmentation supervision causes Gaussians to be assigned to incorrect semantic regions, i.e., sliding from nearby regions (e.g. the lip region) or originating from different layers when Gaussians overlap (e.g. the eye region). Omitting our soft constraints disconnects the Gaussian explanation from their triangles causing the Gaussians to become disorganized with large size and shape variations. In both cases, the quality of the reconstructed mesh is deteriorated. See Fig.~\ref{fig:ablation}. Ablation test for the eye regularization loss and occlusion map are shown in Fig.~\ref{fig:eyes-reg} and Fig.~\ref{fig:occlusion-map}, respectively.





\begin{finalfigure*}[!t]
  \centering

  \begin{minipage}[t]{0.13\linewidth}
    \centering
    \coltitlelarge{Target Image}\\[0.3em]
    \includegraphics[width=\linewidth]{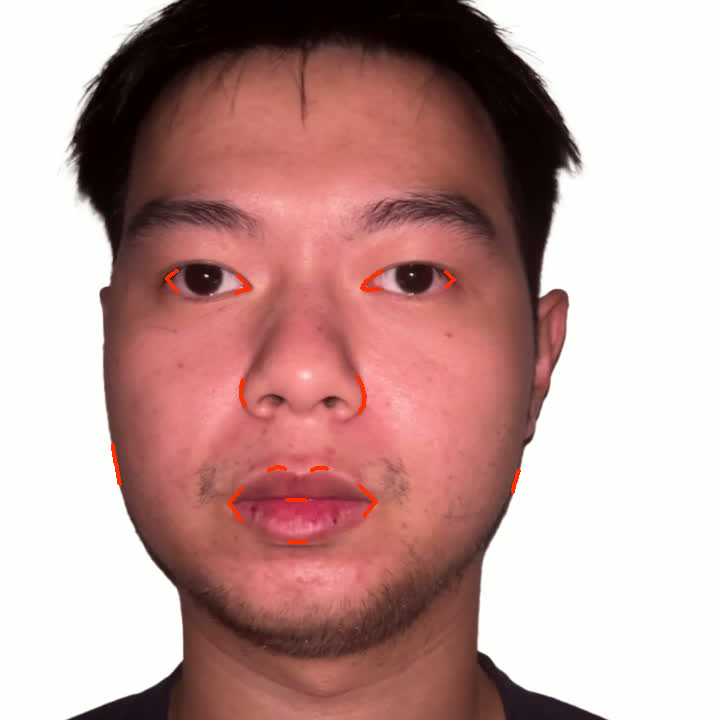}\\
    \includegraphics[width=\linewidth]{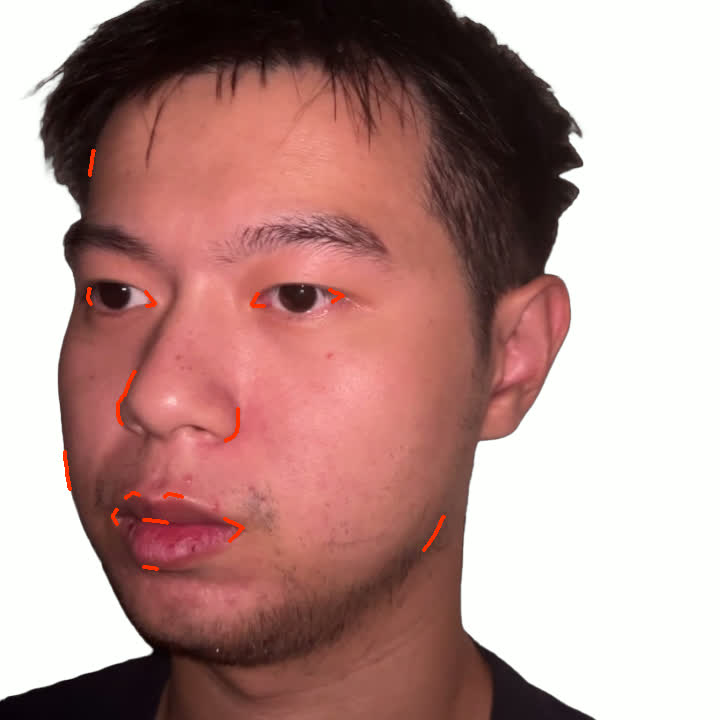}\\
    \includegraphics[width=\linewidth]{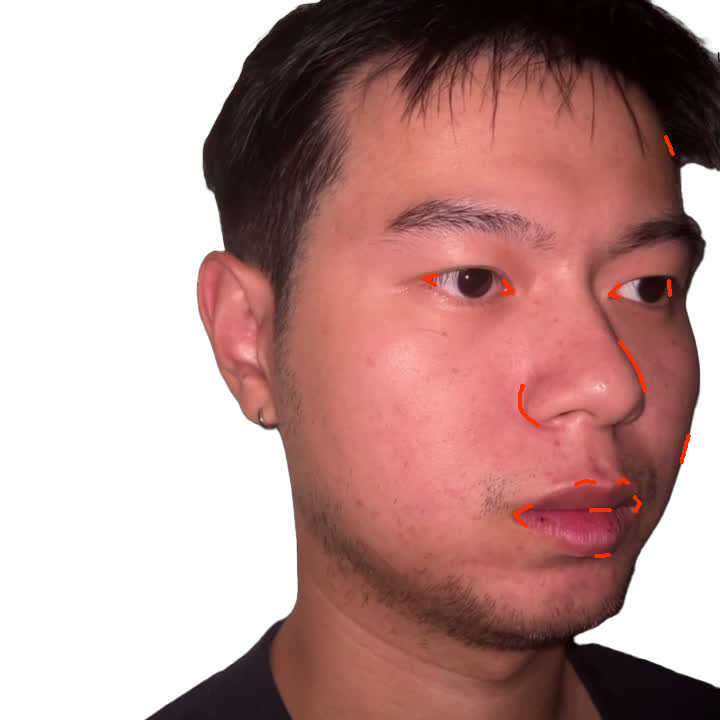}
  \end{minipage}%
  \hfill%
  \begin{minipage}[t]{0.39\linewidth}
    \centering
    \coltitlelarge{Ours}\\[0.3em]
    \begin{minipage}[t]{0.333\linewidth}
      \centering
      \includegraphics[width=\linewidth]{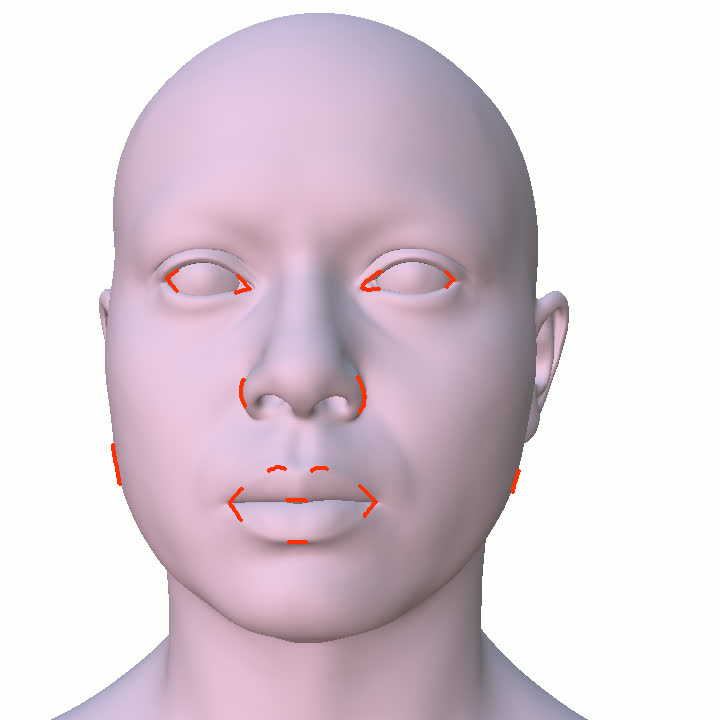}\\
      \includegraphics[width=\linewidth]{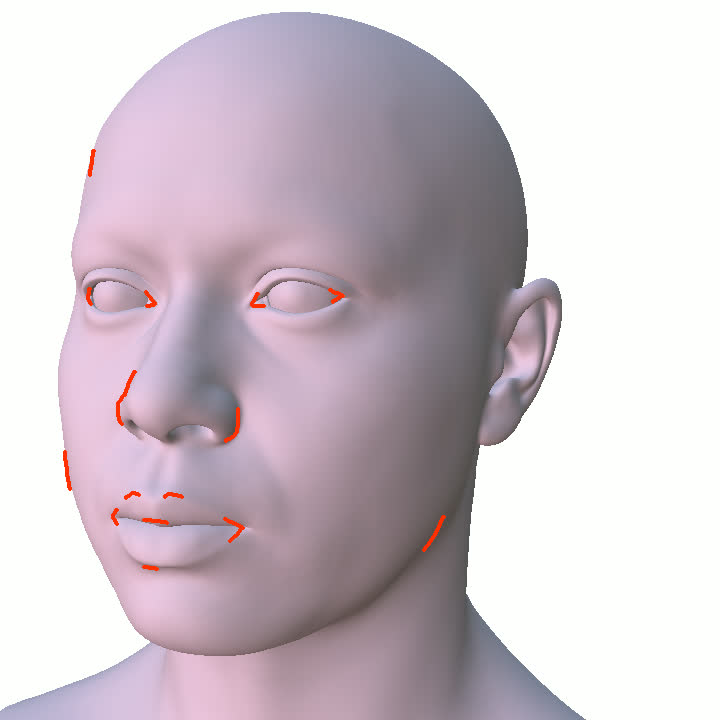}\\
      \includegraphics[width=\linewidth]{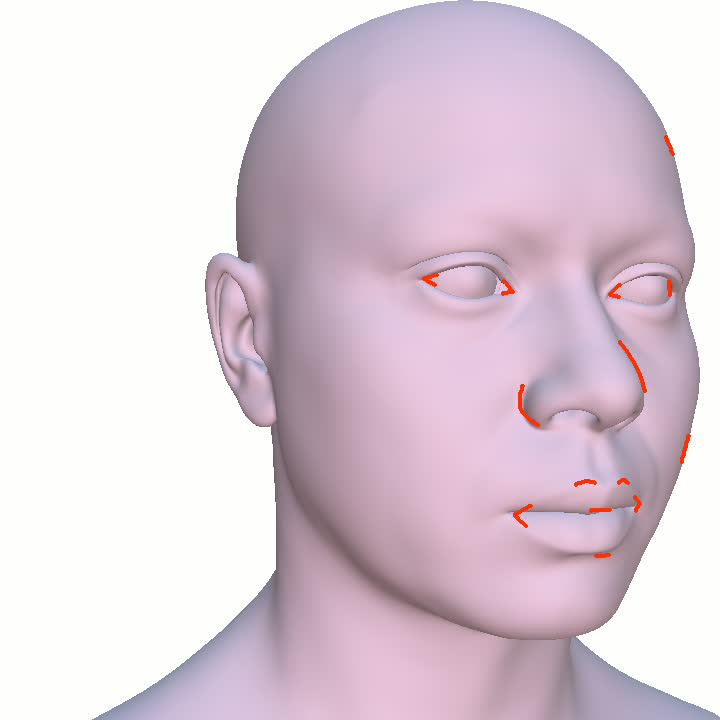}
    \end{minipage}%
    \begin{minipage}[t]{0.333\linewidth}
      \centering
      \includegraphics[width=\linewidth]{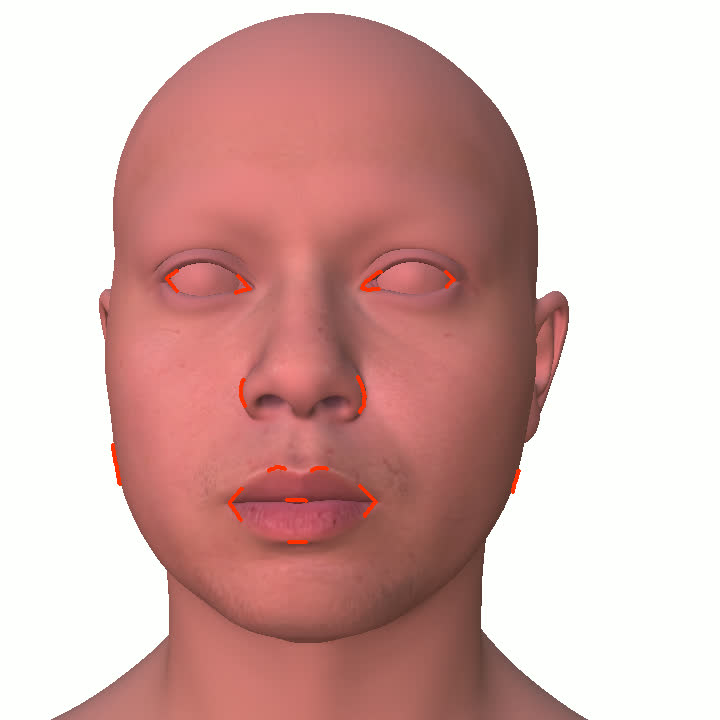}\\
      \includegraphics[width=\linewidth]{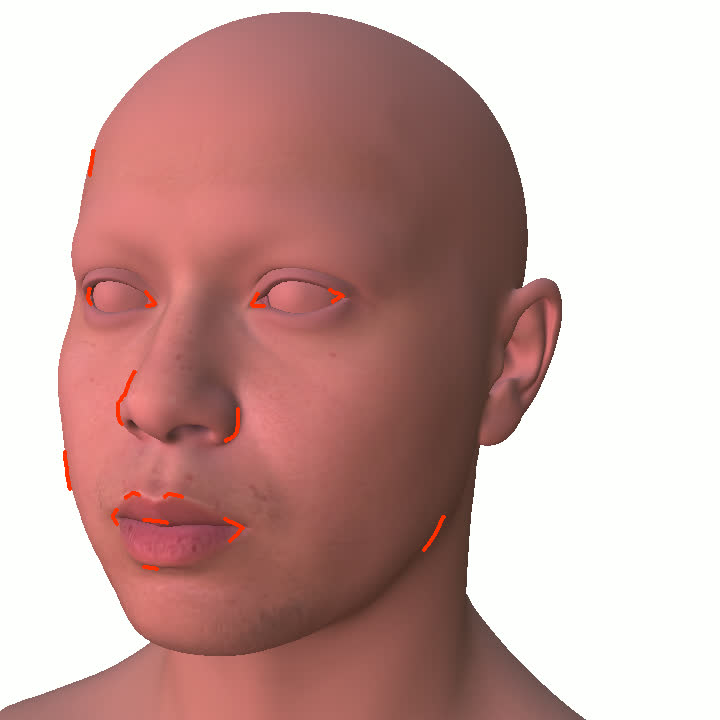}\\
      \includegraphics[width=\linewidth]{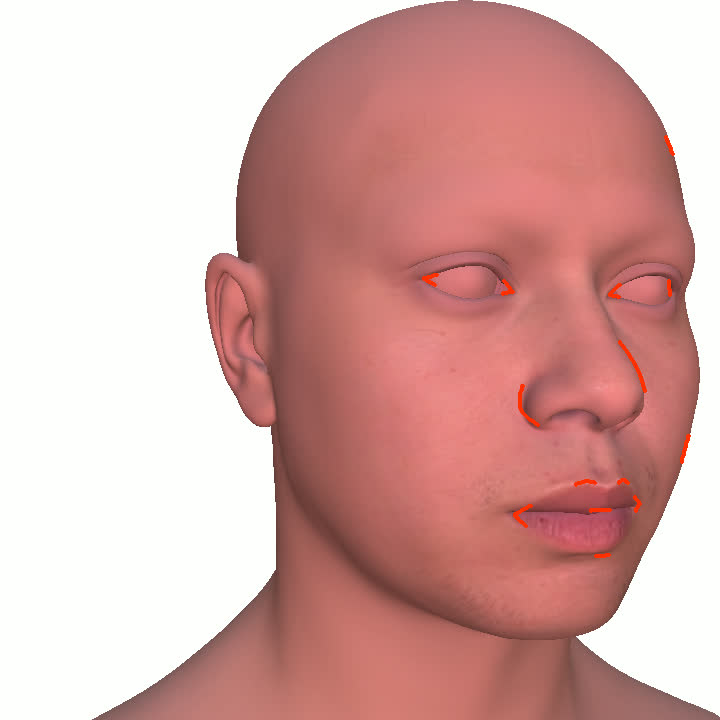}
    \end{minipage}%
    \begin{minipage}[t]{0.333\linewidth}
      \centering
      \includegraphics[width=\linewidth]{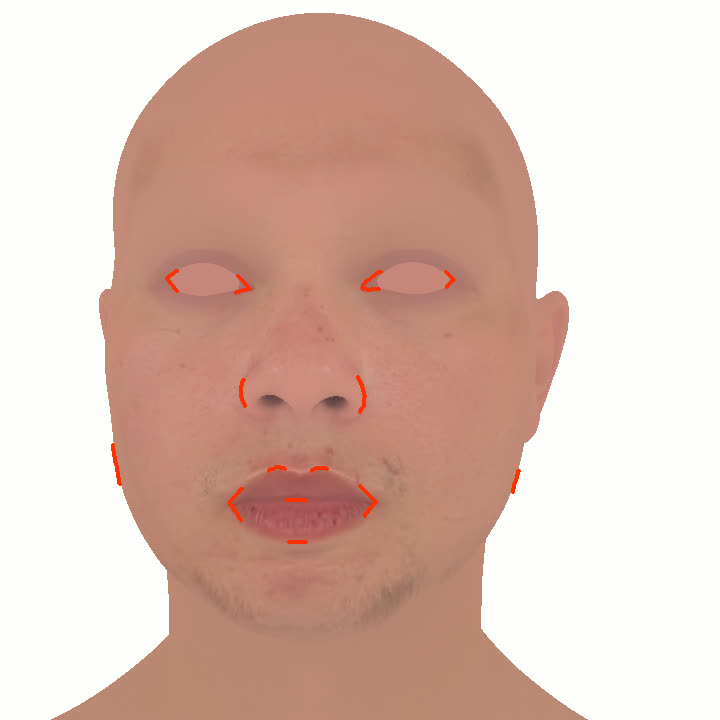}\\
      \includegraphics[width=\linewidth]{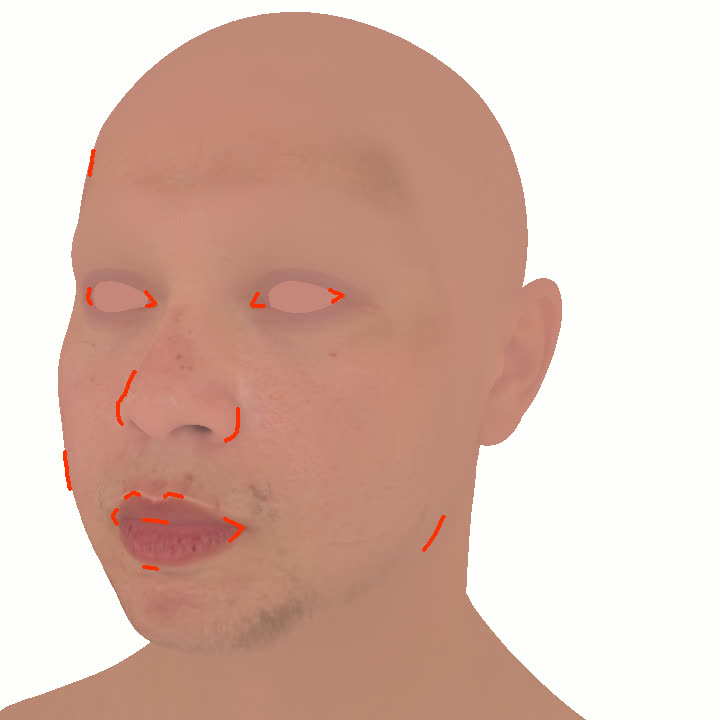}\\
      \includegraphics[width=\linewidth]{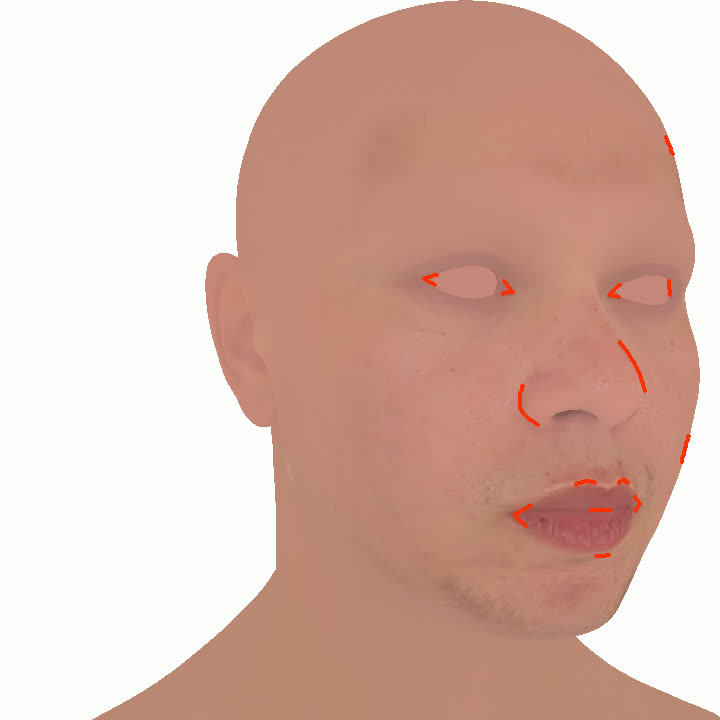}
    \end{minipage}
  \end{minipage}%
  \hfill%
  \begin{minipage}[t]{0.39\linewidth}
    \centering
    \coltitlelarge{CoRA}\\[0.3em]
    \begin{minipage}[t]{0.333\linewidth}
      \centering
      \includegraphics[width=\linewidth]{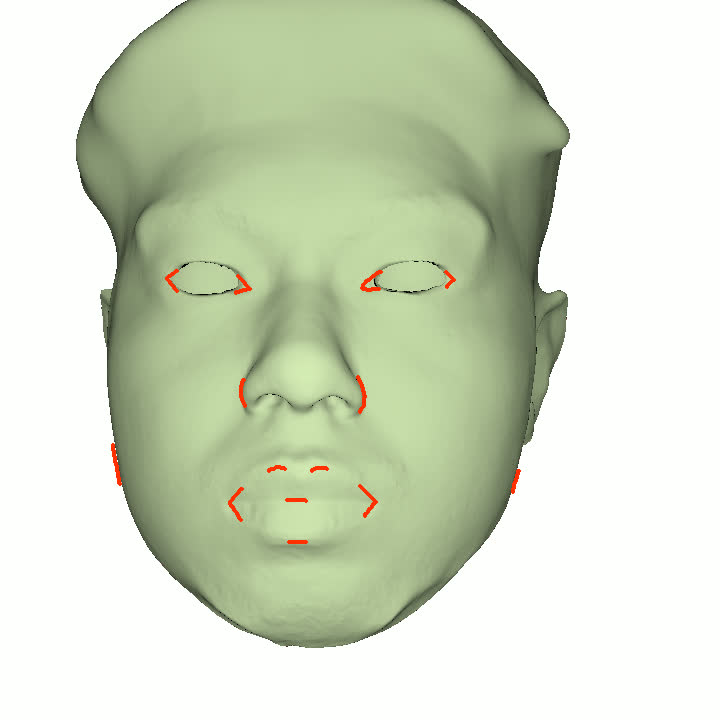}\\
      \includegraphics[width=\linewidth]{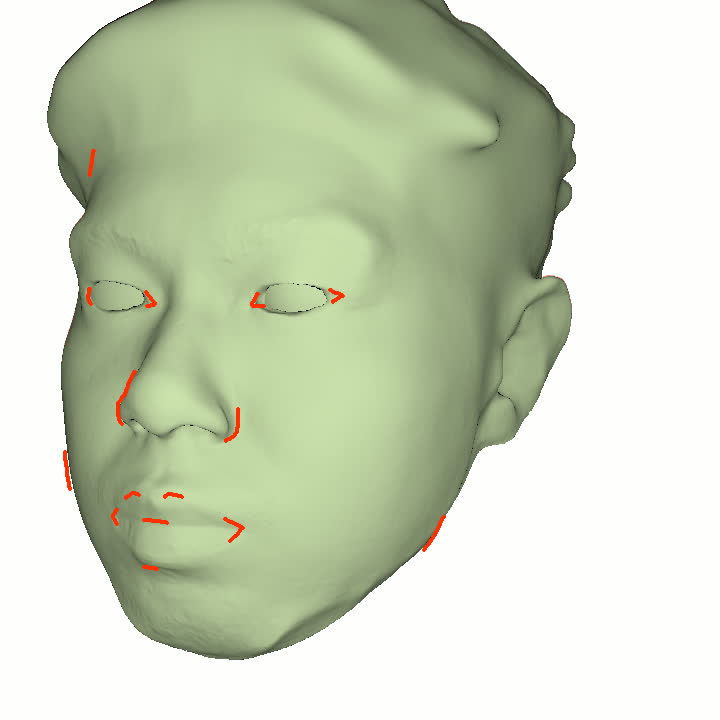}\\
      \includegraphics[width=\linewidth]{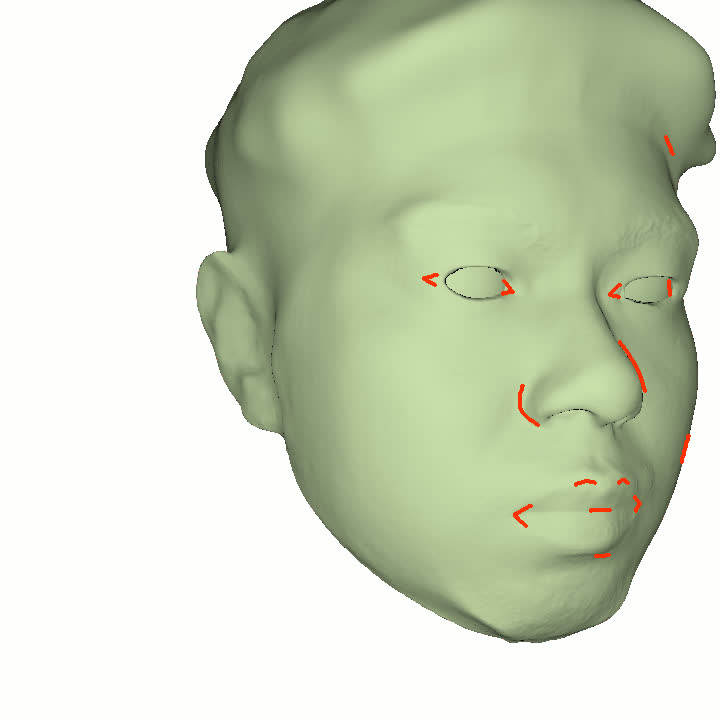}
    \end{minipage}%
    \begin{minipage}[t]{0.333\linewidth}
      \centering
      \includegraphics[width=\linewidth]{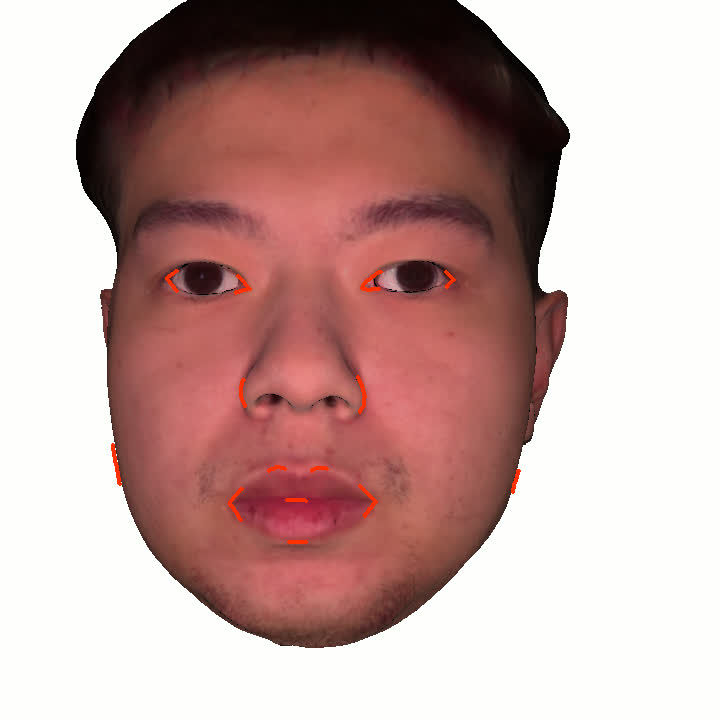}\\
      \includegraphics[width=\linewidth]{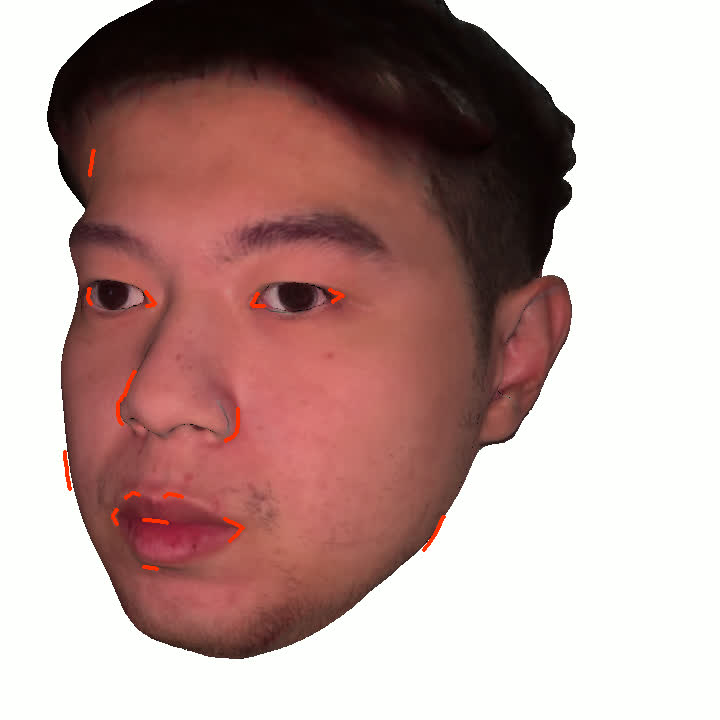}\\
      \includegraphics[width=\linewidth]{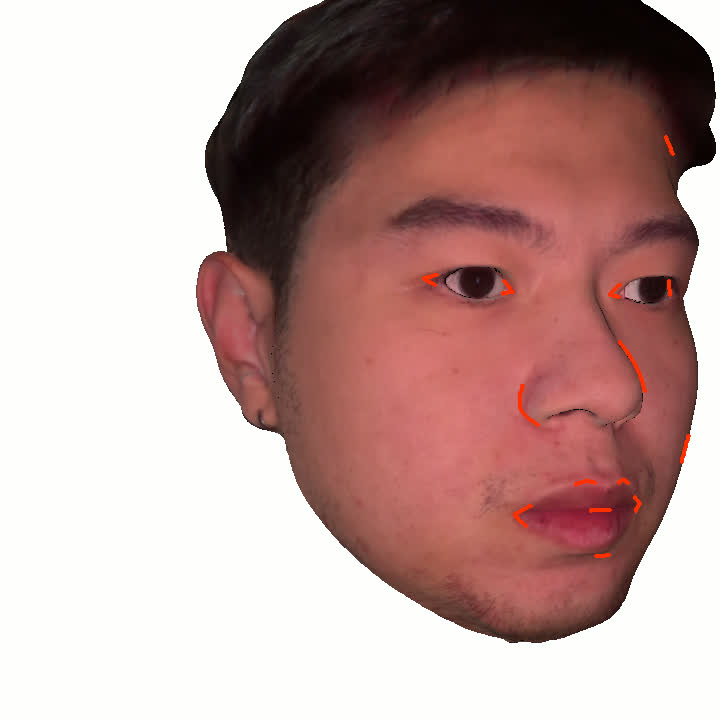}
    \end{minipage}%
    \begin{minipage}[t]{0.333\linewidth}
      \centering
      \includegraphics[width=\linewidth]{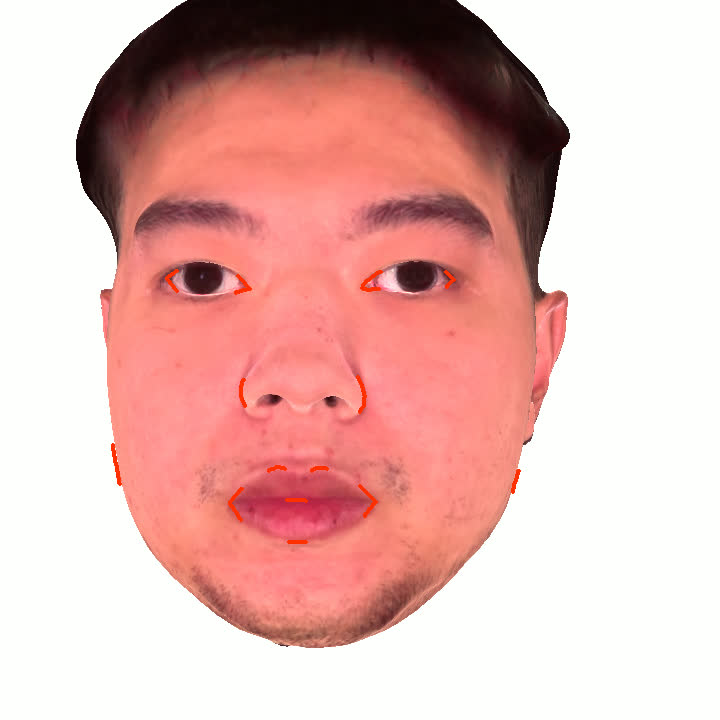}\\
      \includegraphics[width=\linewidth]{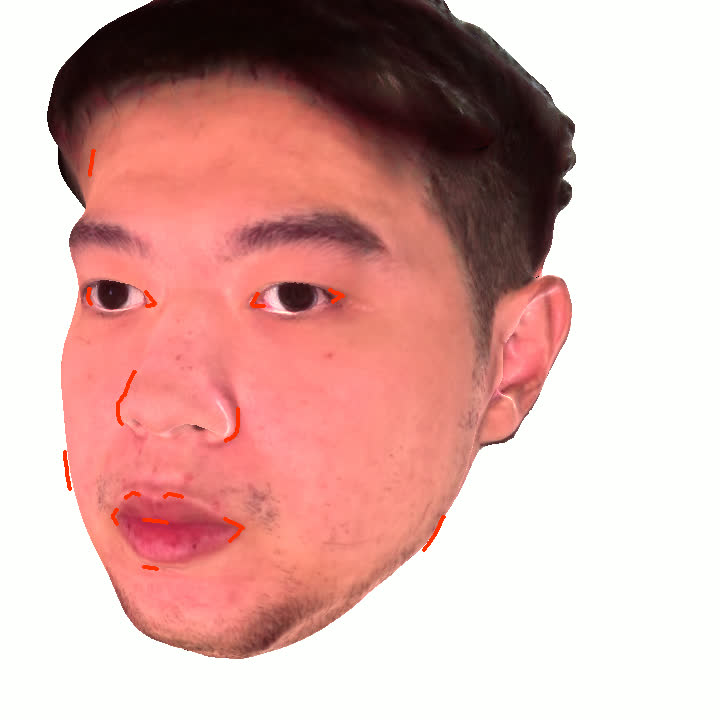}\\
      \includegraphics[width=\linewidth]{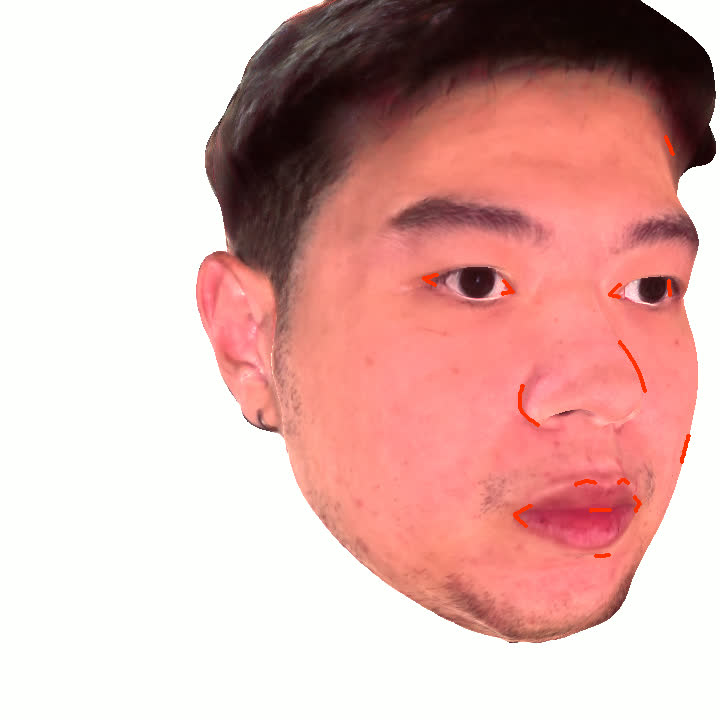}
    \end{minipage}
  \end{minipage}%

  \caption{
    Comparison between our method and CoRA~\cite{han2024high}, using their proposed method for data capture. Contour curves have been overlaid on key facial features in order to facilitate comparisons. Comparing column 2 to column 5 illustrates that our method reconstructs better geometry. CoRA introduces artifacts around the nose and jaw regions. These artifacts are also evident in the de-lit texture (column 7). Comparing column 4 to column 7 illustrates that CoRA has more baked-in lighting in their textures. This makes them incorrectly appear more three-dimensional, rather than flat (a de-lit texture should appear flat, see Fig.~\ref{fig:delit-example}). Columns 3 and 6 show the de-lit textures from columns 4 and 7 combined with the estimated lighting, in order to compare with the target image.} 
  \label{fig:cora_comparison}
\end{finalfigure*}

\begin{finalfigure}[thbp]
  \centering
  \hspace{0.12\linewidth}
  \hfill
  \begin{minipage}{0.31\linewidth}
    \centering \textbf{De-lit}\\[0.3em]
    \includegraphics[width=\linewidth]{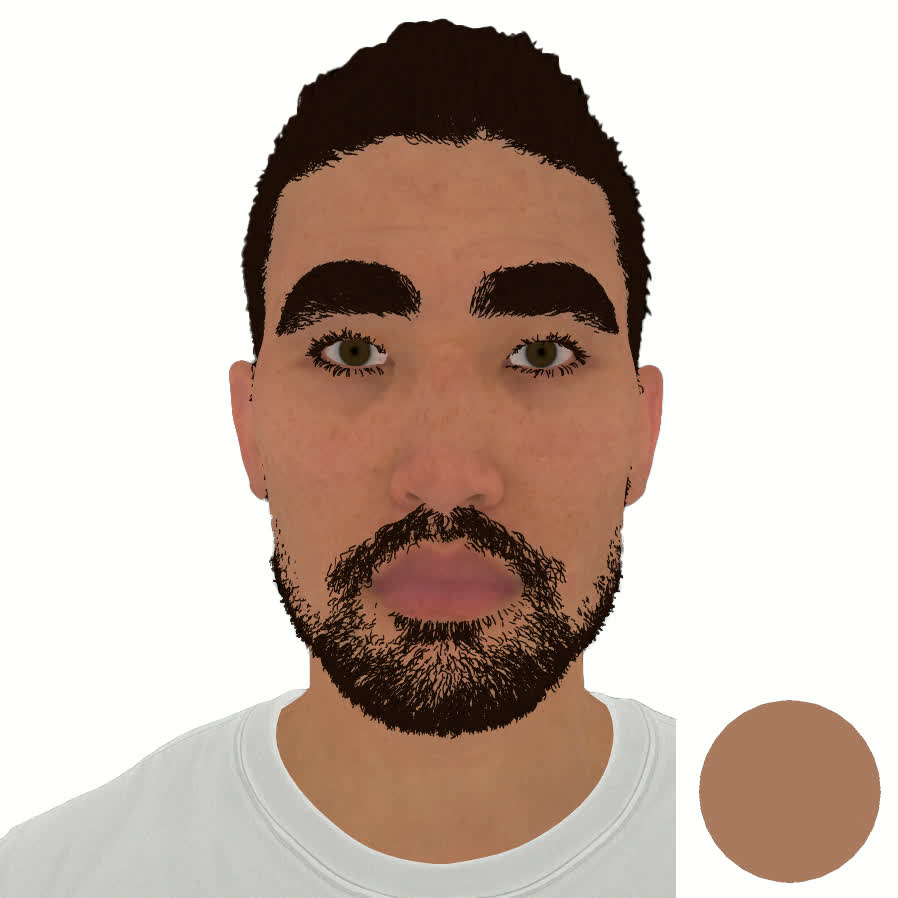}
  \end{minipage}%
  \hfill
  \begin{minipage}{0.31\linewidth}
    \centering \textbf{Lit}\\[0.3em]
    \includegraphics[width=\linewidth]{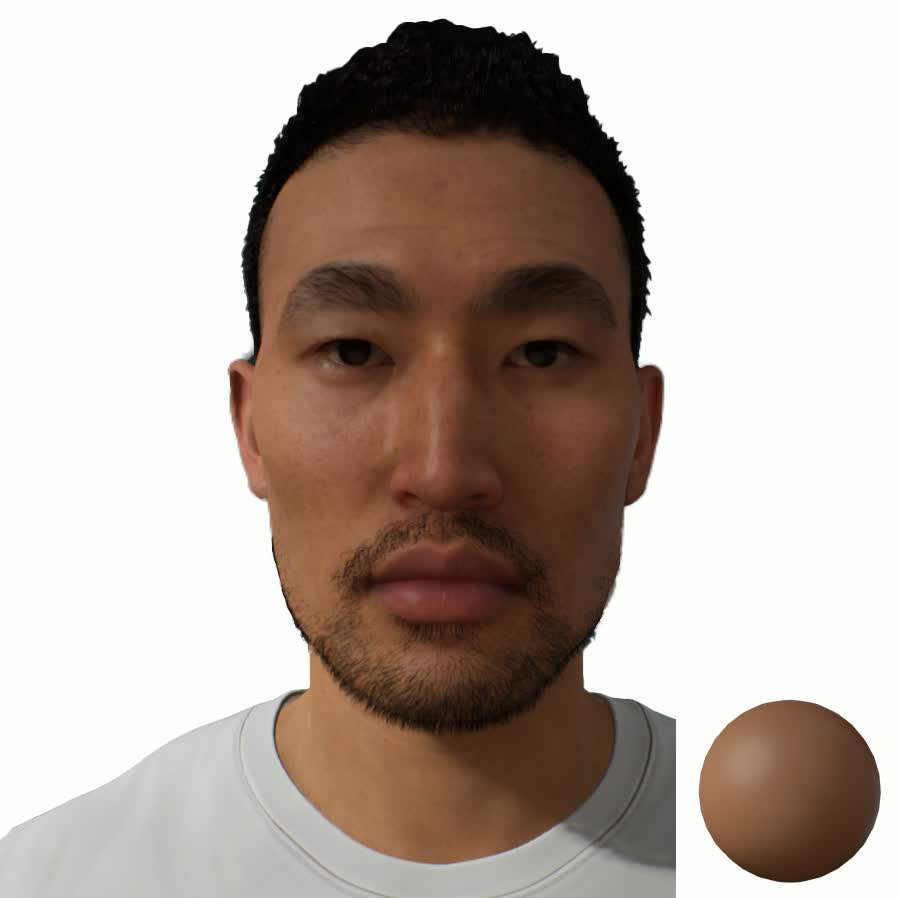}
  \end{minipage}%
  \hspace{0.12\linewidth}
  \hfill
  
  \caption{Left: Ambient-only lighting on both a MetaHuman and a sphere. Right: Full lighting on both. Note how flat the de-lit version appears to be, turning a three-dimensional sphere into a flat two-dimensional circle.}
  \label{fig:delit-example}
\end{finalfigure}

\subsection{Comparison with Prior Work}

Fig.~\ref{fig:nf_comparison} compares our method with NHA~\cite{grassal2022neural} and NextFace~\cite{dib2021practical}. For~\cite{dib2021practical}, we use the multi-image version for their reconstruction. For~\cite{grassal2022neural}, we set their expression and jaw pose coefficients to neutral and set their camera intrinsics to match ours; otherwise, their method overfits to obtain an unusable neutral expression (see Fig.~\ref{fig:nha_comparison}). Both of these state-of-the-art approaches exhibit misalignment of semantic facial regions and (especially~\cite{dib2021practical}) fail to capture the facial silhouette in side views. Lacking a strong semantic correspondence between their reconstructed geometry and the image data, many methods allow texture to slide on the geometry as it overfits to the image. This obfuscates errors in geometry reconstruction, not only from those who intend to use the method, but also from the method itself. That is, methods that allow for texture sliding hinder their own ability to improve the geometry, since the final result no longer differs from the image data. Our semantic segmentation and soft constraints have been devised specifically to address this issue. Since \cite{dib2021practical} and~\cite{grassal2022neural} do not obtain high quality de-lit textures, we instead compare our texture reconstruction approach to other methods.

Fig.~\ref{fig:cora_comparison} compares our method with the approach proposed in~\cite{han2024high}, which uses flashlight capture to obtain image data. Following their instructions, we obtained similar data for our subject. The results labeled ``CoRA'' in the figure use their approach on this data, and the results labeled ``Ours'' in the figure use our approach on a subset of the data (our approach requires only 11 frames, see Fig.~\ref{fig:head_poses}). The artifacts around the lips, nose, and jaw in the CoRA result are primarily due to an incorrect disentanglement of geometry and texture that allows errors in albedo and surface normals to offset each other; however, note that CoRA does achieve much better disentanglement than~\cite{dib2021practical} and~\cite{grassal2022neural}. In comparison to CoRA, our method recovers sharper boundaries for facial features. This is especially important because their results are not readily animatable (ours are), meaning that further errors will result when converting their soft implicit surface boundaries to an explicit mesh that can be rigged for animation. Importantly, our method achieves similar de-lit textures under various lighting conditions (see Fig.~\ref{fig:our-delit}). This consistency foreshadows how well our de-lit texture can be re-lit under novel lighting conditions, as shown in Fig.~\ref{fig:relighting}. 

\begin{finalfigure*}[htbp]
  \centering


  \begin{minipage}[t]{0.15\linewidth}
    \centering
    \coltitle{Target Image}\\[0.3em]
    \includegraphics[width=\linewidth]{figs/nextface_comparison/00_gt.jpg}
  \end{minipage}
  \hfill%
  \begin{minipage}[t]{0.15\linewidth}
    \centering
    \coltitle{Ours \\(lit)}\\[0.3em]
    \includegraphics[width=\linewidth]{figs/nextface_comparison/00_ours_texture.jpg}
  \end{minipage}%
  \hfill%
  \begin{minipage}[t]{0.15\linewidth}
    \centering
    \coltitle{Ours \\(de-lit)}\\[0.3em]
    \includegraphics[width=\linewidth]{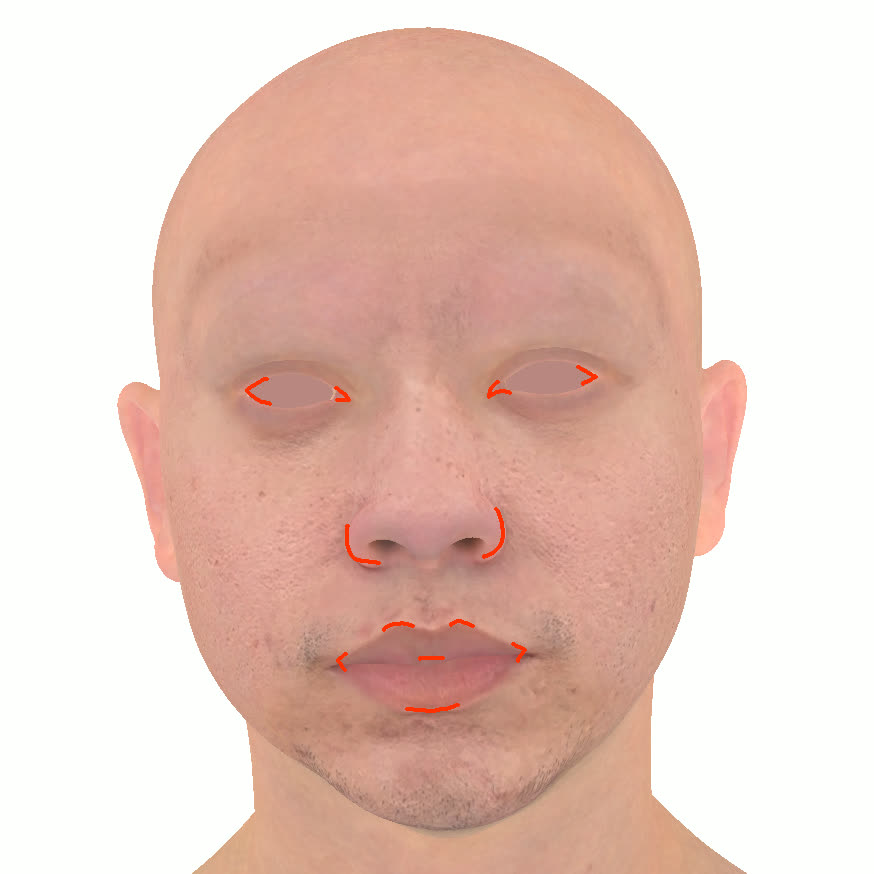}
  \end{minipage}%
  \hfill%
  \begin{minipage}[t]{0.15\linewidth}
    \centering
    \coltitle{Ours \\(de-lit)}\\[0.3em]
    \includegraphics[width=\linewidth]{figs/cora_comparison/01_ours_albedo.jpg}
  \end{minipage}%
  \hfill%
  \begin{minipage}[t]{0.15\linewidth}
    \centering
    \coltitle{Ours \\(lit)}\\[0.3em]
    \includegraphics[width=\linewidth]{figs/cora_comparison/01_ours_texture.jpg}
  \end{minipage}%
  \hfill%
  \begin{minipage}[t]{0.15\linewidth}
    \centering
    \coltitle{Target Image}\\[0.3em]
    \includegraphics[width=\linewidth]{figs/cora_comparison/01_gt.jpg}
  \end{minipage}%

  \caption{Note how similar our de-lit textures appear (compare columns 3 and 4), despite the disparate lighting in the two tests. Also note how well our de-lit textures respond to the lighting (columns 2 and 5), allowing for good matching to the target images (columns 1 and 6).}
    
  \label{fig:our-delit}
\end{finalfigure*}



\begin{finalfigure}[htbp]
  \centering


  \begin{minipage}[t]{0.225\linewidth}
    \centering
    \coltitle{Lighting}\\[0.3em]
    \includegraphics[width=\linewidth]{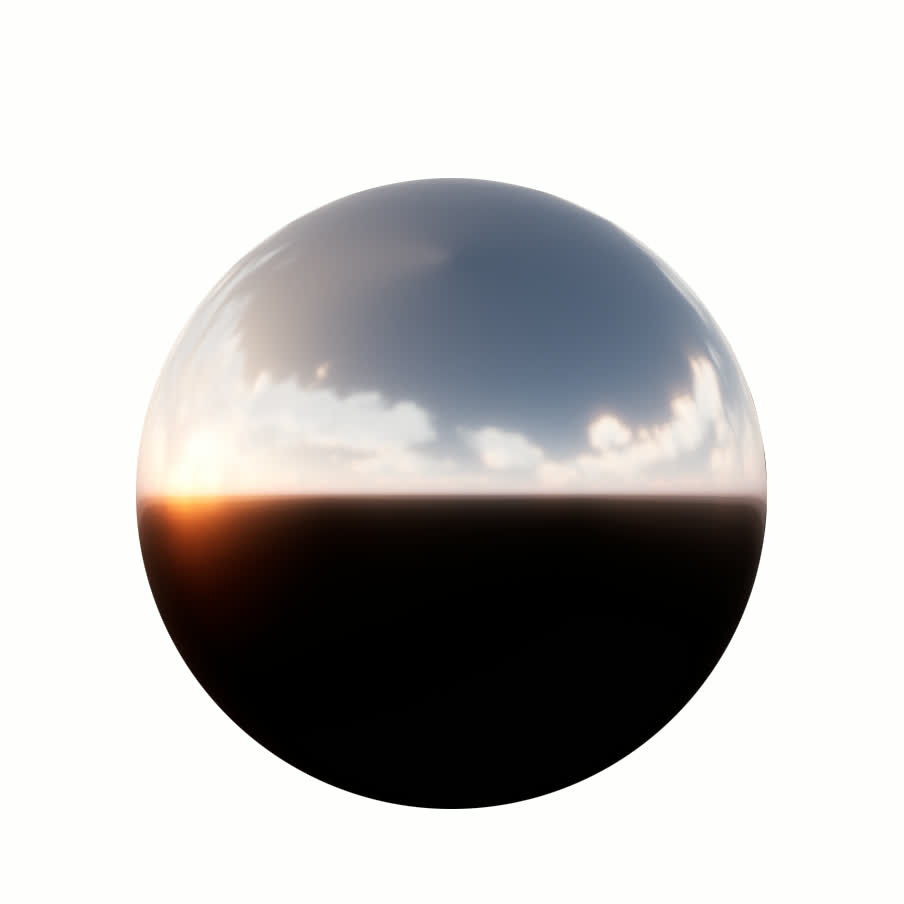}\\
    \includegraphics[width=\linewidth]{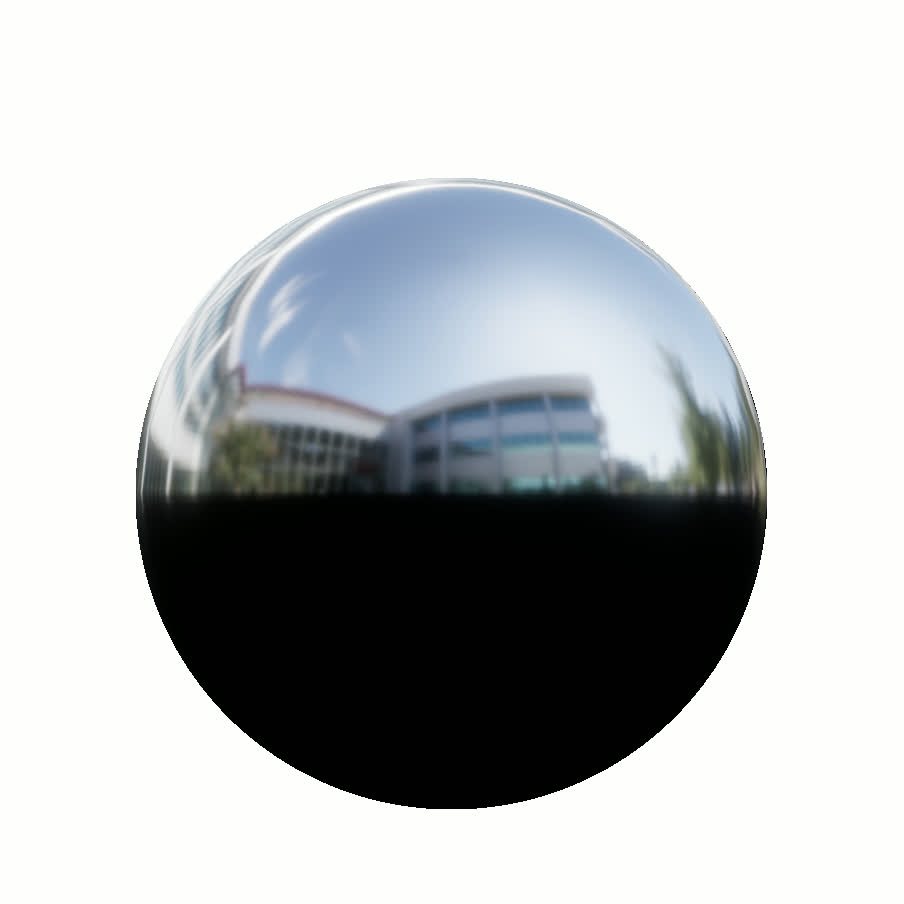}\\
    \includegraphics[width=\linewidth]{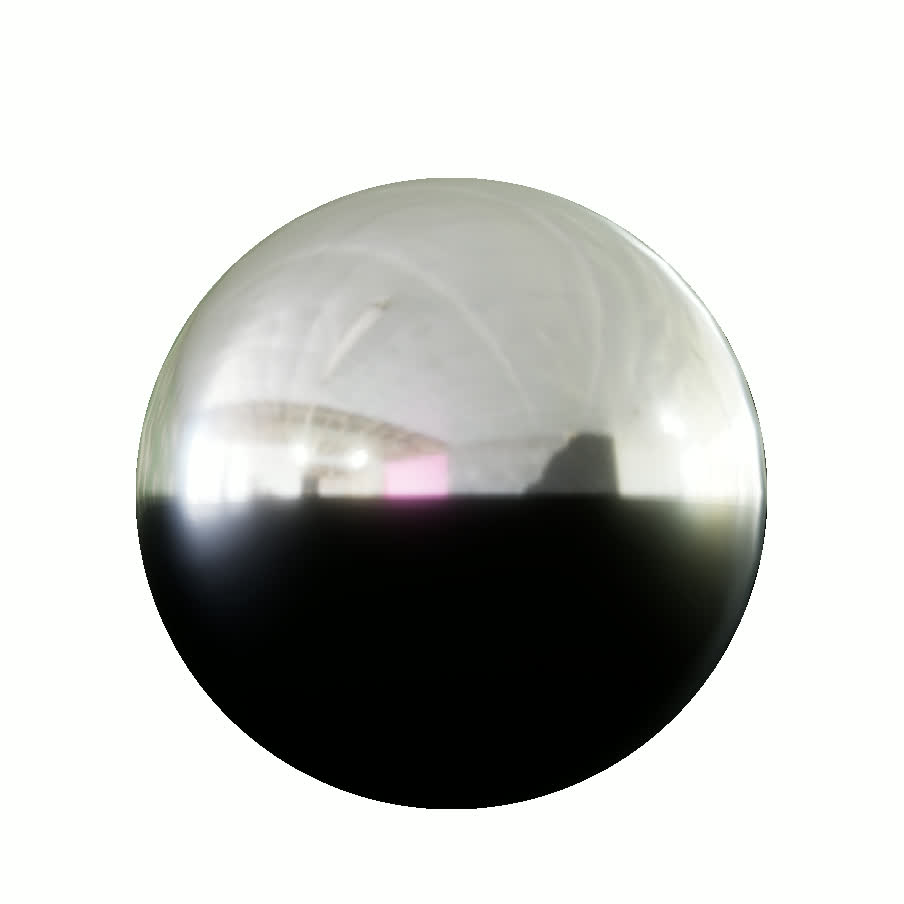}
  \end{minipage}
  \hfill
  \begin{minipage}[t]{0.225\linewidth}
    \centering
    \coltitle{Reference}\\[0.3em]
    \includegraphics[width=\linewidth]{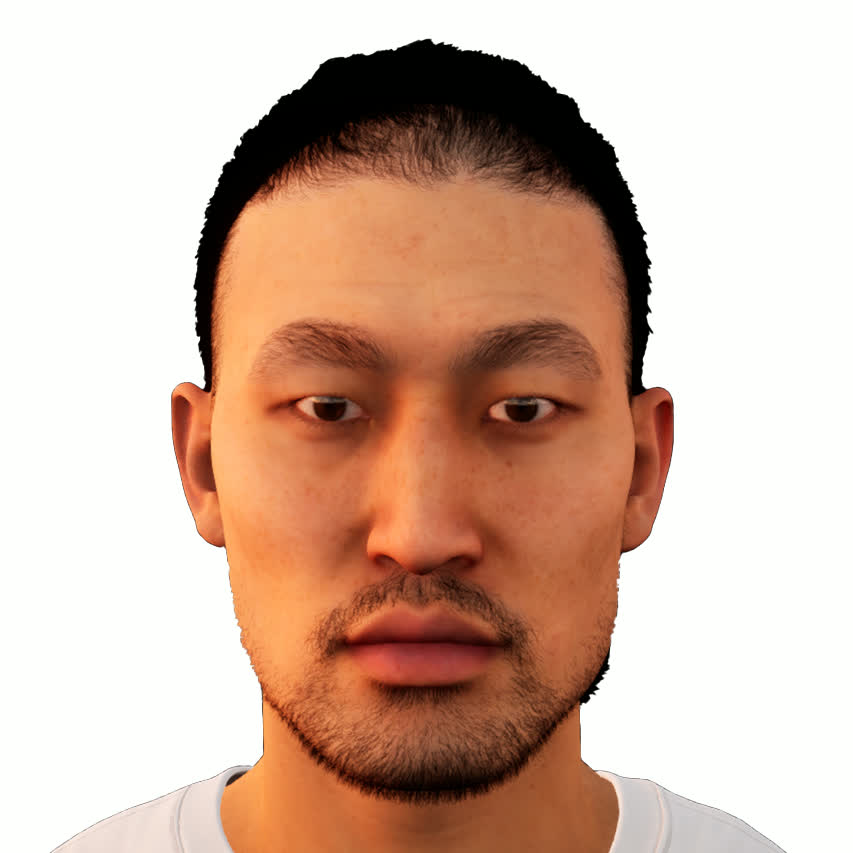}\\
    \includegraphics[width=\linewidth]{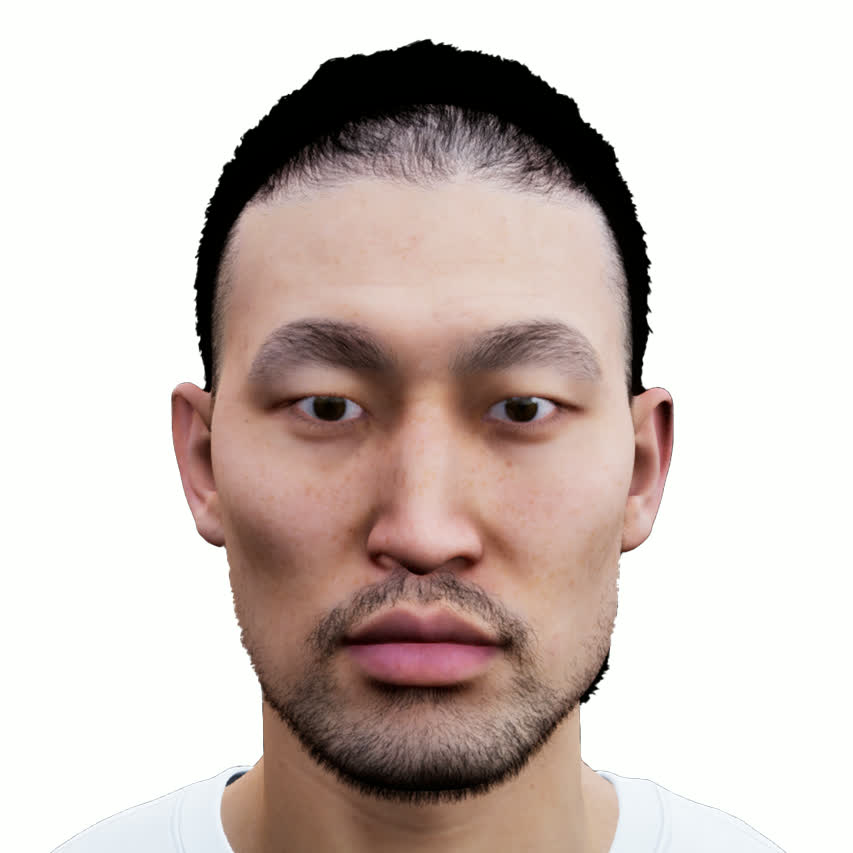}\\
    \includegraphics[width=\linewidth]{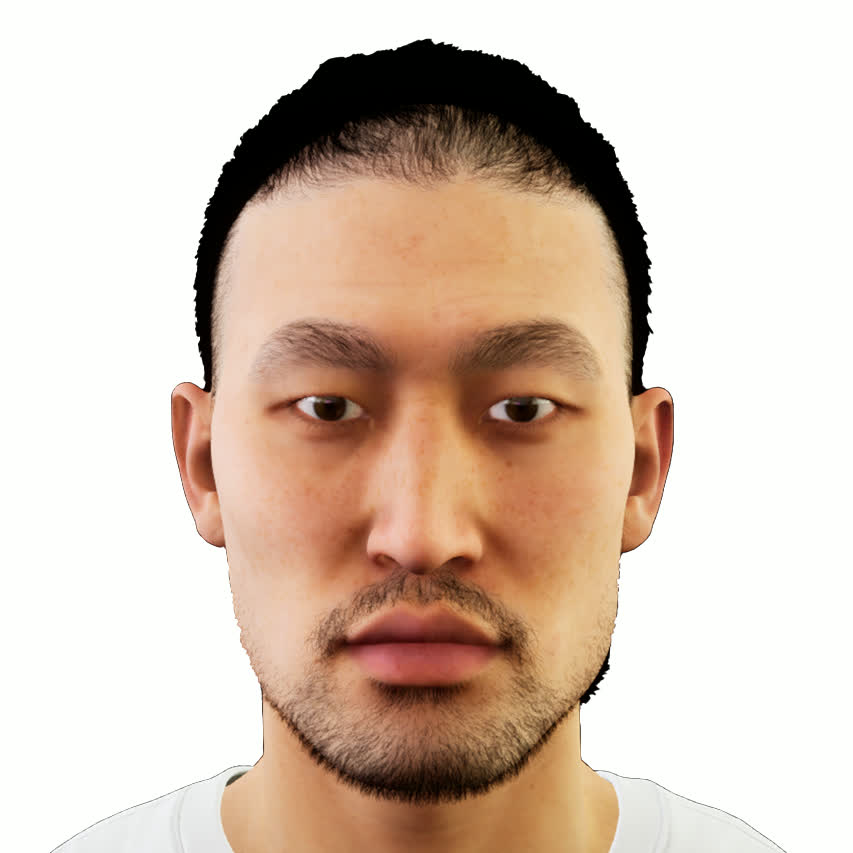}
  \end{minipage}
  \hfill
  \begin{minipage}[t]{0.225\linewidth}
    \centering
    \coltitle{Ours}\\[0.3em]
    \includegraphics[width=\linewidth]{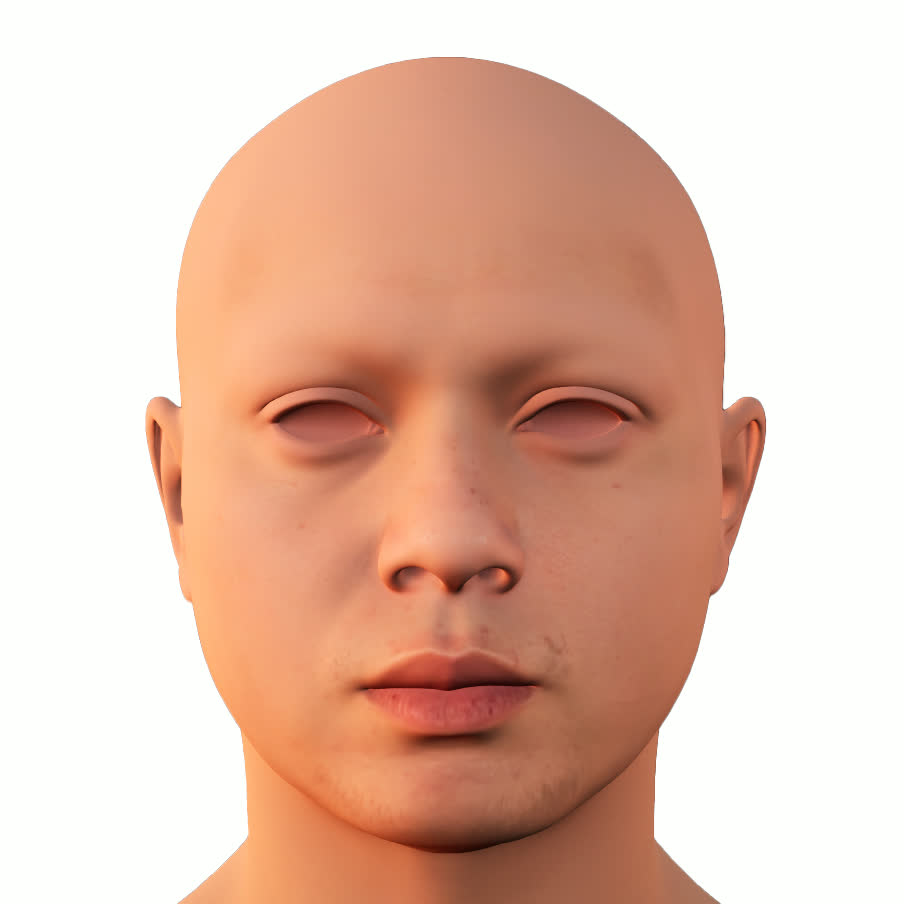}\\
    \includegraphics[width=\linewidth]{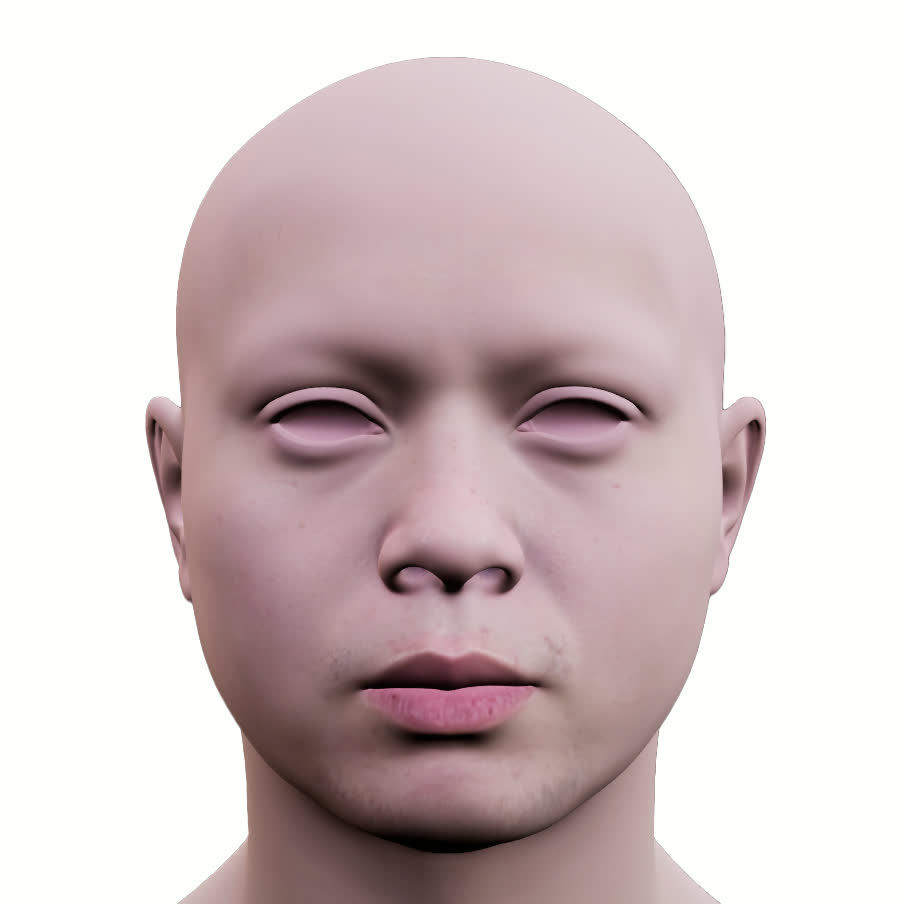}\\
    \includegraphics[width=\linewidth]{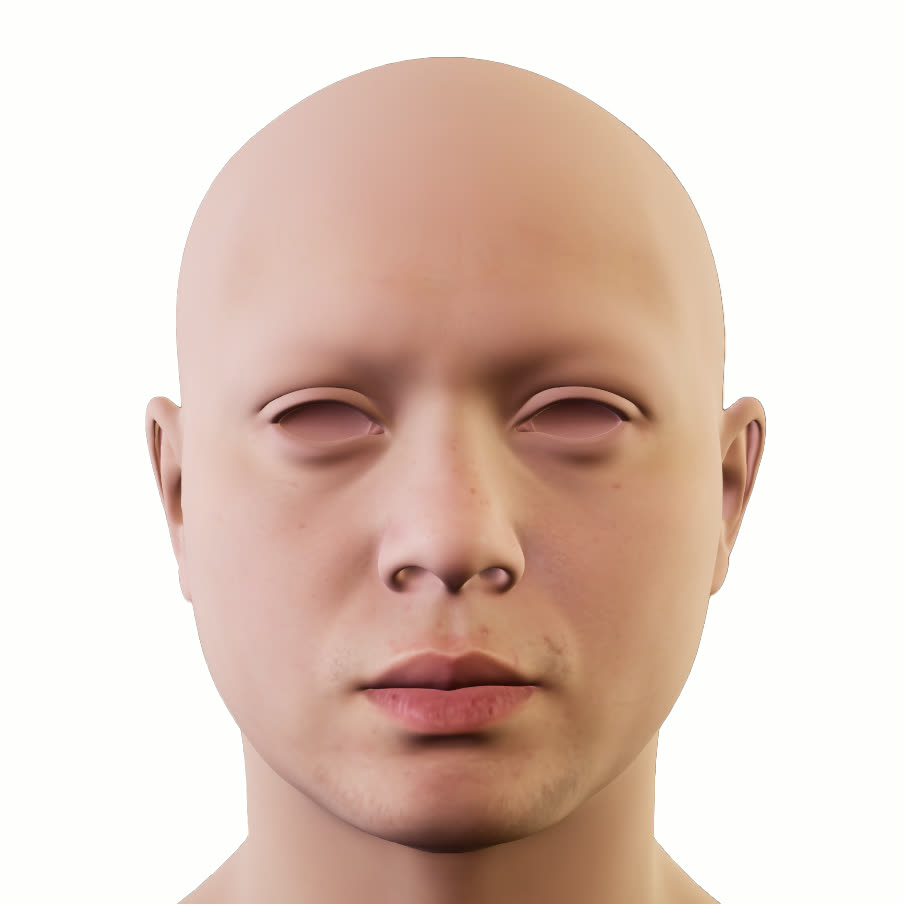}
  \end{minipage}
  \hfill
  \begin{minipage}[t]{0.225\linewidth}
    \centering
    \coltitle{CoRA}\\[0.3em]
    \includegraphics[width=\linewidth]{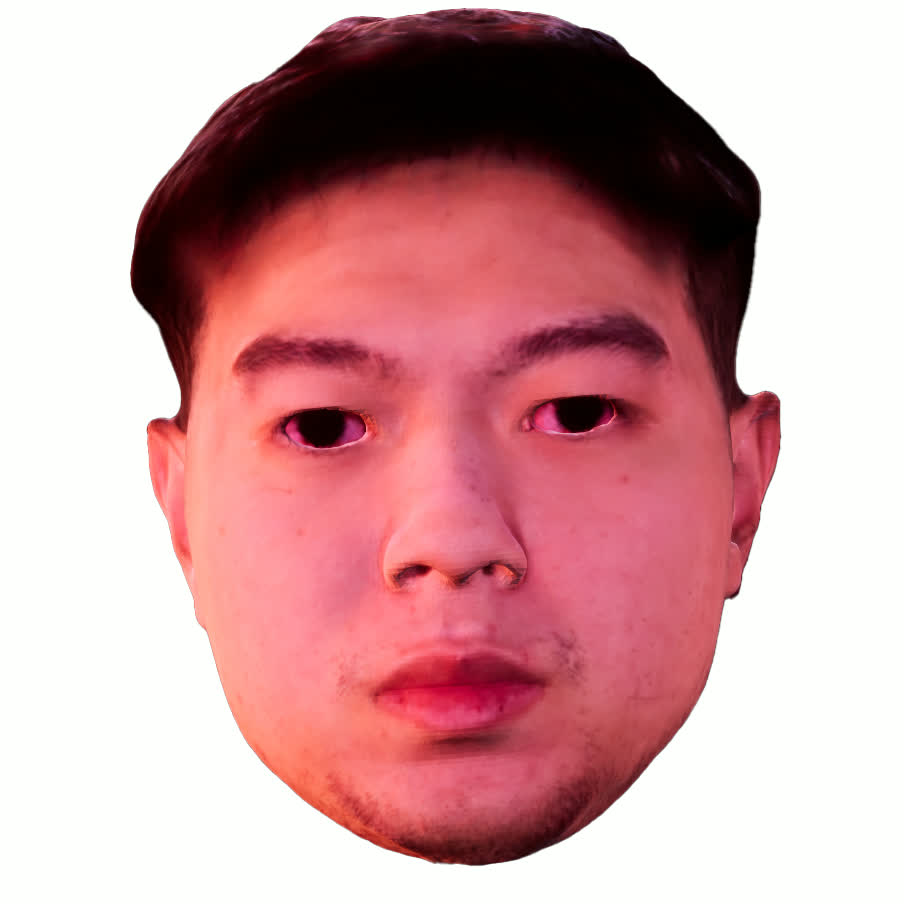}\\
    \includegraphics[width=\linewidth]{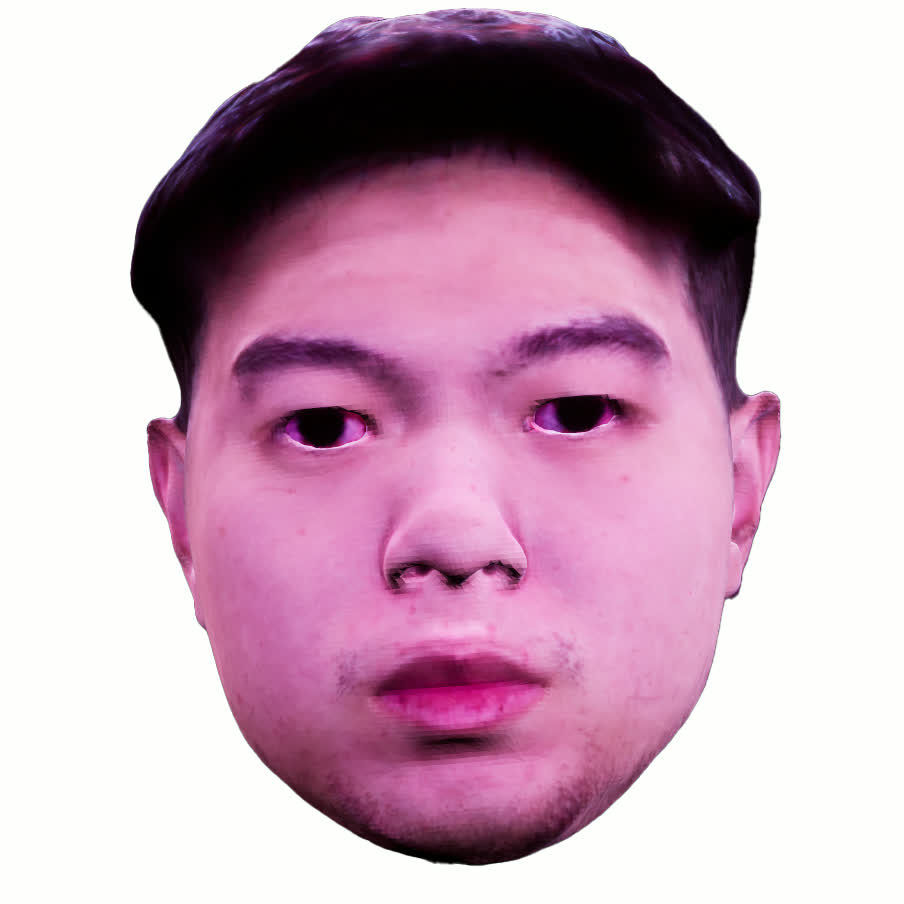}\\
    \includegraphics[width=\linewidth]{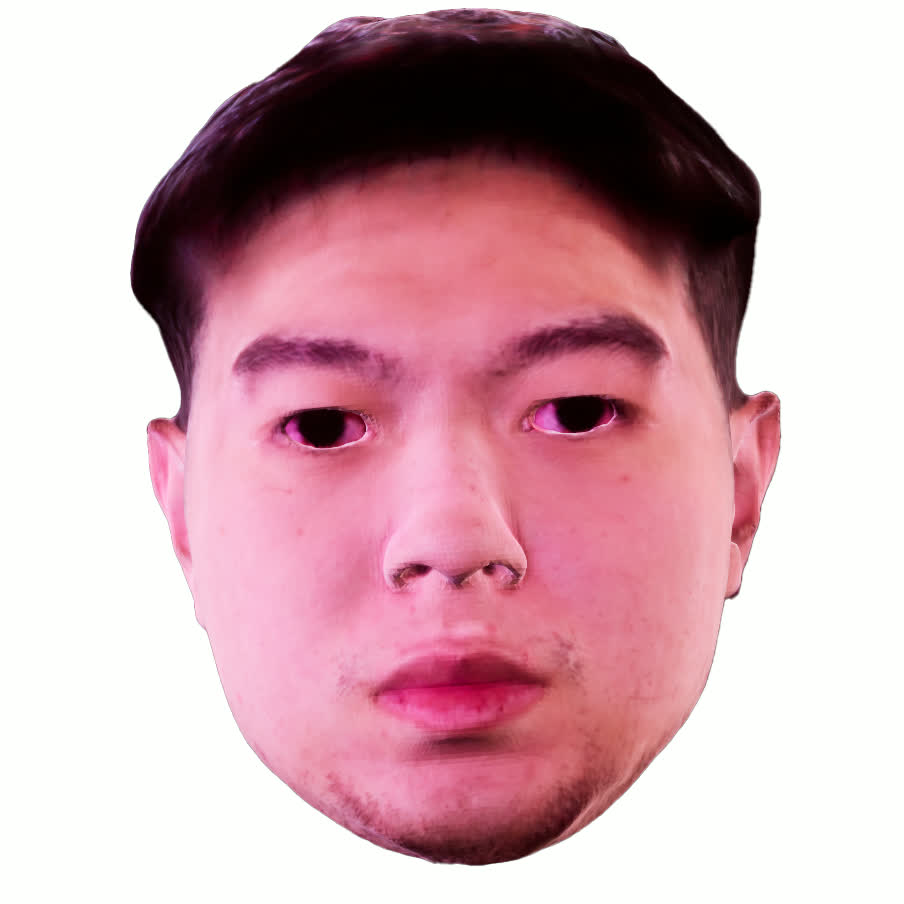}
  \end{minipage}%
  \caption{Each row illustrates a novel lighting condition, depicted by a chrome sphere (column 1) and the rendering of an artist-created MetaHuman (column 2). Our reconstructed texture (column 3) matches the reference quite well, especially when compared to CoRA (column 4).}
  \label{fig:relighting}
\end{finalfigure}

\subsection{Utilizing Disparate Captures}

\begin{figure}[htbp]
  \begin{minipage}[t]{\linewidth}
    \centering
    \begin{minipage}{0.32\linewidth}
      \centering \textbf{Target Image}\\[0.3em]
      \includegraphics[width=\linewidth]{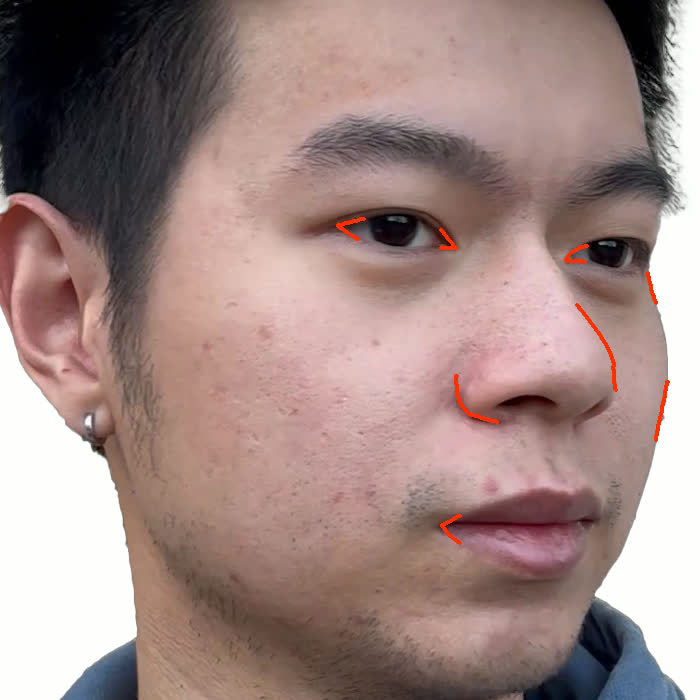}
    \end{minipage}%
    \hfill
    \begin{minipage}{0.32\linewidth}
      \centering \textbf{Ours}\\[0.3em]
      \includegraphics[width=\linewidth]{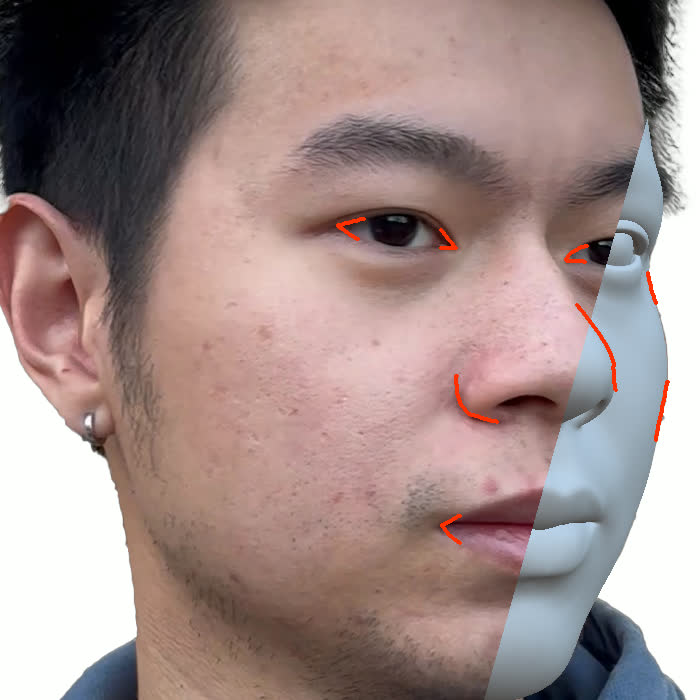}
    \end{minipage}%
    \hfill
    \begin{minipage}{0.32\linewidth}
      \centering \textbf{Joint}\\[0.3em]
      \includegraphics[width=\linewidth]{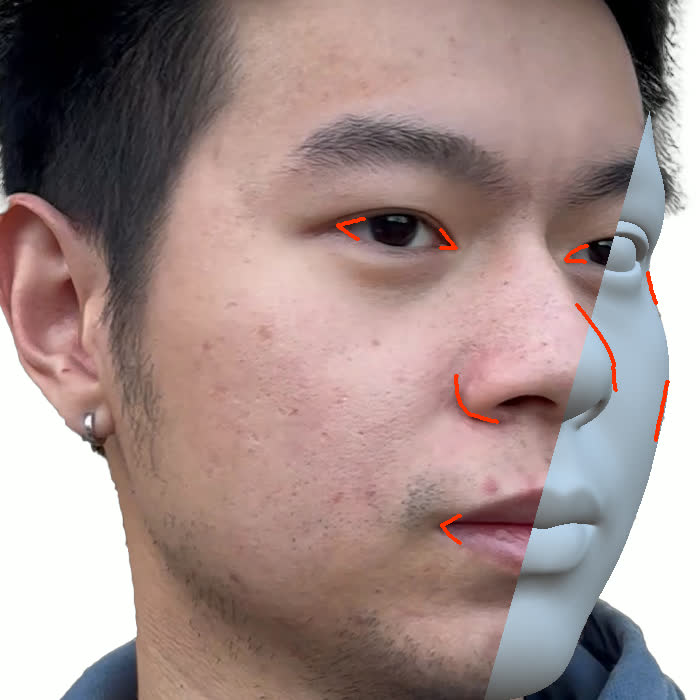}
    \end{minipage}
  \end{minipage}

  \caption{The middle image shows the result obtained using (only) our data capture strategy, while the far right image shows the result obtained when combining our data with the data obtained via flashlight capture. The additional flashlight capture data improves rigid alignment, particularly in side views, as can be seen from the better aligned eye and lip contours.}
    
  \label{fig:additional_data}
\end{figure}

 Our method is flexible enough to utilize images captured with disparate setups and lighting conditions. For example, we can combine images from our outdoor capture (see Sec.~\ref{sec:data_init}) with images acquired using flashlight illumination (see e.g.~\cite{han2024high}). Despite the differences, our approach is able to successfully integrate both datasets in order to produce a consistent geometry. Leveraging heterogeneous inputs facilitates the removal of spurious geometric variations, regularizing away slightly non-neutral facial expressions, small eye-shape differences, imperfections in camera intrinsics/extrinsics, etc. This is readily accomplished by allowing each Gaussian to maintain a separate view-dependent color for each disparate batch of images, while sharing all other attributes (opacity, position, scale, rotation, segmentation label, etc.). Importantly, our segmentation supervision provides reliable geometric cues, preventing the per-capture view-dependent color from overfitting. When one batch of images is deemed more reliable, its influence can be increased by sampling from it more often during training.  In fact, only the most reliable images should be used for the texture reconstruction, as regularization aims to blur important features. 
 See Fig.~\ref{fig:additional_data}.  

\subsection{Text-driven Asset Creation}

\begin{figure}[t]
  \centering


  \setlength{\tabcolsep}{2pt}
  \renewcommand{\arraystretch}{1.0}
  \begin{tabular}{C{0.28\linewidth} C{0.07\linewidth} C{0.28\linewidth} C{0.28\linewidth}}
  
    \multirow{2}{*}[0.6cm]{\begin{minipage}[b]{\linewidth}
      \coltitle{Reference Image}\\
      \includegraphics[width=\linewidth]{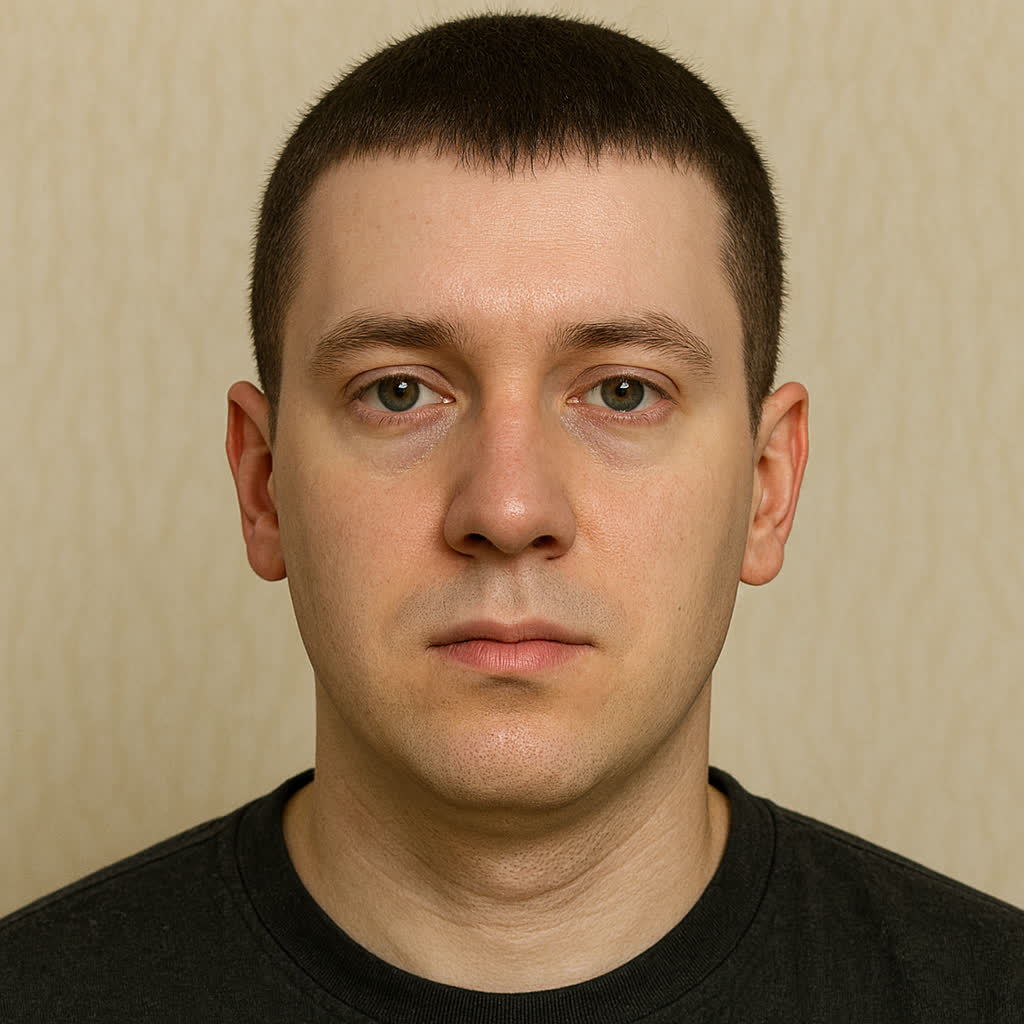}\\
    \end{minipage}} &
    \rowlabel{Images} &
    \includegraphics[width=\linewidth]{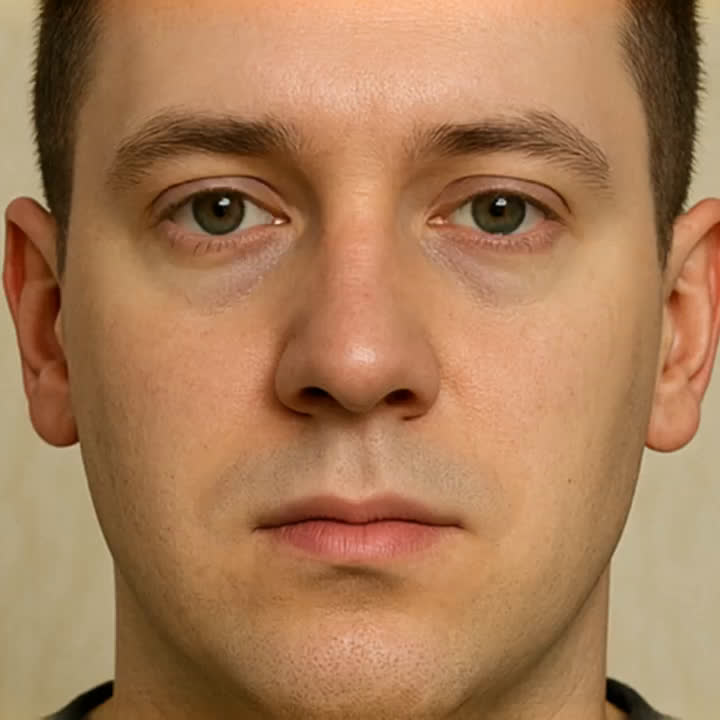} &
    \includegraphics[width=\linewidth]{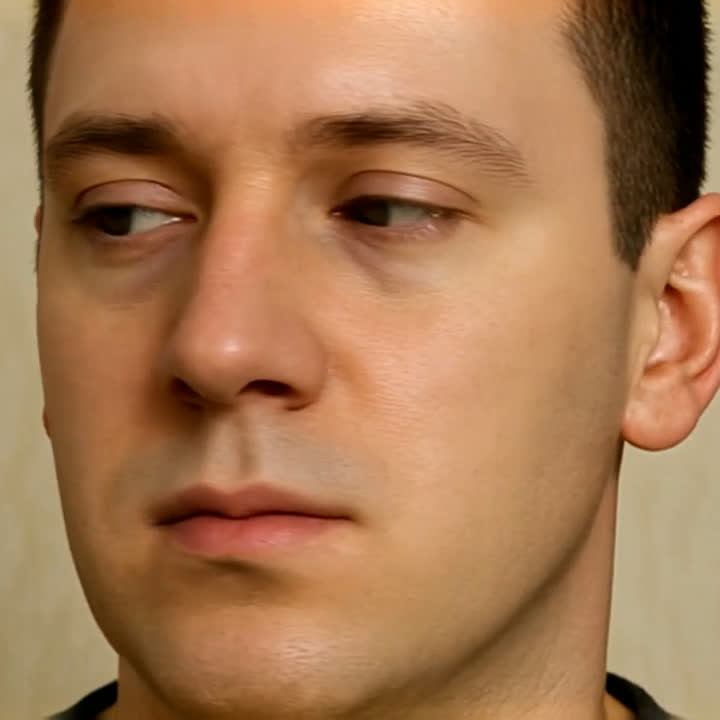}\\[0.35em] 

    & \rowlabel{MetaHuman} &
    \includegraphics[width=\linewidth]{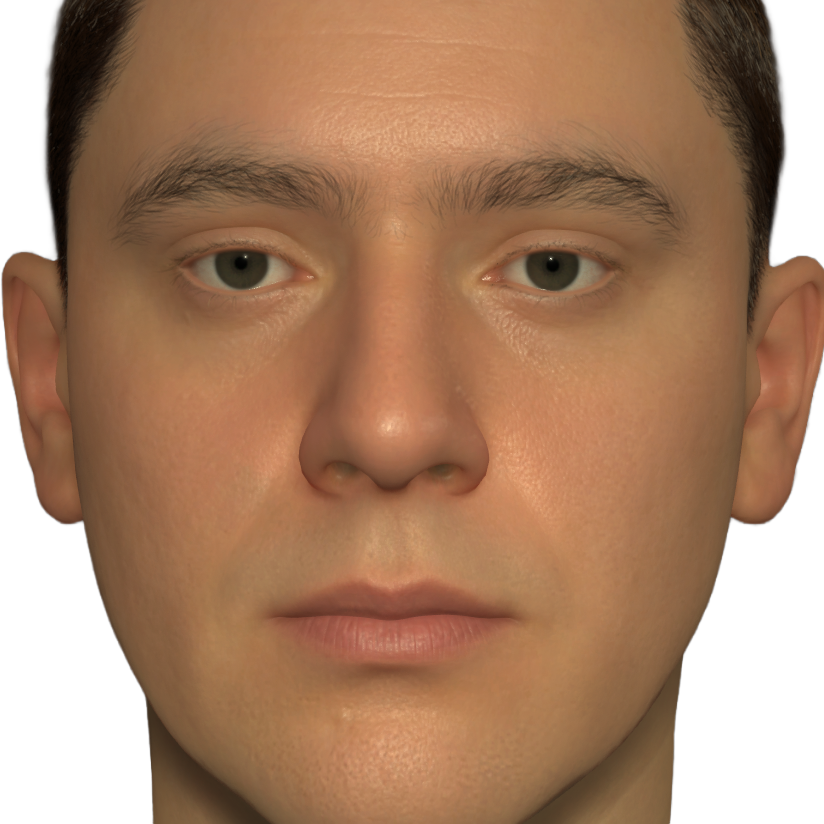} &
    \includegraphics[width=\linewidth]{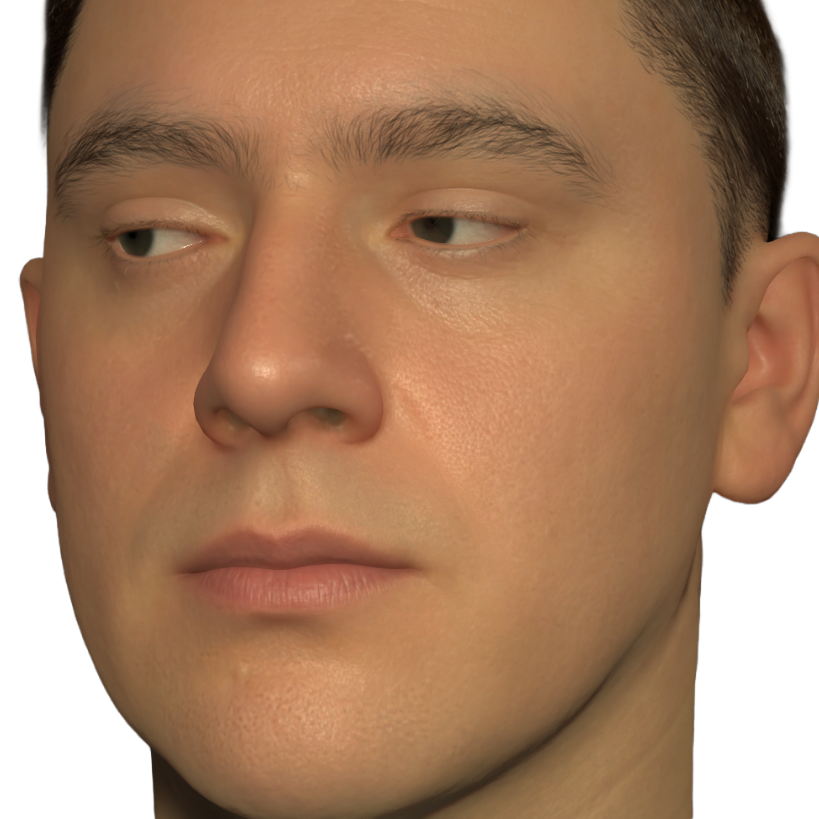} 
  \end{tabular}

  \caption{The reference image (left) was generated via a ChatGPT prompt, and it was subsequently used to create a video via Veo~3. The top row shows selected images from the video, and the bottom row shows the result of our pipeline. }
  \label{fig:text_to_metahuman}
\end{figure}

We demonstrate the efficacy of our approach by integrating it into a text-driven asset creation pipeline. First, we use ChatGPT~\cite{chatgpt2025} to generate an image with a neutral expression. Then, we use Veo~3~\cite{google2024veo} to create a video from the image using the instruction ``The actor rotates their head around.'' This creates data similar to our capture setup. Afterwards, our pipeline can be executed as usual. See Fig.~\ref{fig:text_to_metahuman}. 

\subsection{Additional Results}

\begin{figure}[t]
  \centering
  
  \begin{minipage}[t]{0.31\linewidth}
    \centering
    \coltitlelarge{Target Image}\\[0.3em]
    \includegraphics[width=\linewidth]{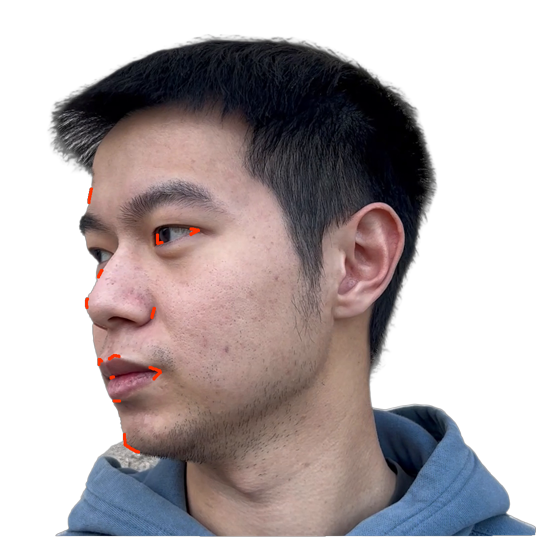}\\
    \includegraphics[width=\linewidth]{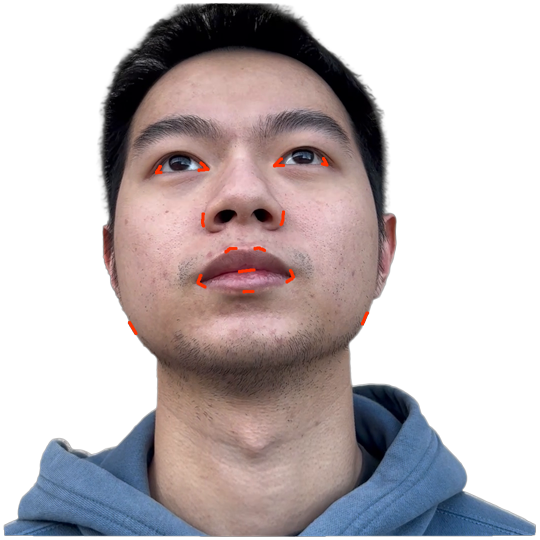}\\
    \includegraphics[width=\linewidth]{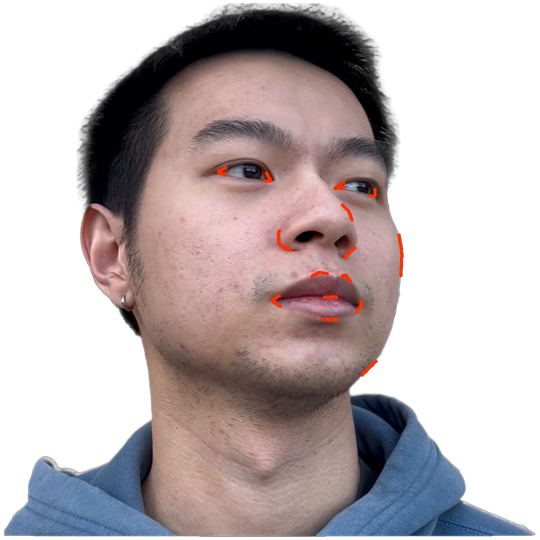}
  \end{minipage}%
  \hfill%
  \begin{minipage}[t]{0.62\linewidth}
    \centering
    \coltitlelarge{Ours}\\[0.3em]
    \begin{minipage}[t]{0.5\linewidth}
      \centering
      \includegraphics[width=\linewidth]{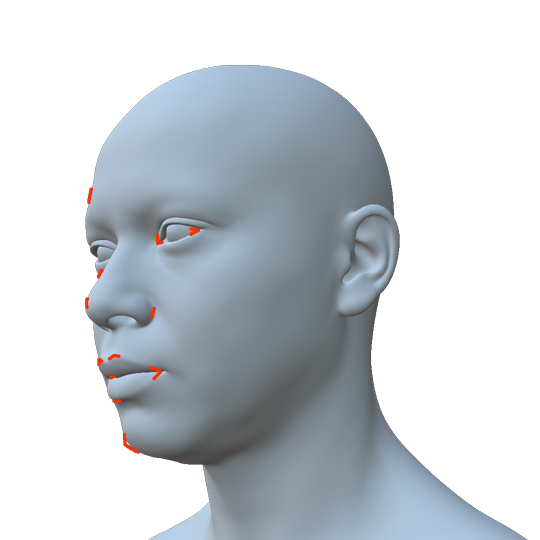}\\
      \includegraphics[width=\linewidth]{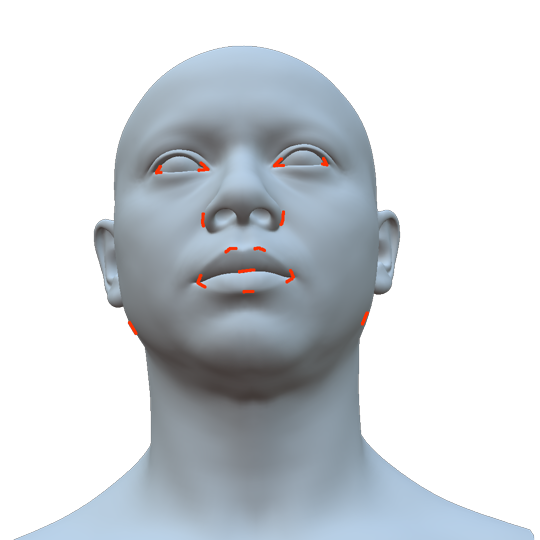}\\
      \includegraphics[width=\linewidth]{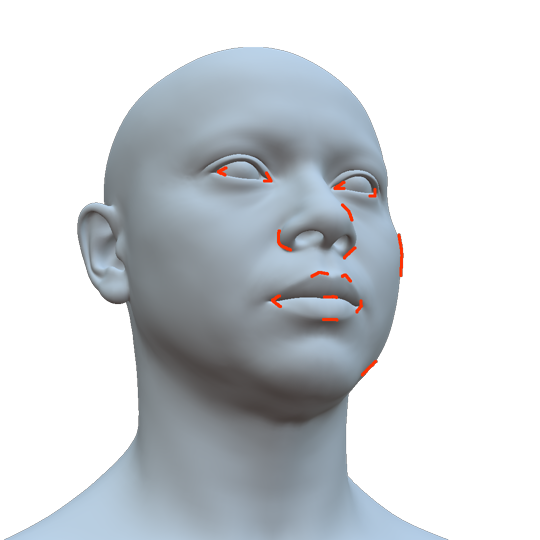}
    \end{minipage}%
    \begin{minipage}[t]{0.5\linewidth}
      \centering
      \includegraphics[width=\linewidth]{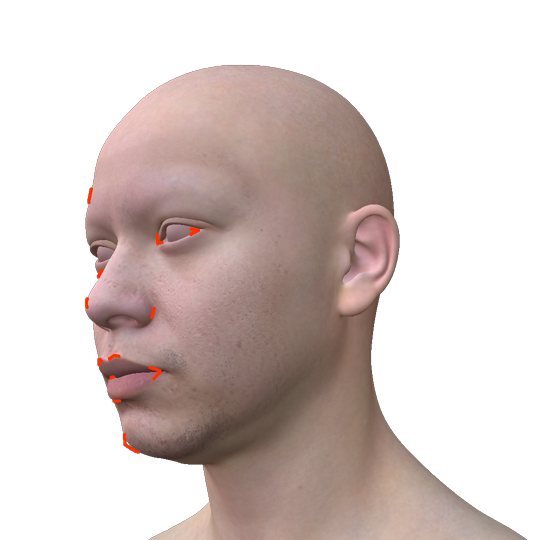}\\
      \includegraphics[width=\linewidth]{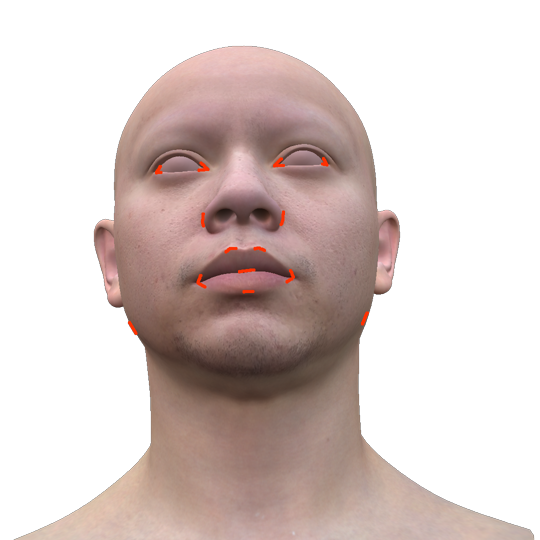}\\
      \includegraphics[width=\linewidth]{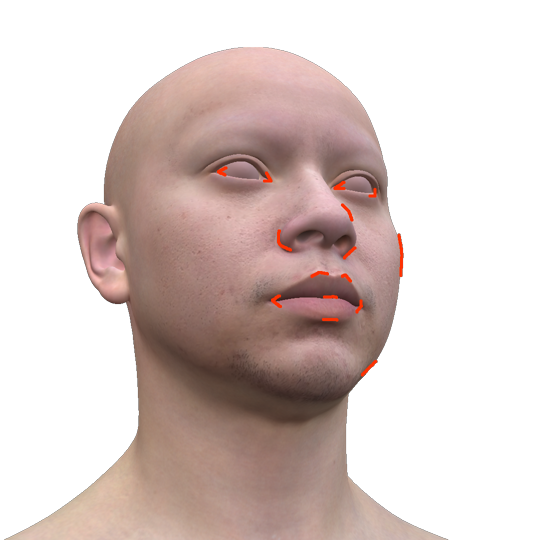}
    \end{minipage}
  \end{minipage}%

  \caption{Reconstruction results of the identity shown in Fig.~\ref{fig:nf_comparison} from additional views.}
  \label{fig:additional_views}
\end{figure}

We provide reconstruction results for the identity shown in Fig.~\ref{fig:nf_comparison} from additional views in Fig.~\ref{fig:additional_views}., and results for an additional identity in Fig.~\ref{fig:additional_identity}.

\begin{finalfigure}[t]
  \centering
  
  \begin{minipage}[t]{0.31\linewidth}
    \centering
    \coltitlelarge{Target Image}\\[0.3em]
    \includegraphics[width=\linewidth]{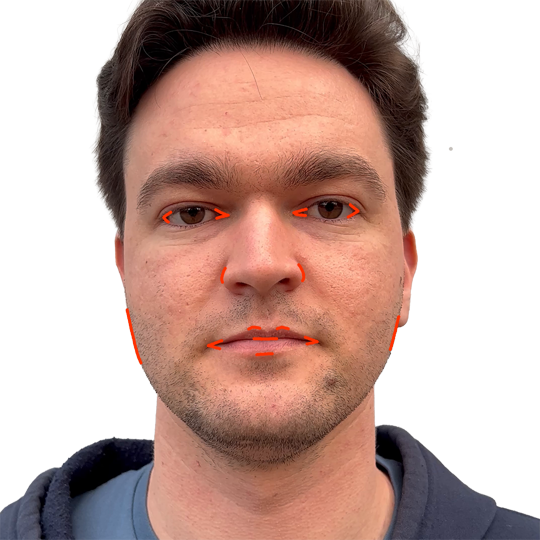}\\
    \includegraphics[width=\linewidth]{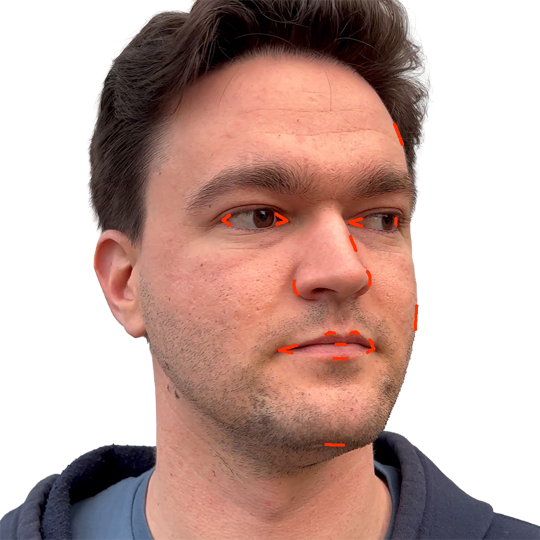}\\
    \includegraphics[width=\linewidth]{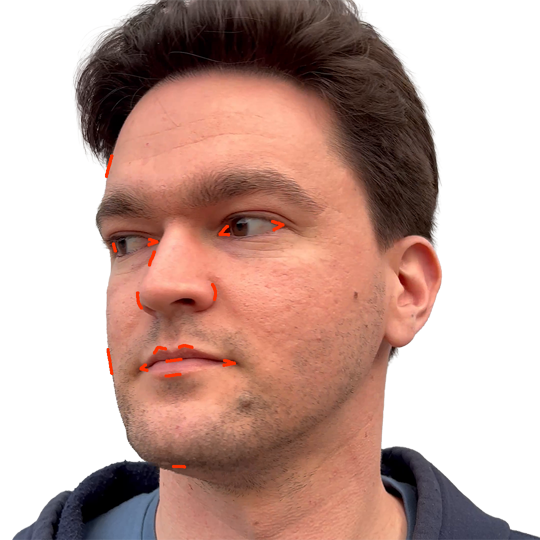}
  \end{minipage}%
  \hfill%
  \begin{minipage}[t]{0.62\linewidth}
    \centering
    \coltitlelarge{Ours}\\[0.3em]
    \begin{minipage}[t]{0.5\linewidth}
      \centering
      \includegraphics[width=\linewidth]{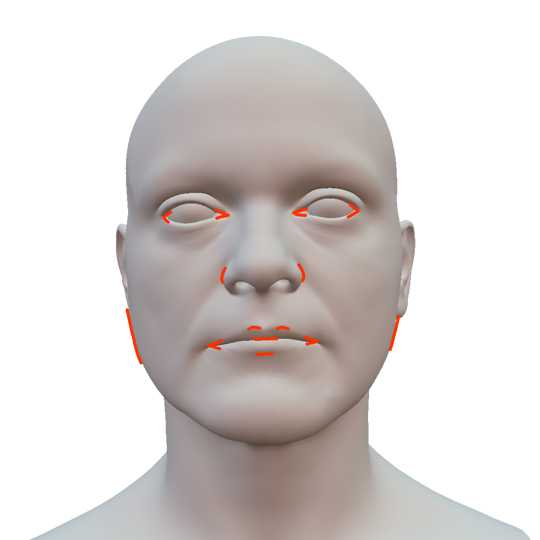}\\
      \includegraphics[width=\linewidth]{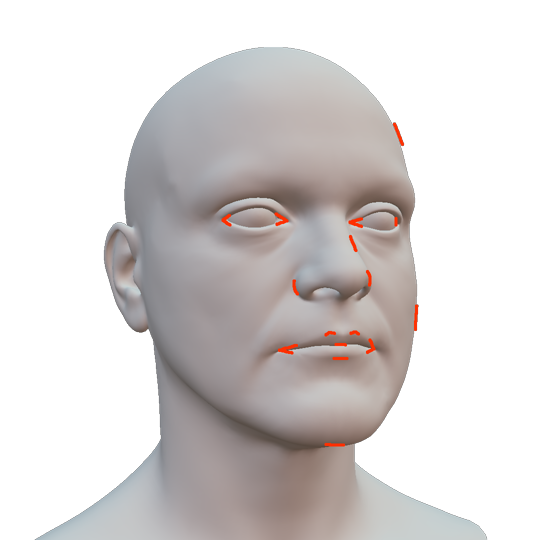}\\
      \includegraphics[width=\linewidth]{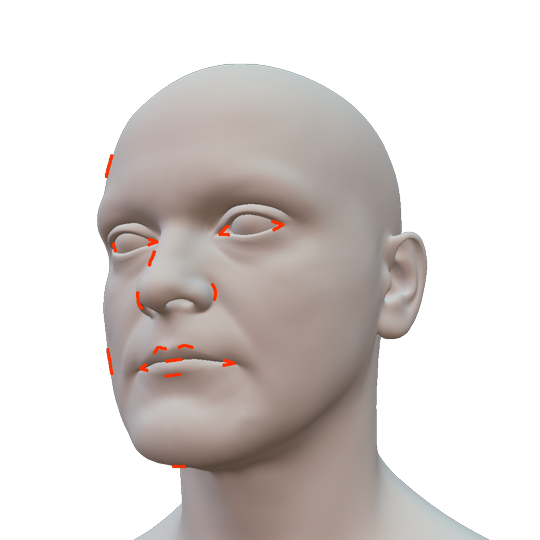}
    \end{minipage}%
    \begin{minipage}[t]{0.5\linewidth}
      \centering
      \includegraphics[width=\linewidth]{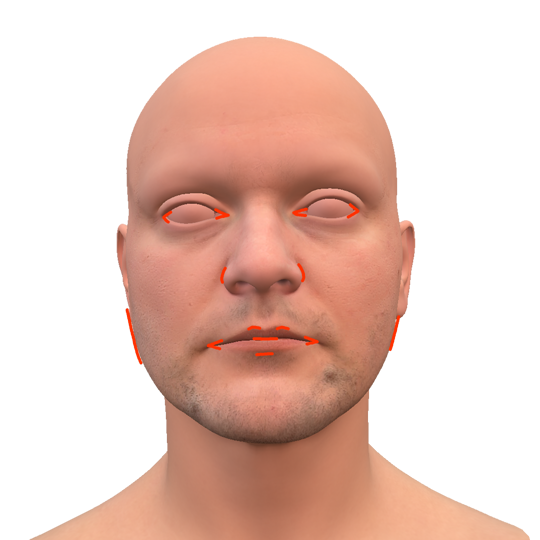}\\
      \includegraphics[width=\linewidth]{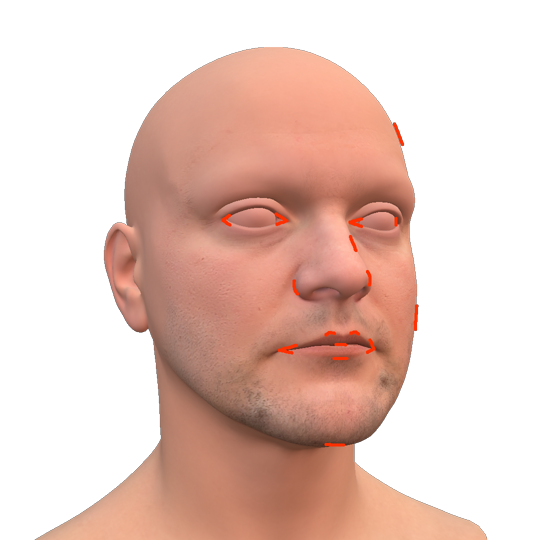}\\
      \includegraphics[width=\linewidth]{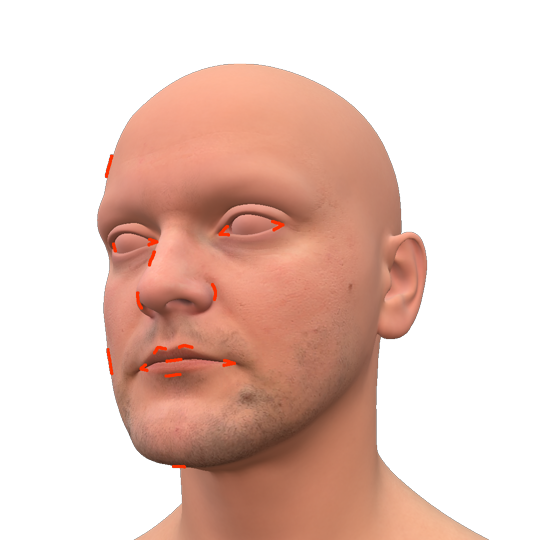}
    \end{minipage}
  \end{minipage}%

  \caption{Reconstruction results for an additional identity.}
  \label{fig:additional_identity}
\end{finalfigure}

\section{Limitations} 
In spite of the progress we have made addressing de-lighting, it remains difficult to address without a light-stage or similarly non-democratizable specialized equipment. Our de-lighting approach sacrifices some fine-grained geometric details, such as wrinkles. Although surface normal information could help to address this, Gaussians are not yet able to predict surface normals accurately enough to provide a remedy. 

The eyes and eyelids are the most difficult regions for our method to properly capture. This is because the overlapping Gaussians used in these regions do not have precise segmentation or strong silhouette guidance. While semantic segmentation does provide some supervision, it operates at a coarse granularity. Improvements could be achieved by incorporating high-quality, fine-grained landmark prediction. 

Our framework focuses primarily on the face and does not extract or regularize geometry for the hair, neck, etc. See e.g.~\cite{wang2024mega,junkawitsch2025eva,zhang2025fate} for various options/discussions. Gaussians in these areas remain unstructured and are thus not used to optimize the geometry of the triangulated surface. This may make camera pose estimation less reliable and sometimes cause the reconstructed geometry to deviate from the true head shape.

\section{Conclusion}

\begin{figure}[t]
  \centering
  \setlength{\tabcolsep}{0.01\linewidth} 
  \begin{tabular}{C{0.31\linewidth} C{0.31\linewidth} C{0.31\linewidth}}
    \coltitle{Original Image} &
    \coltitle{SwitchLight} &
    \coltitle{SwitchLight + Ours}  \\[0.4em]

    \multirow{2}{*}{\includegraphics[width=\linewidth]{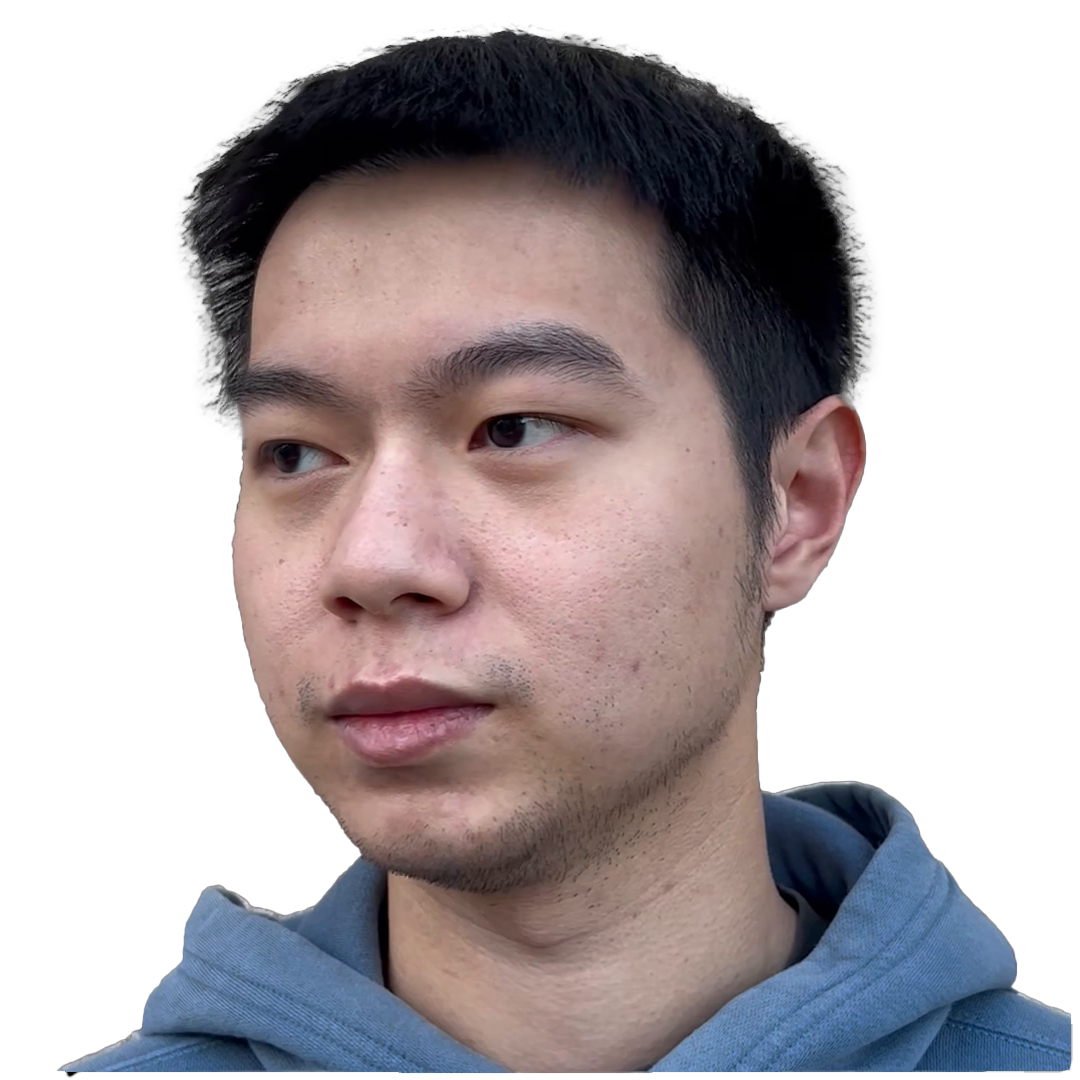}} &
    \includegraphics[width=\linewidth]{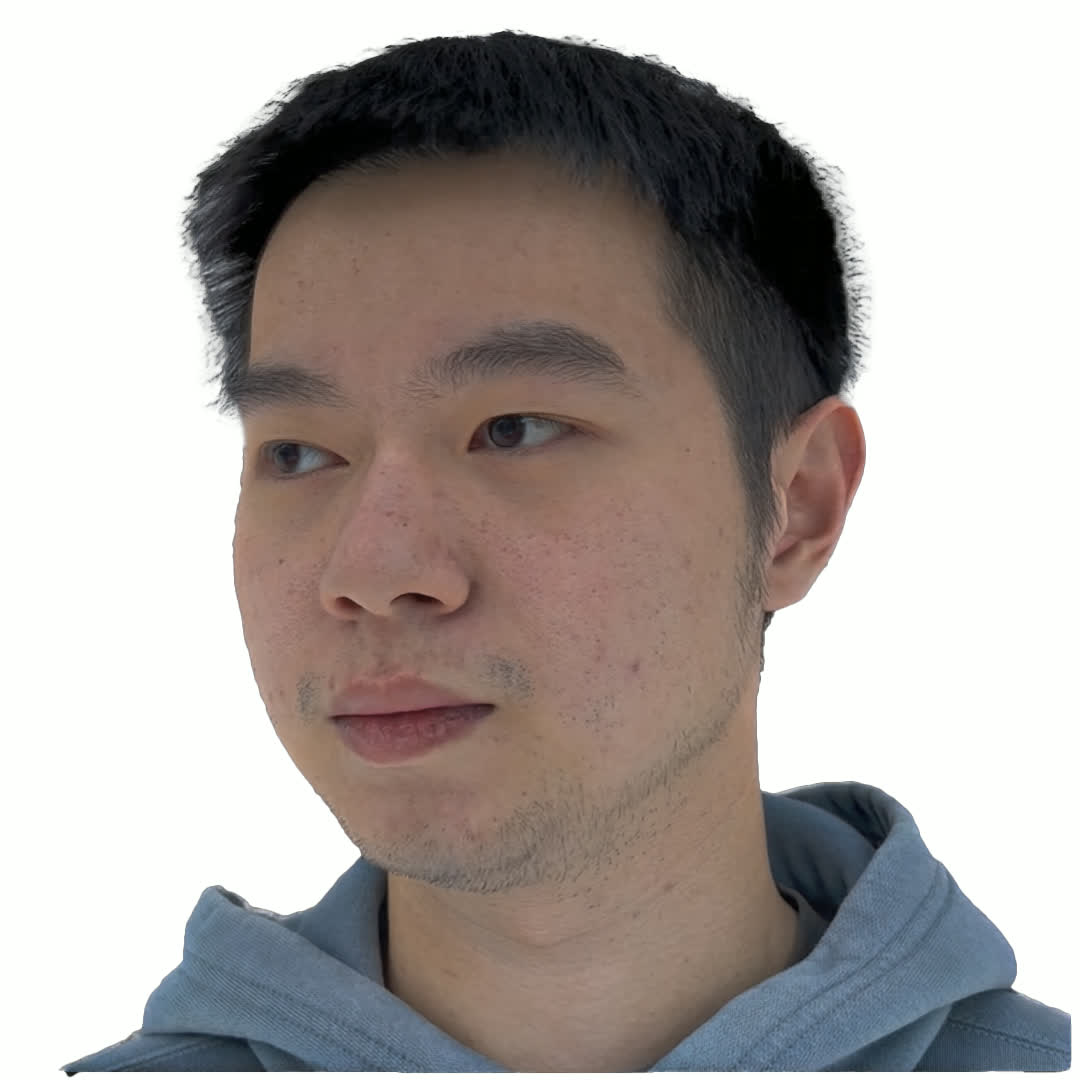} &
    \includegraphics[width=\linewidth]{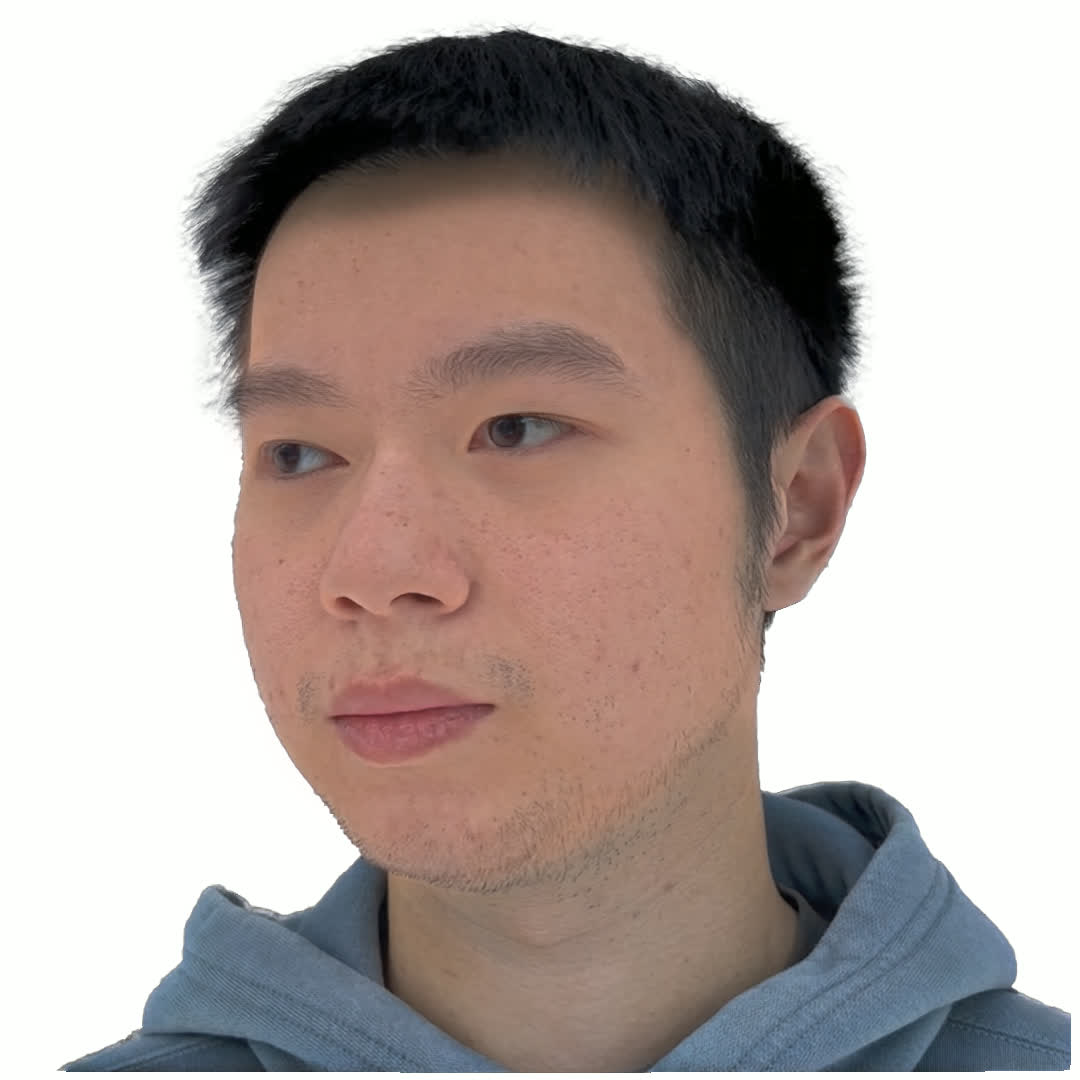} \\[0.35em]

    {} &
    \includegraphics[width=\linewidth]{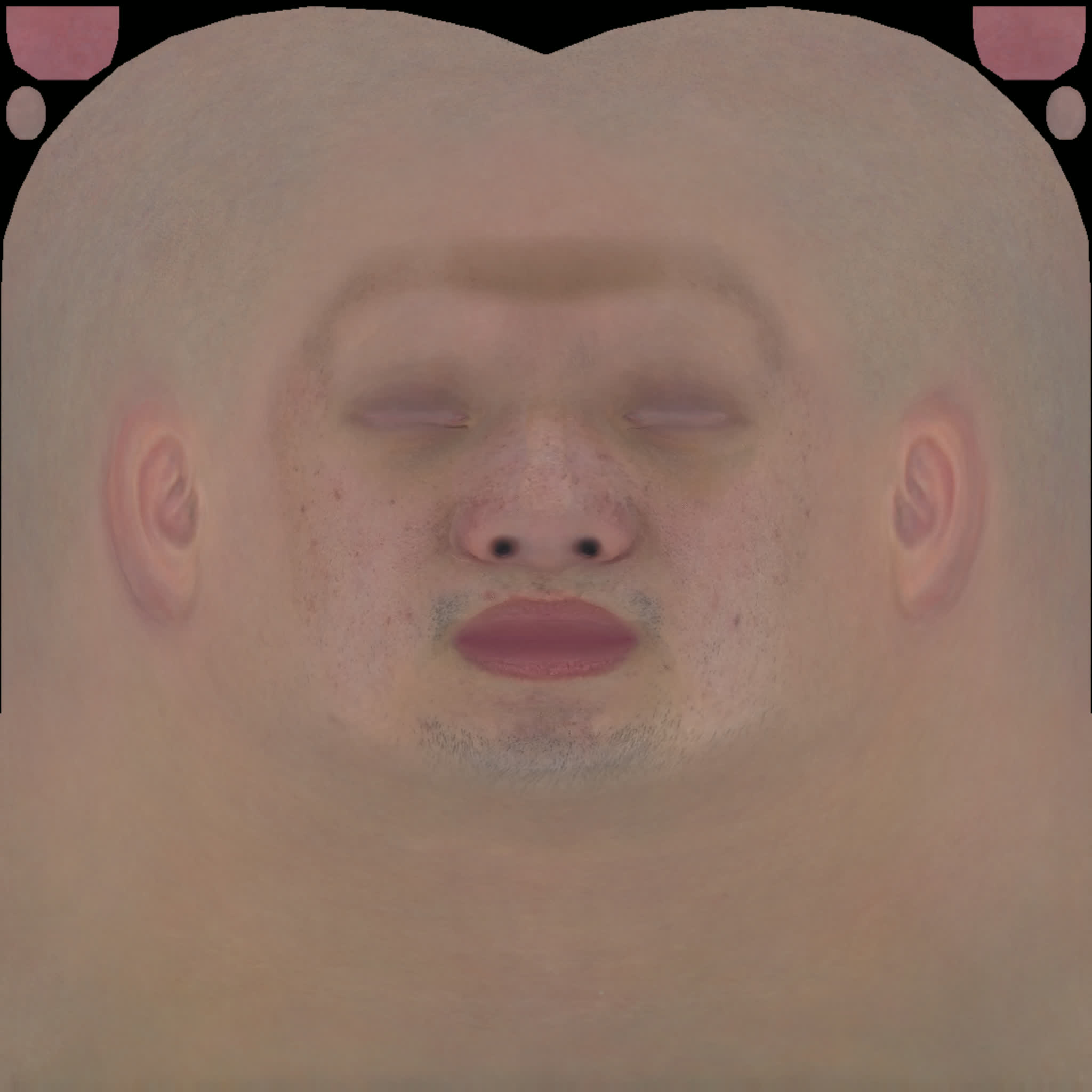} &
    \includegraphics[width=\linewidth]{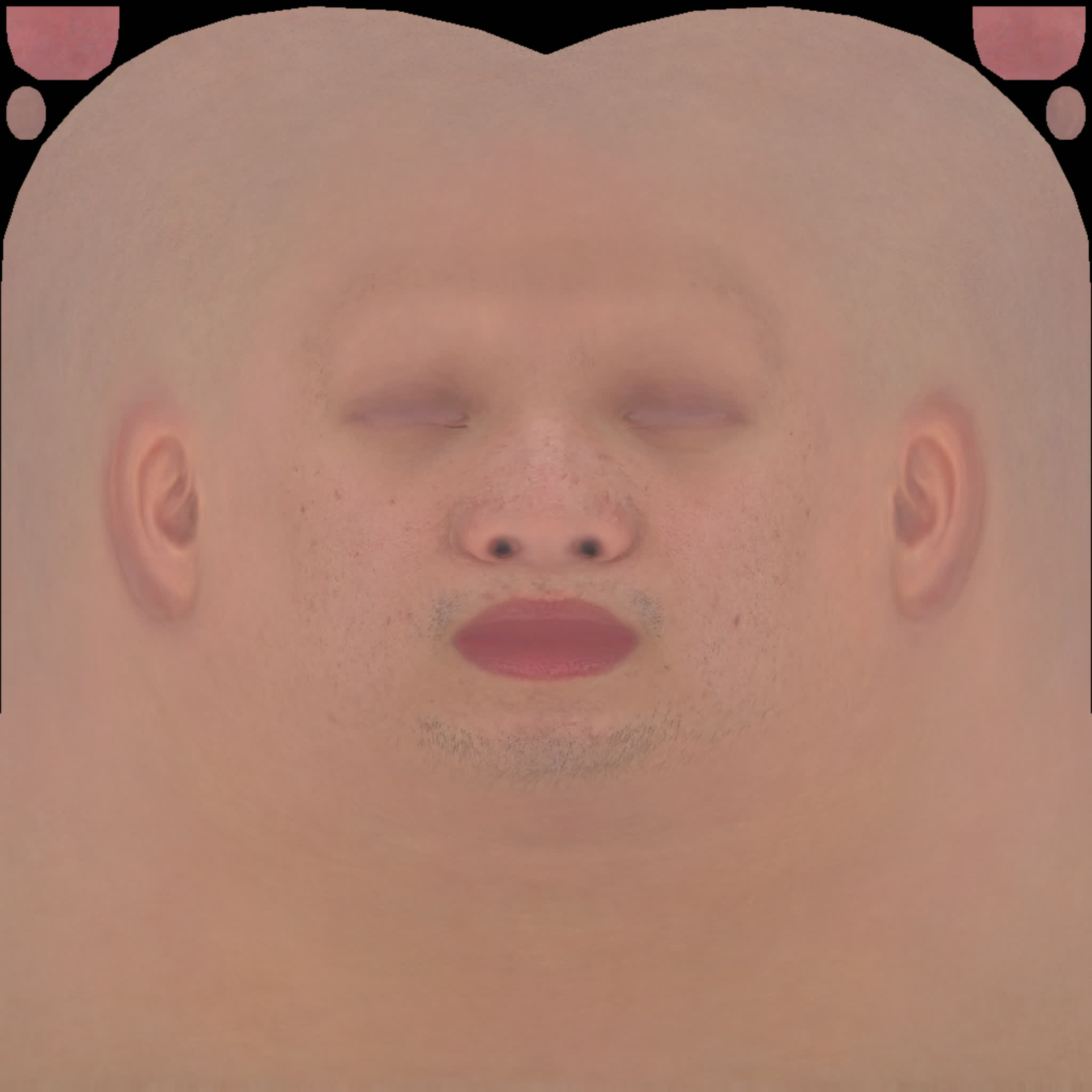} \\
  \end{tabular}

  \caption{Column 1 shows the results obtained using SwitchLight~\cite{kim2024switchlight} to preprocess the images before using~\cite{lin2022leveraging} to project and gather a texture. Note that the eyebrows, neck, hairline etc.~have been inpainted via the PCA-based projection approach discussed at the end of Sec.~\ref{sec:texture_opt}.  Column 2 shows the result obtained using SwitchLight to preprocess the images before fitting them into our pipeline. Although SwitchLight does a reasonable job, our method further suppresses baked-in shadows and aligns the texture with the MetaHuman color space (e.g. the SwitchLight result is too dark).}
    
  \label{fig:switchlight}
\end{figure}

Our method leverages the remarkable ability of Gaussian Splatting to explain a set of disparate images in a cohesive way with a disentangled view, essentially solving the camera extrinsics problem for democratizable approaches. Our modifications of Gaussian Splatting facilitate its use in geometry reconstruction. The semantic segmentation supervision allows the Gaussians to create accurate correspondences between the image and the triangulated surface. The soft constraints provide regularization and structure to the correspondence. Our geometry reconstructions compare favorably to the state-of-the-art. For texture, leveraging a PCA-based albedo prior and a Relightable Gaussian Splatting model allowed us to disentangle illumination from reflectance. This resulted in a cleaner, flatter, and more consistent de-lit texture than that which could be obtained by other democratizable methods not relying on controlled lighting or a light stage. Together, these design choices yield an accurate, relightable, and animatable reconstruction that generalizes well across subjects and lighting conditions, while remaining fully compatible with standard graphics pipelines.

Our approach is flexible enough to be used on a wide range of data sources and capture setups, readily accommodates disparate data, and can leverage inputs from other approaches. Notably, our ability to utilize disparate image data facilitates the hybridization of our approach with portrait-editing approaches (see e.g.~\cite{meng2024mm2latent,chaturvedi2025synthlight,rao2024lite2relight}). In particular, ~\cite{kim2024switchlight} achieves impressive portrait de-lighting and can be used to de-light images before they are used in our pipeline. See Fig.~\ref{fig:switchlight}.

Even in the case where Gaussian Splatting is preferred over the geometry and texture of a standard graphics pipeline, obtaining good geometry (as our method enables) is still rather useful. For example, better geometry provides better accuracy when driving Gaussian Splats during animation. Accurate geometry also allows, as shown in Sec.~\ref{sec:neural_texture}, the Gaussian Splats to be transformed into texture space where they can be used as a view-dependent neural texture. This allows any asset (or any subset of any asset) to be treated with Gaussian Splats without the need to modify any other assets or any other aspect of the graphics pipeline, facilitating immediate use of Gaussian Splats in high-end applications.

\section{Acknowledgement}
\begin{flushleft}
Research supported in part by ONR N00014-24-1-2644, 
ONR N00014-21-1-2771, and ONR N00014-19-1-2285. 
We would like to acknowledge Epic Games for additional support.
\end{flushleft}

{
    \small
    \bibliographystyle{ieeenat_fullname}
    \bibliography{main}
}

\end{document}